\documentclass[lettersize,journal]{IEEEtran}
\usepackage{amsmath,amsfonts}
\usepackage{algorithm}
\usepackage{algorithmic}
\usepackage{booktabs}
\usepackage{multirow}
\usepackage{array}
\usepackage[caption=false,font=normalsize,labelfont=sf,textfont=sf]{subfig}
\usepackage{textcomp}
\usepackage{stfloats}
\usepackage{url}
\usepackage{verbatim}
\usepackage{graphicx}
\usepackage{pifont}
\usepackage{xcolor}
\def\BibTeX{{\rm B\kern-.05em{\sc i\kern-.025em b}\kern-.08em
    T\kern-.1667em\lower.7ex\hbox{E}\kern-.125emX}}
\usepackage{balance}

\begin{document}
\title{Progressive$^2$: A Teacher-Student Progressive Co-Evolving Knowledge Distillation Method for Substantial Model Compression}
\author{Tiancong Cheng, \IEEEmembership{Student Member,~IEEE}, Ying Zhang\textsuperscript{*}, \IEEEmembership{Member,~IEEE}, Zhiwen Yu\textsuperscript{*}, \IEEEmembership{Senior Member,~IEEE}, Yifang Yin, Bin Guo, \IEEEmembership{Senior Member,~IEEE}

\thanks{Tiancong Cheng is with the School of Computer Science, Northwestern Polytechnical University, Xi'an 710129, China. E-mail: ctc1116@mail.nwpu.edu.cn.}
\thanks{Corresponding author: Zhiwen Yu and Ying Zhang.}
\thanks{This work has been submitted to IEEE Transactions on Services Computing for possible publication. Copyright may be transferred without notice, after which this version may no longer be accessible.}

}

\markboth{Journal of \LaTeX\ Class Files,~Vol.~18, No.~9, September~2020}%
{How to Use the IEEEtran \LaTeX \ Templates}

\maketitle

\begin{abstract}
Knowledge distillation (KD) is a widely utilized technique for transferring knowledge from a large model (the teacher) to a smaller model (the student). Owing to its flexibility and broad applicability, KD has been extensively applied in the compression of server-side models to meet the Quality of Service (QoS) requirements of client users. Despite significant advancements, the performance of distillation is substantially compromised when a large disparity exists between the capabilities of the server and the requirements of the client. To alleviate this problem, we propose a novel distillation approach, named Progressive$^2$, which operates through the combination of a progressively stronger teacher and a progressively smaller student. On the side of the teacher, rather than involving all layers simultaneously, we progressively select additional layers for distillation following a raw-to-rich semantic progression, establishing a systematic learning curriculum. Furthermore, we design a teacher-side multi-feature fusion adapter for the teacher to improve training stability, which is theoretically supported by the framework of Lipschitz continuity. On the side of the student, rather than directly training a tiny model, we gradually reduce the size of the network to facilitate an iterative co-evolution with the teacher. Progressive$^2$ serves as a flexible framework; the progressive strategy of the teacher can be deployed independently to achieve an optimal balance between accuracy and training efficiency, while the joint integration of the teacher and the student yields further improvements in overall performance.
\end{abstract}


\section{Introduction}
\IEEEPARstart{D}{eep} neural networks have demonstrated significant achievements in various {domains such as computer vision}~\cite{o2020deep,pasquadibisceglie2023jarvis}, data mining~\cite{wang2020deep,xia2021joint}, and cloud computing~\cite{xue2021ddpqn,huang2023secure}. {These advances have often relied on deeper or wider neural networks}, resulting in a significant number of parameters, which prevents their practical deployment in resource-limited scenarios ({e.g.,} robots, drone swarms, and low-cost computing devices)~\cite{lin2020mcunet}. {With the continued development of Internet of Things} (IoT) technology, cloud servers are increasingly delivering model compression~\cite{lu2022treenet} services to meet QoS requirements of mobile devices. {Server-side model compression services are commonly employed to equip mobile devices with lightweight and high-performance models~\cite{shuvo2022efficient}. Extensive research has been dedicated to developing smaller yet effective architectures. Knowledge distillation {has emerged as a widely adopted approach}~\cite{mirzadeh2020improved,wang2021knowledge}, effectively transferring knowledge from a large teacher model to a small student model.}

\begin{figure}[!t]
    \centering
    \includegraphics[width=3.5in]{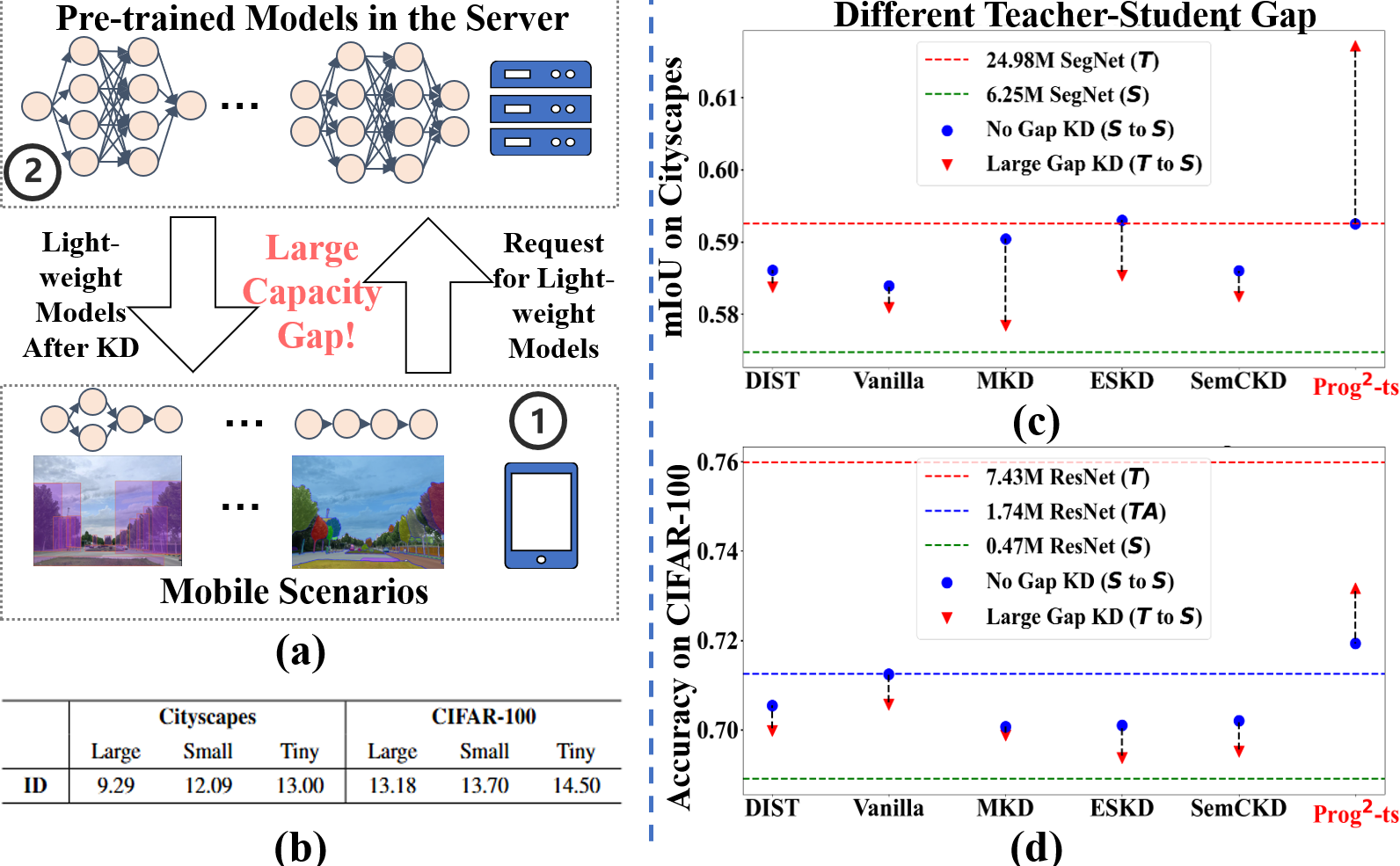}
    \vspace{-6mm}
    \caption{(a) Substantial compression of pre-trained models for mobile scenarios, highlighting the significant capacity gap between the models. (b) Differences in intrinsic dimensions (ID)~\cite{ansuini2019intrinsic} for models of varying sizes across different datasets. (c-d) Previous research (DIST~\cite{huang2022knowledge}, Vanilla~\cite{hinton2015distilling}, {MKD}~\cite{li2020knowledge}, ESKD~\cite{cho2019efficacy}, SemCKD~\cite{chen2021cross}) demonstrates that knowledge distillation performance diminishes when a substantial gap exists between teacher and student models. Experiments are conducted on {CIFAR-100}~\cite{krizhevsky2009learning} and Cityscapes~\cite{cordts2016cityscapes} using ResNet~\cite{he2016deep} and SegNet~\cite{badrinarayanan2017segnet} backbones.}
    \label{fig:background}
    \vspace{-3mm}
\end{figure}

Despite advancements in knowledge distillation, empirical performance typically degrades severely given a substantial capacity gap between teacher and student models (Fig.~\ref{fig:background} (c-d)). Recent methodologies calculating the intrinsic dimension (ID) of feature representations~\cite{facco2017estimating,ansuini2019intrinsic} quantitatively analyze this degradation by characterizing deep neural network training dynamics ({Fig.~\ref{fig:background}(b)}). A lower ID of feature representations extracted from the last hidden layer typically indicates superior predictive performance on the same task~\cite{ansuini2019intrinsic}. Although linking ID to precise generalization bounds via closed-form analytical proofs remains challenging, ID serves as an effective proxy for representation complexity~\cite{levina2004maximum,li2023rethinking}. {Conceptually, drawing on statistical learning theory~\cite{chen2022nonparametric}, a substantial ID mismatch between teacher and student is closely associated with increased distillation difficulty. Furthermore, empirical observations suggest that forcing a divergent high-dimensional student manifold to abruptly collapse into a specific low-dimensional teacher structure correlates with optimization instability~\cite{cho2019efficacy,mirzadeh2020improved}. Rather than implying a strict causal mechanism, this instability is often observed to coincide with gradient directional conflicts and surging local loss curvature.}

To bridge the teacher-student gap, recent methods design medium-sized assistant models for progressive distillation~\cite{mirzadeh2020improved,liu2023iterde}. {However, these studies leverage only hard or soft labels from the final layer for label-based distillation ($L$-$KD$)}, ignoring rich information from intermediate teacher layers~\cite{chen2021cross}. Thus, advanced methods require students to integrate knowledge from teacher layers, frequently via layer-wise feature alignment~\cite{zhu2021student,yang2022masked,he2022knowledge}. In these approaches ($F$-$KD$), knowledge transfers occur between corresponding teacher-student layer pairs or across multiple layers~\cite{ji2021show,chen2021cross,chen2021distilling}. However, as teachers typically comprise numerous layers, these methods introduce substantial parameters. Furthermore, student learning can suffer from inconsistencies across multi-layer features~\cite{zhu2021student}, diminishing training stability.

To overcome these challenges, we propose Progressive$^2$, which progressively and stably transfers a large teacher's multi-layer knowledge to a small student. The method features a progressively stronger teacher (a basic distillation framework) and a progressively smaller student (an optional enhancement), hence Progressive$^2$. The progressive teacher alone achieves a good balance between accuracy and training time, but accuracy further improves when combined with the progressive student.

Specifically, 1) \textbf{A progressively stronger teacher ($Prog^2$-$t$)}: To distill a small student, a progressively stronger teacher ($Prog^2$-$t$) is preferred, as utilizing a very strong teacher exacerbates the capacity gap~\cite{cho2019efficacy,huang2022knowledge}. Consequently, we introduce a curriculum over teacher representation levels, requiring the teacher to gradually incorporate additional layers during training. Empirically, Fig.~\ref{fig:crossKD}(a) demonstrates that utilizing shallow features with sparse semantics for early distillation warm-up yields higher performance gains than direct distillation. To capture comprehensive teacher knowledge~\cite{chen2021cross,ji2021show,ji2021refine,jacob2023online} while accommodating student training, we design a raw-to-rich layer involvement mechanism that progressively integrates additional teacher knowledge. Concurrently, a teacher-side knowledge adapter condenses the involved layers into an aggregated feature to guide the student. {Theoretical proxy analysis based on Lipschitz continuity suggests that this design reduces training complexity and enhances stability, while visualizations in Fig.~\ref{fig:crossKD}(d) illustrate the efficient weak-to-strong transfer of teacher feature knowledge.}

2) \textbf{A progressively smaller student ($Prog^2$-$ts$)}: While the previous strategy mitigates the teacher-student gap via progressive feature representations, progressive model capacities further reduce this disparity. Existing research indicates that gradually transferring knowledge via auxiliary models facilitates effective distillation~\cite{mirzadeh2020improved,li2023curriculum}. Inspired by this insight, we systematically reduce student size to promote its co-evolution with the progressively stronger teacher, achieving superior results. Notably, the progressively stronger teacher, establishing a curriculum over representation levels, can be deployed independently. In contrast, the progressively smaller student, forming a curriculum over model capacities, serves as an optional accuracy enhancement module. Our contributions are summarized as follows:

\begin{figure}[!t]
\centerline{\includegraphics[width=3.5in]{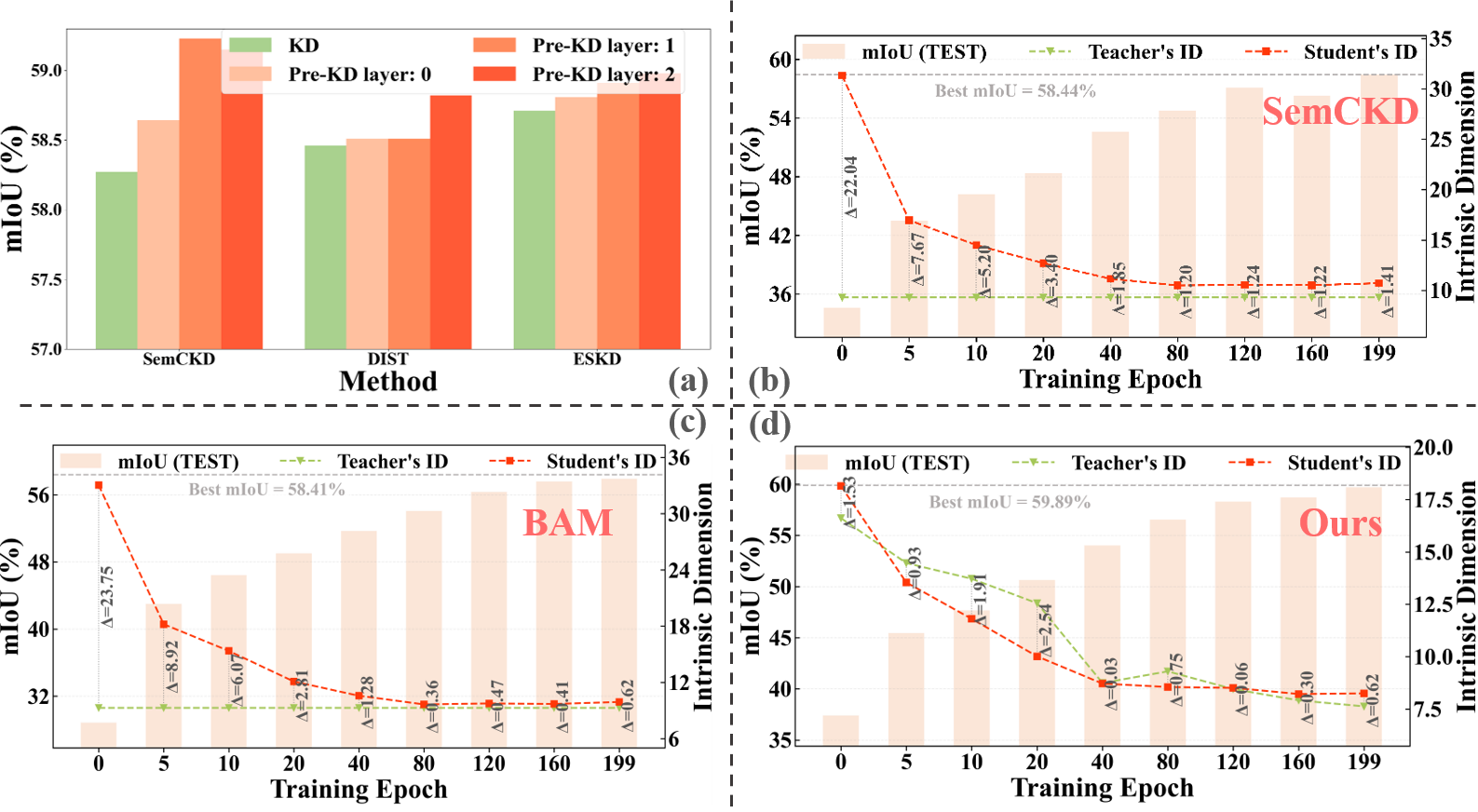}}
\vspace{-3mm}
\caption{{(a) Results of several classic distillation methods. For each method, the leftmost bar represents the original distillation result, while the three adjacent bars on the right illustrate the new results obtained by adopting a weakened teacher with a single shallow layer during the early training period. (b) Variation of the Intrinsic Dimension (ID)~\cite{ansuini2019intrinsic} of the student model's last hidden features with performance for classical methods SemCKD~\cite{chen2021cross} and BAM~\cite{clark2019bam} on the Cityscapes dataset~\cite{cordts2016cityscapes}. (d) Dynamic changes in the ID of the teacher's feature knowledge transferred to the student under our method.}} 
\label{fig:crossKD}
\end{figure}

\begin{itemize}
\item We introduce a novel method for knowledge distillation that progressively reduces the teacher-student capacity gap. This method achieves optimal performance through a teacher-student co-evolving strategy, while the progressive teacher can be deployed independently to balance accuracy and training efficiency. This flexibility perfectly accommodates the diverse requirements of Quality of Service (QoS) in practical mobile applications.
\item We design a multi-layer feature adapter for the teacher, which condenses the complexity of the multi-layer architecture and facilitates the training process with enhanced stability, as theoretically supported by the framework of Lipschitz continuity. This approach guarantees the quality of distillation services provided on the server side.
\item We evaluate the proposed method on multiple classic vision tasks using the Cityscapes, CIFAR-100, NYU-V2, and Tiny-ImageNet benchmark datasets. {Our proposed approach consistently outperforms state-of-the-art distillation methods.}
\end{itemize}

\section{Related Work}

Knowledge distillation compresses models by transferring information from a high-capacity teacher to a compact student. Existing approaches are mathematically categorized by transferred knowledge into logits-based distillation, feature-based distillation, and progressive schemes.

{\textbf{Logits-based Distillation ($L$-$KD$): Probability Distribution Matching.} Logits-based distillation $L$-$KD$ popularized by Hinton et al.~\cite{hinton2015distilling} minimizes the Kullback-Leibler (KL) divergence between the softened softmax outputs of the teacher ($P_T$) and the student ($P_S$). Utilizing a temperature parameter $\tau$ smooths the probability landscape, allowing students to learn rich inter-class relationships. Recent works refine the target distribution $P_T$ by avoiding excessive label smoothing~\cite{muller2019does}, decomposing the KL divergence loss to decouple gradient contributions~\cite{zhao2022decoupled}, or extending probability matching to cross-task settings~\cite{yang2022cross}. Despite their efficiency, $L$-$KD$ methods are limited to the final logits output. {Under the assumption that dimension-reduction layers, such as average pooling ($\theta \in \mathbb{R}^{CHW \times C}$) and fully connected projections ($\theta \in \mathbb{R}^{C_1 \times C_2}$, where $C_1 > C_2$), function as information bottlenecks, these methods may inadvertently attenuate the rich spatial and structural information embedded within intermediate layers during the generation of logits.}


\textbf{Feature-based Distillation: Intermediate Structural Alignment.}
Unlike logits semantic alignment, feature-based distillation~\cite{romero2014fitnets} enforces structural consistency in the intermediate representation space by minimizing the $L_2$ norm between the teacher feature maps ($F_T$) and the projected student features ($\phi(F_S)$): $\mathcal{L}_{feat} = ||F_T - \phi(F_S)||^2$. The transformation function $\phi(\cdot)$ bridges dimensional mismatches. Advanced methods optimize $\phi$ using Singular Value Decomposition (SVD) to reduce linear mapping redundancy~\cite{he2022knowledge} or align gradient directions to facilitate convergence~\cite{chen2021cross,liu2021conflict,liu2023famo}. However, simultaneously distilling multi-layer features causes distillation gradients to conflict with the primary task optimization gradients, expressed as $\text{Cosine\_Similarity}(\nabla \mathcal{L}_{task}, \nabla \mathcal{L}_{feat}) < 0$. {This directional conflict can unfavorably alter the local curvature of the task loss landscape, which often exacerbates training instability.}

\textbf{Progressive Distillation: Relaxed Optimization Constraints.}
To mitigate optimization difficulties caused by large capacity gaps (e.g., model capacity gap~\cite{cho2019efficacy,mirzadeh2020improved} or intrinsics dimension gap~\cite{ansuini2019intrinsic}), progressive distillation introduces a dynamic training curriculum. Progressive methods modulate transfer difficulty over time via a time-dependent loss weight $\lambda(t)$ or an evolving teacher target. Existing response-level strategies dynamically adjust the KL divergence loss weight~\cite{cho2019efficacy,clark2019bam,jacob2023online,li2023curriculum}, introduce intermediate-sized assistant models~\cite{mirzadeh2020improved,liu2023iterde}, or employ softened loss functions~\cite{huang2022knowledge,xu2023multi} to decompose optimization steps, {collectively establishing a weak-to-strong learning paradigm}.

While existing progressive methods optimize $\mathcal{L} = \lambda(t) \cdot \mathcal{L}(P_T, P_S)$, their scalar weight adjustments primarily scale the gradient magnitude ($\nabla \mathcal{L} = \lambda(t) \cdot \nabla \mathcal{L}_{KD}$). {In optimization dynamics, $\lambda(t)$ primarily modulates gradient step size. Although loss interactions and training schedules shape practical optimization, scalar weighting alone cannot explicitly reshape descent geometry or fundamentally alter convergence trajectories in the non-convex KD loss landscape. Consequently, particularly in severe cases involving an excessively large intrinsic dimension gap between teacher and student manifolds, simple magnitude scaling may be insufficient to fully overcome topological misalignment (e.g., as observed with BAM~\cite{clark2019bam} in Fig.~\ref{fig:crossKD} (c)).} Beyond conceptual scalar adjustments, our work introduces a unified analytical framework of feature space evolution. By establishing a stepwise evolution of feature representations, from simple to complex on the teacher side while iteratively adapting the student scale, our approach systematically reshapes the KD optimization landscape, mitigates severe curvature fluctuations, and maintains a manageable student capacity gap throughout the alignment process (Fig.~\ref{fig:crossKD} (d)).

\begin{figure*}[t]
\centering
\subfloat[$Prog^2$-$t$]
{
\label{fig:subfig1}\includegraphics[width=0.34\textwidth]{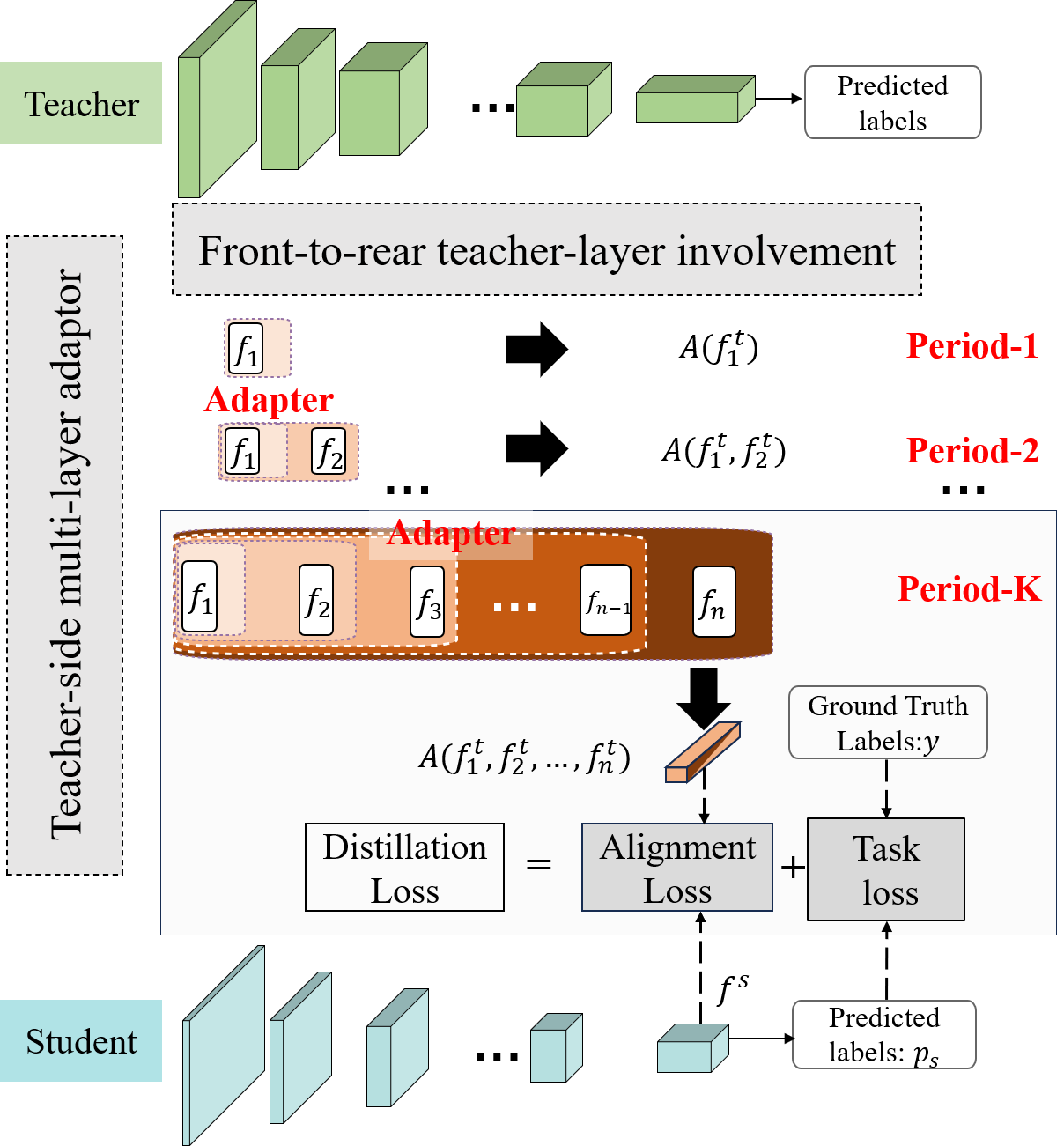}
}\hspace{20pt}
\hspace{10pt}\subfloat[$Prog^2$-$ts$]
{
\label{fig:subfig2}\includegraphics[width=0.41\textwidth]{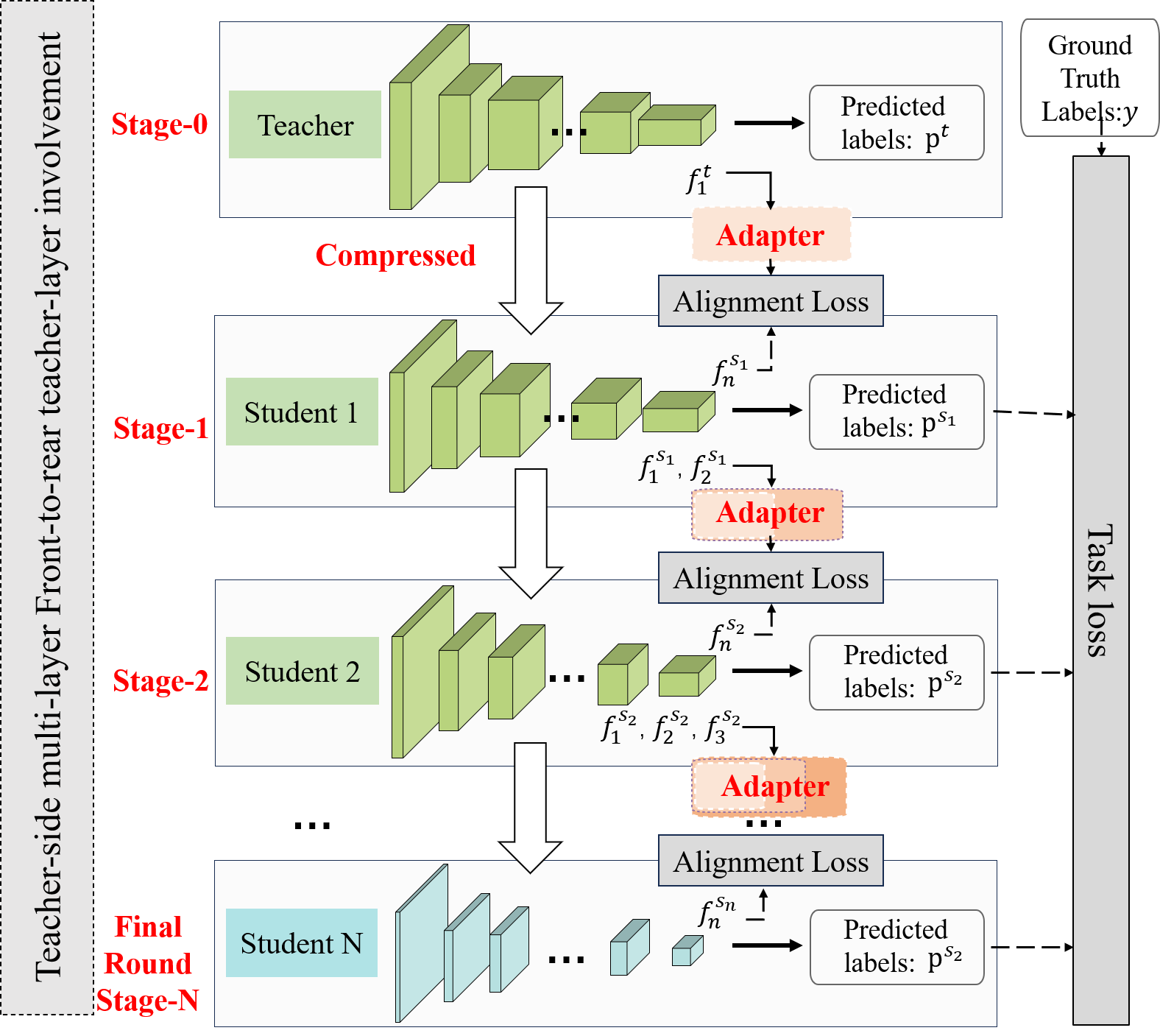}
}
\vspace{-3mm}
\caption{The overview of two versions of our proposed progressive$^2$ distillation approaches. (a) $Prog^2$-$t$ selects multiple teacher features in a {raw-to-rich} way and these features are integrated through a teacher-side multi-layer adapter which is then transferred to the student with better stability. (b) $Prog^2$-$ts$ enhances $Prog^2$-$t$ by gradually downsizing the student model while simultaneously assimilating an expanding pool of knowledge from the teacher model.}
\label{progressive}
\end{figure*}

\section{Method}
\label{sec:methods}
{To address the challenging distillation scenario featuring a large capacity gap between teacher and student models, we adopt a weak-to-strong curriculum distillation strategy. By progressively fusing the intermediate features of the teacher, we effectively reduce the complexity of the teacher knowledge from the perspective of intrinsic representation dimensionality (Fig.~\ref{fig:crossKD} (d)). Consequently, we propose Progressive$^2$, a novel distillation approach incorporating a progressively stronger teacher, serving as the basic framework detailed in Section~\ref{sec:one-stage}, and a progressively smaller student, acting as an optional enhancement detailed in Section~\ref{sec:multi-stage}. The progressive teacher (feature-level) can operate independently as the $Prog^2$-$t$ algorithm, but integrating the progressive student (model-level), denoted as the $Prog^2$-$ts$ configuration, further improves the overall predictive accuracy. Fig.~\ref{progressive} provides an overview of both algorithms, which are detailed in the subsequent subsections.}

\subsection{A progressively stronger teacher-based distillation framework ($Prog^2$-$t$)} \label{sec:one-stage}
{\subsubsection{A raw-to-rich teacher feature involvement} 
Raw-to-rich describes feature knowledge progression within the teacher model, with pixel-level semantic abstraction gradually increasing from shallow to deep representations. In standard deep learning architectures (e.g., VGG~\cite{simonyan2014very}, ResNet~\cite{he2016deep}), this corresponds to sequential feature outputs from initial to terminal blocks.} Assume we have a teacher model and its corresponding layer-index sets are denoted as $L^{t}$. To achieve the progressive distillation, we divide the training timeline into $K$ time-windows. Within each window, only a subset of teacher layers should be distilled and along the timeline, more and more layers will be included for distillation. Specifically, each time-window $T_i, i=1,\cdots, K$ will be associated with a subset of teacher layers(index), say $L_i^t$, from which the knowledge will be distilled to the student. Then, for the next period $T_{i+1}^t$, the $L_{i+1}^t$ is updated as following:
\begin{equation}
L_{i+1}^t = L_{i}^t + \Delta_{i+1}^t
\label{equ_l}
\end{equation} where $\Delta_{i+1}^t$ is an incremental layer-index set that is selected from the undistilled teacher layers. In this work, we require the $\Delta_{i+1}$ to be the next immediate adjacent teacher layers, and such ordered layer selection should be satisfied for all the time-windows until all teacher layers are included. As the layers in a neural network are stacked sequentially and the rear layers compose higher semantics, this work adopts a front-to-back layer involvement by selecting the front shallow layers first and moving the rear and deeper later. Therefore, the knowledge distillation loss at the current time-window $T_{i+1}$ is calculated as below:
\begin{equation}  \label{equ_stu} 
\mathcal{L}_{i+1}= \mathcal{L}_{task}(p^s,y)+\sum_{f_j^t\in  L_{i+1}^t} \alpha_j\times \mathcal{L}_{a}(\theta_{Map}(f^{s}),f_j^{t}) 
\end{equation} 
where $p^s$ and $y$ are the predictions of the student and the truth labels,$f^s$ is the feature from the last student layer with the mapping parameter $\theta_{Map}(\cdot)$ employed to align with the features of the teacher, $\theta_{Map}(\cdot)$ is conventionally conceptualized as $1\times1\;or\;3\times3$ convolution~\cite{li2020knowledge,chen2021cross,chen2021distilling}, a $1 \times 1 \; Conv$ is explicitly adopted in our architectural design, $f_{j}^t$ is
the feature map of $j_{th}$ teacher layer corresponding to the index
set $L_{i+1}^t (j \epsilon L_{i+1}^t)$ in the current time-window $T_{i+1}$. $\mathcal{L}_{task} (\cdot)$ is the task-specific loss ({e.g.,} dice loss for segmentation task, or cross-entropy loss for classification task). The hyperparameter $\alpha_j$ is the weight to adjust the layer-wise loss contribution and is determined empirically. The function $\mathcal{L}_{a}(x,y)$ refers to the feature alignment loss between two feature vectors $x$ and $y$, calculated by the Equation below:
\begin{equation}
\mathcal{L}_{a}\left ( \textsl{x}, \textsl{y}\right )=\small \left\| \:\frac{\textsl{x}}{\left\| \textsl{x}\right\|_{2}} - \frac{\textsl{y}}{\left\| \textsl{y}\right\|_{2}} \: \small\right\|_{2}^{2} 
\label{equ_l2}
\end{equation}  where $\left \| \cdot \right \| _2^2$ is the Euclidean distance function to calculate the distance between the L2 normalized feature maps.

\subsubsection{A teacher-side multi-layer adapter} \label{sec:tea_adapter} 
{This section explains why introducing excessive gradient directions during distillation can complicate model convergence}. {We further show how the proposed teacher-side adapter preserves the structural organization of teacher feature knowledge. This design also improves optimization stability and is naturally coupled with the raw-to-rich extraction mechanism described above. }
Eqn.~\ref{equ_stu} shows that the overall distillation is determined by the alignment level between the student feature and all selected teacher features (by $\theta_{Map}$). Such alignment is usually determined by a student-side adapter that maps the corresponding features~\cite{chen2021cross,chen2021distilling,li2020knowledge}. {Fig.~\ref{fig_Adapter} illustrates}three common adapter designs for mapping between a $m$ layered student and a $n$ layered teacher. {The teacher features are extracted immediately after each downsampling operation. Therefore, the number of teacher features varies across architectures and tasks, such as 10 for SegNet in semantic segmentation and 5 for ResNet in classification.} In Fig.~\ref{fig_Adapter}(a), one student {feature} learns from a single teacher {feature} through a mapping function without involving all teacher features knowledge. In Fig.~\ref{fig_Adapter}(b), each student {feature} is aligned to multiple teacher {features} and thus forms a $m$-to-$n$ mapping ~\cite{chen2021cross,ji2021show}. In Fig.~\ref{fig_Adapter}(c), all student {features} {are first integrated} and then mapped to multiple teacher {features} which form a 1-to-$n$ mapping. 

The above three designs initialize the mapping from the student's side and we name the mapping as \textbf{a student-side adapter}. {Since the feature distillation loss of each teacher feature represents a potential direction for the learning gradient~\cite{zhu2021student} and encapsulates unique knowledge, the student learns from multiple distillation directions.} {We hypothesize that a capacity-constrained student model faces challenges in extracting the most informative knowledge from numerous gradient directions, which typically leads to high training uncertainty and slow convergence (Fig.~\ref{fig_Adapter}(b,c)). We substantiate this hypothesis in subsequent Section~\ref{sec:proof} under the proxy assumption of Lipschitz continuity by analyzing the uncertainty, convergence rate, and stability.} To overcome the problem, we propose a \textbf{teacher-side adapter} that first integrates the multiple teacher layers and then aligns them only with the last student layer (Fig.~\ref{fig_Adapter}(d)). {This design captures comprehensive teacher knowledge, reduces optimization uncertainty, and accelerates convergence.}

\begin{figure}[t]
\centerline{\includegraphics[width=3.4in]{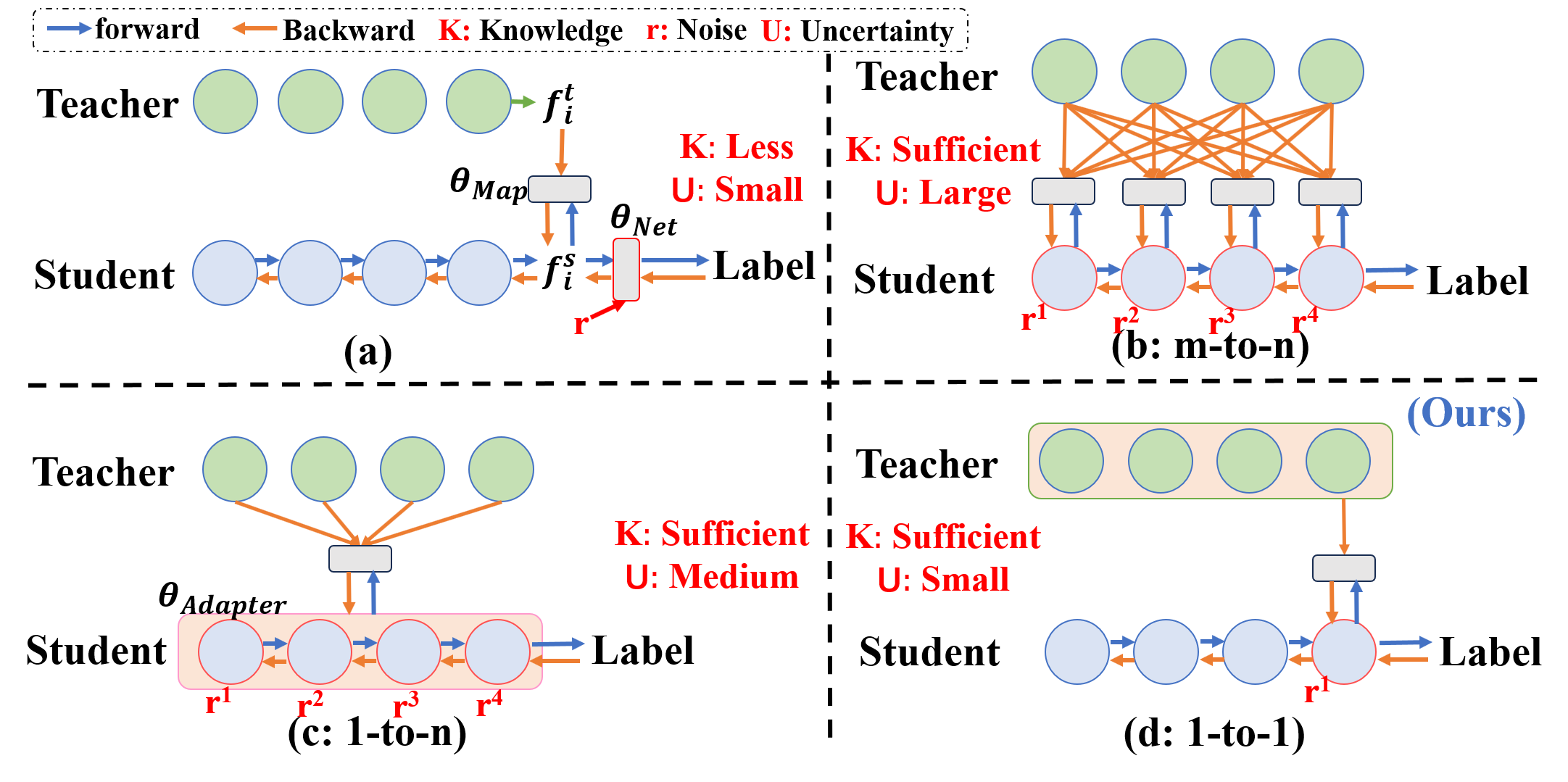}}
\vspace{-3mm}
\caption{{The backpropagation gradient conflict between teacher-student distillation and student-supervised learning.} 'U' and 'K' represent whether the uncertainty is large and the knowledge transfer is sufficient.} 
\label{fig_Adapter}
\vspace{-6pt}
\end{figure}

For the implementation of the teacher-side adapter, we design a simple but effective architecture as shown in Fig.~\ref{fig_FlowFPN}. For each teacher layer, it is firstly passed to a $1 \times 1$ convolution for channel alignment and then passed to a merge block which includes the nearest neighbor downsampling for dimension alignment (for $h$ and $w$), and $3 \times 3$ convolution for eliminating the discontinuity and aliasing caused by interpolation. Following each convolutional layer, a subsequent Batch Normalization (BN) layer and Rectified Linear Unit (ReLU) layer are used to enhance the model's generalization capability. The final output is a feature map that condenses the multi-layer knowledge of the teacher and is used to calculate the loss as in Eqn.~\ref{eq:teacher-adapter-loss}. $A(\cdot)$ represents our adapter function. Compared to Eqn.~\ref{equ_stu}, we only need to adjust one hyperparameter $\lambda$.
\begin{equation} \label{eq:teacher-adapter-loss}
    \mathcal{L}_{i+1}= \mathcal{L}_{task}(p^s,y)+\lambda \mathcal{L}_{a}(\theta_{Map}(f^{s}),A(f_1^{t},...,f_{|L^t_{i+1}|}^t)).
\end{equation}
Based on the above, we will distill a small student using the progressively stronger teacher with a front-to-back layer selection strategy and a multi-feature adapter, and we name the algorithm as $Prog^2$-$t$ (Algorithm~\ref{alg:t}).

\begin{figure}[!t]
\centerline{\includegraphics[width=2.9in]{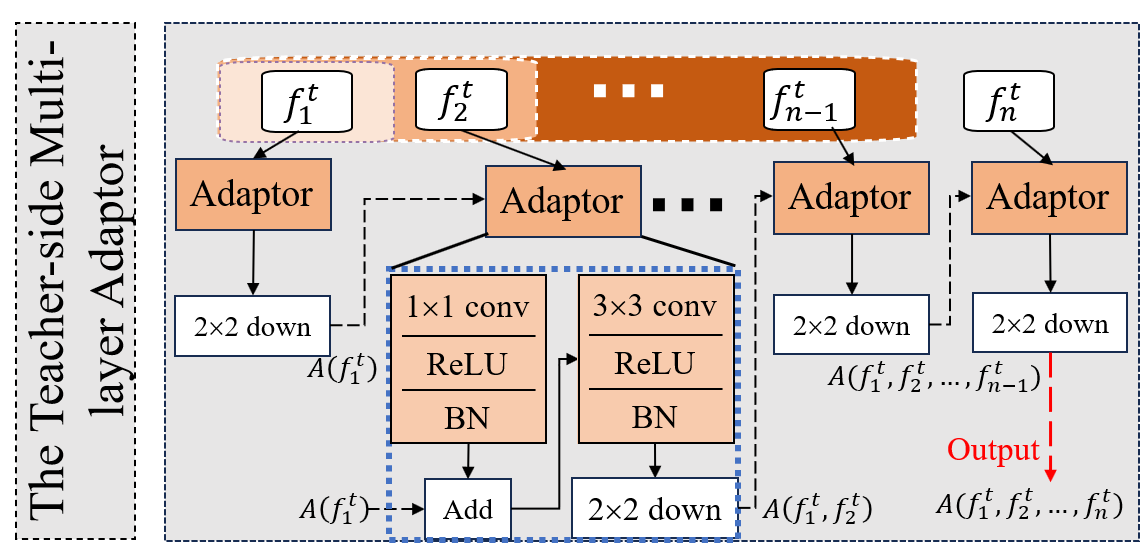}}
\vspace{-2mm}
\caption{The schematic diagram of our teacher-side multi-layer adapter with a bottom-to-up Feature Pyramid Network inspired integration\protect\cite{lin2017feature}.} 
\label{fig_FlowFPN}
\end{figure}

\subsection{A progressively smaller student co-evolved distillation enhancement ($Prog^2$-$ts$)} \label{sec:multi-stage}
The preceding sections detailed how $Prog^2$-$t$ can be utilized to distill a small student model by progressively and stably leveraging multiple intermediate {features} of the teacher. Throughout this training process, the capacity of the student model remains constant. {Existing studies demonstrate that a progression in model capacity, specifically through the introduction of medium-sized auxiliary teachers, effectively mitigates the capacity gap~\cite{liu2023iterde,cho2019efficacy,mirzadeh2020improved}. We corroborate this phenomenon by analyzing the intrinsic dimensions of the model features (Fig.~\ref{fig:background} (b)). Consequently, we enhance the feature-level progression of $Prog^2$-$t$ by gradually adjusting the size of the student model to facilitate a co-evolution with the progressively stronger teacher. We denote this enhanced algorithm as $Prog^2$-$ts$ (Fig.~\ref{progressive}(b)).}

The key idea of $Prog^2$-$ts$ is to expand the progressively stronger teacher setting from a single round training to multiple rounds. Here the round is the basic unit ({e.g.,} N epochs) and it contains multiple time-windows in $Prog^2$-$ts$. Assume the training is repeated $N$ rounds and denote the teacher and the student in the $k^{th}$ round is $t_k$ and $s_k$, respectively for $k=[1, N]$. In each round, 1) student-wise, we gradually reduce its model size, say $s_{k} = \Phi(s_{k-1}, g(r))$, where $\Phi(\cdot)$ is a compression operator (e.g, channel compression in this work), $r\in[0,1]$ is a compression ratio required by the application, $g(r)$ is a function to determine the current round compression rate. 2) Teacher-wise, considering the student model is updating, we use the last student as the new teacher to further reduce the gap, i.e.\,, $t_{k} = s_{k-1}$ and the distillation layer index of $t_{k}$ are incremented included as below: 
\begin{equation}\label{eq:layerset-ts}
    L^{t_k} = L^{t_{k-1}} + \Delta^k
\end{equation} where $\Delta^k$ is the incremental layer-index set determined by our previously introduced {raw-to-rich} selection strategy. Note that although the teacher is updated in each round, during the distillation, we only use partial teacher layers instead of all of them. Based on the above, the eventual distillation loss of $Prog^2$-$ts$ is calculated by the following:
\begin{equation} \label{eq:teacher-adapter-loss-multi}
    \mathcal{L}^k= \mathcal{L}_{task}(p^s,y)+\lambda \mathcal{L}_{a}(\theta_{Map}(f^{s_k}),A(f_1^{t_k},...,f_{|L^{t_k}|}^{t_k})), 
\end{equation} where $|L^{t_k}|$ is the number of to-be-distilled layers of the teacher $t_k$. 
We summarized the full version of $Progressive^2$ in Alg.~\ref{alg:ts} where the input is the teacher model, a compression ratio, and the number of repeated rounds. The output is a well-distilled student model. In each of the following rounds, $Prog^2$-$ts$ first determines the student model by a compression operator (Line 4) and uses the last round student as the teacher model (Line 5). Then it determines the distillation layer set of the teacher using our {raw-to-rich} selection strategy (Line 5). Distillation training is then carried out by optimization of the distillation loss function as in Eqn.~\ref{eq:teacher-adapter-loss-multi} (Line 6). Eventually, we will obtain a well-distilled student model that satisfies the requirement compression ratio. 

\begin{algorithm}[!t]
\caption{The algorithm of $Prog^2$-$t$.}
\label{alg:t}
\begin{algorithmic}[1] 
\REQUIRE ~~\\ 
    A large pre-trained teacher model $T$. The compressed student model $S$.
    \STATE \textbf{Initialization of $S$.} 
    \FOR{period $i=1,2,...K$}
        \STATE set time-window $T_i$ with the subset of teacher layers(index): $L_i^t=L_{i-1}^t+\bigtriangleup_i^t$.
        \STATE get the fused teacher feature: $f^t=A(f_1^{t},...,f_{|L^t_{i}|}^t)$.
        \STATE Training $S$ by optimizing the loss in Eqn.~\ref{eq:teacher-adapter-loss}.
    \ENDFOR
\RETURN Target well distilled student model $S$.
\end{algorithmic}
\end{algorithm}

\subsection{$Prog^2$-$t$ vs.$Prog^2$-$ts$ } 
The progressive teacher in Sec.~\ref{sec:one-stage} can be used alone ($Prog^2$-$t$), or together with the progressive student in Sec.~\ref{sec:multi-stage} ($Prog^2$-$ts$). Both versions could achieve superior accuracy than the recent advanced methods (as seen in Experiments). The full version $Prog^2$-$ts$ can further improve $Prog^2$-$t$ in terms of accuracy but with additional training time cost. In practice, the version selection is determined by the accuracy and efficiency requirement of the downstream application on mobile clients.


\begin{figure*}[!t]
\centerline{\includegraphics[width=6.0in]{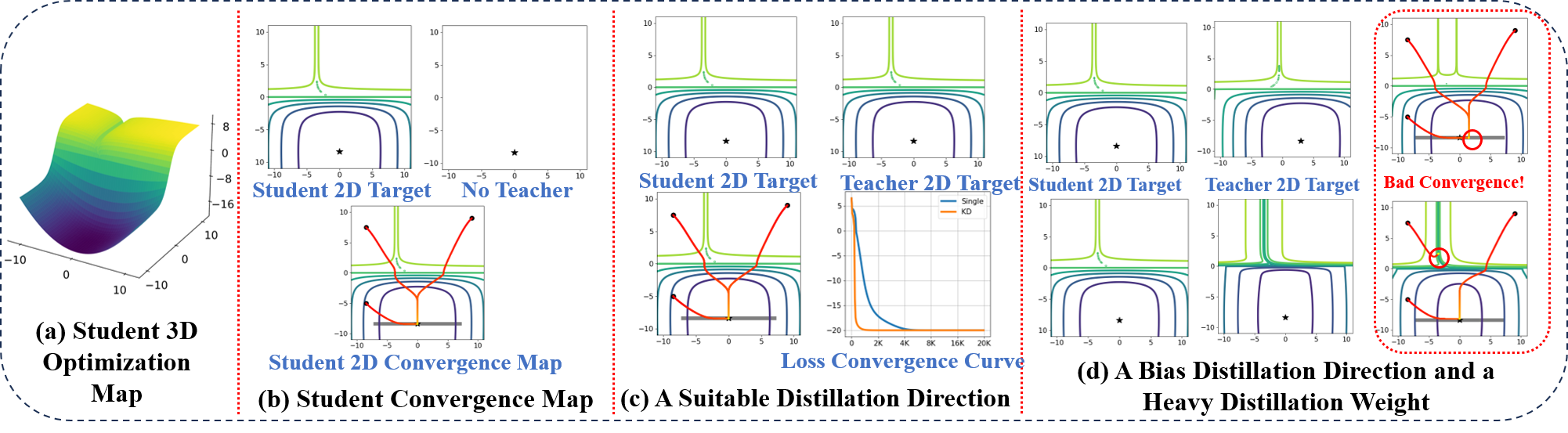}}
\caption{Illustration of why getting a suitable gradient direction from knowledge distillation loss is hard.} 
\label{fig_toy}
\vspace{-10pt}
\end{figure*}

\begin{algorithm}[!t]
\caption{The algorithm of $Prog^2$-$ts$.}
\label{alg:ts}
\begin{algorithmic}[1] 
\REQUIRE ~~\\ 
    A large pre-trained teacher model $T$. The stage number $n$. The compression ratio $r$.
    \STATE \textbf{Initialization:} 
    \STATE $t_0=T$, $s_0=t_0$, $g(r) = \frac{r}{n}$
    \FOR{$k=1,2,...n$}
        \STATE set the student model by channel compression: $s_k=\Phi(s_{k-1}, g(r))$
        \STATE set teacher model: $t_k = s_{k-1}$, determine layer (index) set by Eqn.~\ref{eq:layerset-ts}.
        \STATE Training $s_k$ by optimizing the loss in Eqn.~\ref{eq:teacher-adapter-loss-multi}.
    \ENDFOR
\RETURN Target compressed student model $S=s_{n}$.
\end{algorithmic}
\end{algorithm}

\subsection{{Instability Quantification and Theoretical Proxy Analysis of the Teacher-Side Adapter}} \label{sec:proof}

The analytical progression, as outlined in the flowchart (Fig.~\ref{fig_flowchart}), is structured as follows: we first demonstrate through a toy example that the distillation gradient inherently introduces noise into the optimization of the primary task and quantify the associated uncertainty. Subsequently, we utilize a proxy analysis based on Lipschitz continuity to establish that multi-layer distillation approaches (specifically, $1$-to-$n$ and $m$-to-$n$ configurations) amplify this gradient noise and correspondingly decelerate model convergence. {As these conditions are not strictly guaranteed in deep non-convex neural networks, the ensuing convergence and uncertainty results serve as theoretical proxies elucidating optimization dynamics, rather than formal guarantees.}

\noindent \textbf{A toy example of distillation gradient noise and empirical quantification of instability.} \label{sec:instability}
We first present an example in Fig.~\ref{fig_toy}: (a) illustrates the 3D optimization landscape, where deeper colors indicate greater curvature; (b) visualizes the 2D optimization convergence map of an independent student model, demonstrating trajectories from three distinct initial points toward the star-marked optimal target; (c) demonstrates a distillation gradient perfectly aligned with the optimization path of the student model, accelerating convergence; and (d) depicts scenarios where the direction of the distillation gradient is biased or the distillation strength is inappropriate, causing the student model to converge toward a suboptimal solution or diverge from the optimal target. Therefore, the stability of the distillation process is fundamentally influenced by first-order gradient conflicts and further modulated by the magnitude of the distillation gradient and second-order curvature. Previous works address this instability through conservative distillation losses~\cite{huang2022knowledge} or adaptive strength scheduling~\cite{clark2019bam,li2023curriculum} but do not quantify its changes as training.

\begin{figure}[!t]
\centerline{\includegraphics[width=3.3in]{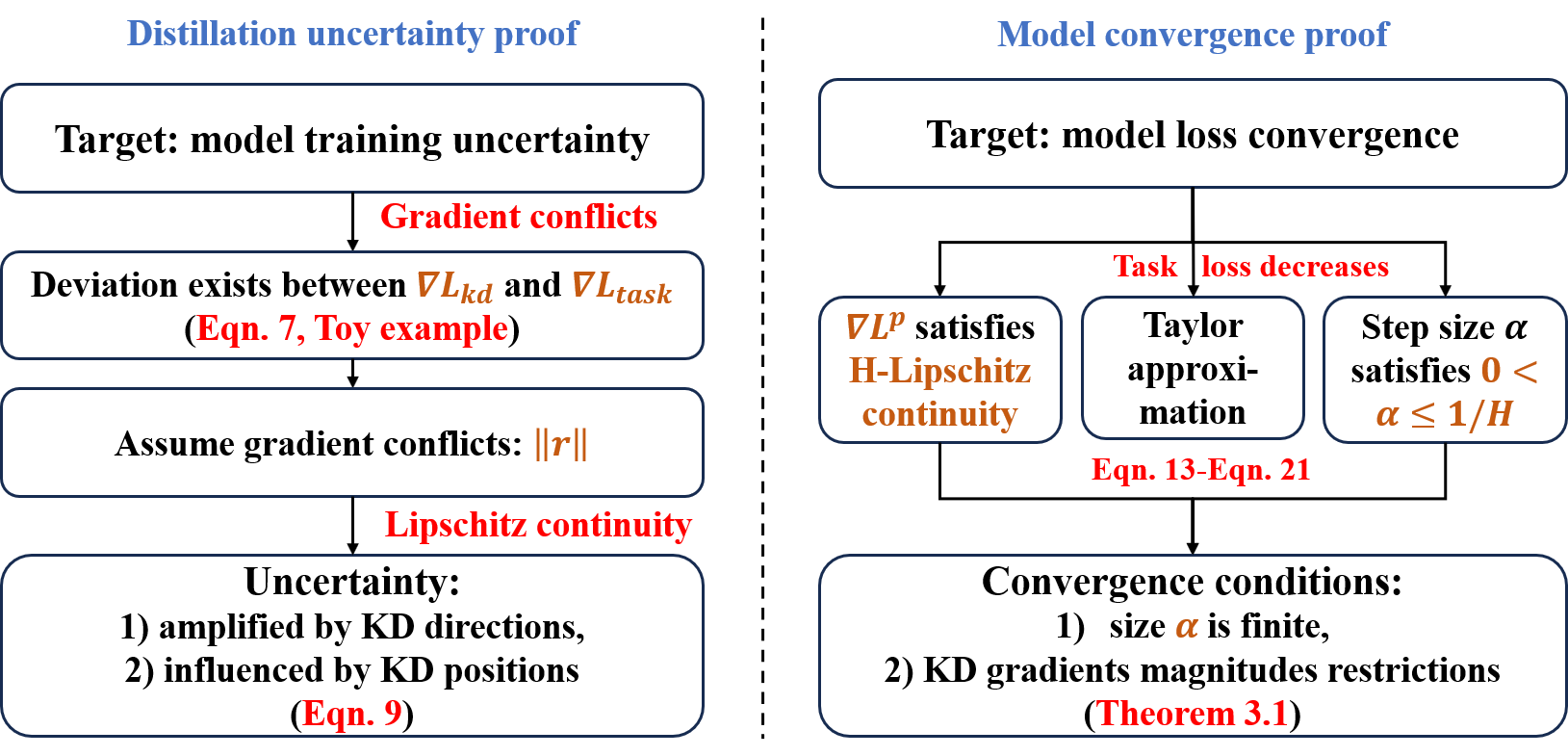}}
\caption{A flowchart illustrating the proof process of model training uncertainty and loss convergence.} 
\label{fig_flowchart}
\vspace{-6pt}
\end{figure}

The preceding analysis of the toy example indicates that distillation stability is jointly determined by the directional similarity between the distillation gradient $\nabla_{\theta^p_{Net}} \mathcal{L}_{kd}$ and the supervised learning gradient $\nabla_{\theta^p_{Net}} \mathcal{L}_{task}$, alongside the susceptibility of the model $\theta_{Net}$ to noisy gradients governed by its inherent curvature. We define an instability metric based on the product of the normalized curvature and a gradient alignment penalty:
\begin{align}
    S_{\mathrm{instability}}
    &= \sum_{x_i \in \mathcal{D}} \left[ 1 - \cos\!\left(\nabla_{\theta^p_{Net}} \mathcal{L}_{task},\; \nabla_{\theta^p_{Net}} \mathcal{L}_{kd}\right) \right] \notag\\
    &\qquad \qquad\qquad\cdot \mathcal{N}\left( \mathbb{H} (\mathcal{L};\theta^p_{Net},\theta^{p+1}_{Net}) \right),
    \label{eqn:instable}
\end{align}
where $\mathcal{D}$ is the dataset, $\mathcal{N}(\cdot)$ denotes a normalization function of the curvature $\mathbb{H} (\mathcal{L};\theta^p_{Net},\theta^{p+1}_{Net})$ of the student model updates during distillation, $\mathcal{N}(\nabla\mathcal{L}) = \nabla \mathcal{L} / (\mathcal{L}+1)$ in this work. Considering only the parameter updates on the student side, the update rule is formulated as $\theta^{p+1}_{Net}-\theta^p_{Net} = \alpha \cdot (\nabla \mathcal{L}_{task}(\theta^p_{Net}) + \nabla \mathcal{L}_{kd}(\theta^p_{Net}))$, where $\alpha$ denotes the learning rate and $\theta^p_{Net}$ represents the student parameters at iteration $p$. {Inspired by prior studies on gradient conflict~\cite{yu2020gradient}, variance~\cite{mccandlish2018empirical}, and loss landscape smoothness~\cite{santurkar2018does}, we formulate $\mathbb{H}$ as a first-order empirical proxy for local optimization fluctuations:}
\begin{equation}
    \mathbb{H} (\mathcal{L}; \theta^p_{Net},\theta^{p+1}_{Net}) \approx \left( \nabla \mathcal{L}(\theta^p_{Net})^T \nabla \mathcal{L}(\theta^{p+1}_{Net}) \right)^2,
\end{equation}
When the distillation process introduces inappropriate knowledge, the total gradient fluctuates erratically. This instability creates a substantial disparity and misalignment between $\nabla \mathcal{L}(\theta ^p)$ and $\nabla \mathcal{L}(\theta ^{p+1})$, ultimately causing violent oscillations in their inner product. {From a manifold perspective, this optimization instability fundamentally correlates with the intrinsic dimension (ID) gap between student ($d_S$) and teacher ($d_T$) feature manifolds. Based on statistical learning theory~\cite{chen2022nonparametric}, finite-sample manifold approximation error scales exponentially with intrinsic dimension. A massive ID gap ($d_S \gg d_T$) implies direct feature distillation forces a divergent high-dimensional manifold into abrupt geometric compression. This forced topological collapse maps to our instability formulations: it severely misaligns the distillation descent direction from the optimal task gradient (inducing $\cos < 0$), rendering the loss landscape highly corrugated and structurally causing a surge in local curvature $\mathbb{H}$.}

\noindent\textbf{{Theoretical proxy analysis of reducing uncertainty.}}

This proxy analysis addresses the "gradient conflict" problem where the distillation gradient acts as a noise vector ($r$) interfering with the task gradient. The uncertainty bound $U$ under perturbation $r$ can be formalized as: $U(\theta_{Net}, r) \propto \|\theta_{Net}\| \cdot \|r\|$. {Based on the aforementioned manifold approximation theory, we reasonably assume that the introduced distillation gradient noise bound $\|r\|$ acts as a monotonically increasing function positively correlated with the intrinsic dimension gap $\Delta d = |d_S - d_T|$.} It is deduced from this: $U_{1-to-1(ours)} \leq U_{1-to-n} \leq U_{m-to-n}$

The following section details how to estimate the uncertainty involved in the distillation process. 
Fig.~\ref{fig_Adapter}(a) illustrates a typical feature distillation process, where the student feature learns from the teacher feature through mapping parameters $\theta_{Map}$. The gradient of the $i^{th}$ feature $f_i^s$ is expressed by Eqn.~\ref{equ_gra}:
\begin{equation} \label{equ_gra}
{\nabla_{f_i^s}\mathcal{L} = \nabla_{f_i^s}\mathcal{L}_{kd} + \nabla_{f_i^s}\mathcal{L}_{task}.}
\end{equation}
{This formulation indicates that the gradient update comprises two distinct directions. A conflict emerges when the newly introduced distillation gradient diverges from the original supervised learning gradient $\nabla_{f_i^s}\mathcal{L}_{task}$ by an angle exceeding 90 degrees, which corresponds to a negative cosine similarity, denoted as $\cos(\nabla \mathcal{L}_{task}, \nabla \mathcal{L}_{feat}) < 0$. We characterize the additional parameter updates imposed on the features caused by this gradient conflict as optimization noise $r$.}

Lipschitz theory indicates that introducing random noise into a neural network causes parameter variations with a theoretical Lipschitz upper bound. Szegedy et al. approximate this bound using the norms of the network parameters and the noise~\cite{szegedy2014intriguing}. Specifically, denoting the supervised model parameters as $\theta_{Net}$ and a noise perturbation in feature $f_i^s$ as $r$, the upper bound of the model uncertainty is expressed as:
\begin{equation}  
\left \|   \Phi (\theta _{Net};f_i^s+r)- \Phi (\theta _{Net};f_i^s)\right \|\le \left \| \theta_{Net}  \right \|\left \|r\right \|,
\end{equation} with $\Phi(\theta_{Net};f_i^s)=max(0,\theta _{Net}\!\cdot\! f_i^s)$ represents the output of $\theta_{Net}$ with input $f_i^s$. Under our teacher-student distillation setting, the uncertainty is as follows:

\begin{equation}  
     U(\theta_{Net} , r) \propto  \| \theta_{Net}\| \cdot  \left \|r\right \|,
\end{equation}  where $\theta_{Net}$ represents the parameters associated with magnifying the incorrect knowledge in the noise propagation. {Observations indicate the relative magnitudes of the parameters associated with magnifying the distillation noise of the methods in Fig.~\ref{fig_Adapter}.}
\begin{equation}
\|\theta_{Net}^{m-to-n}\|\geq
\|\theta_{Net}^{1-to-n}\|\geq
\|\theta_{Net}^{1-to-1}\|,
\end{equation}
and the comparison of their training is as follows where our teacher-side adapter-based model has the least uncertainty:
\begin{equation}
    U_{1-to-1}\leq U_{1-to-n}\leq U_{m-to-n}.
\end{equation} 
The aforementioned relationship indicates that our 1-to-1 teacher-side adapter achieves the lowest level of uncertainty.

\noindent \textbf{Theoretical proxy analysis of accelerating convergence.}
{Under idealized smoothness assumptions, this theoretical proxy analysis serves as explanatory support, suggesting that the training process of our method converges with fewer constraints and its convergence speed is faster compared to other methods.} First, we give the sufficient condition for the loss to decrease ($\mathcal{L}(\theta^{p+1}) - \mathcal{L}(\theta^p) < 0$). Then, we define the relationship between adapter complexity $|\theta|$ and the update step size $\alpha$ as: $|\theta_{m-to-n}| > |\theta_{1-to-1 (ours)}| \implies \alpha_{m-to-n} < \alpha_{1-to-1 (ours)}$.

The goal of training deep neural networks lies in finding an optimal set of parameters $\theta^{*}$ that achieves low loss $\mathcal{L}_{task}(\theta_{Net})$ where the independent variable is $\theta_{Net}$ on the target task. This implies that the task loss must be continually minimized $\mathcal{L}_{task} (\theta_{Net}^p )\downarrow$ over time $p\uparrow$ and parameters $\theta_{Net}$ are continually optimized to $\theta^*$. 

\textit{Theorem 3.1} \textit{Based on~\cite{liu2021conflict}, assume that the gradient of the loss function satisfies H-Lipschitz continuity, i.e. $\left \| \nabla \mathcal{L}_{task}(x) -\nabla \mathcal{L}_{task}(y) \right \| \le H\left \| x-y \right \| $, where $0\le H\le \infty$. Deep neural networks $\mathcal{L}_{task}(\theta_{Net})$ can converge to a Pareto stationary point with the update step size $\alpha$ meets $0< \alpha \le \frac{1}{H}$ and the searched gradient $g$ satisfies:}
\begin{equation}
    \left \|\nabla \mathcal{L}_{task}(\theta_{Net}^{p})-g\right \|< \left \| \nabla \mathcal{L}_{task}(\theta_{Net}^{p}) \right \|,
\end{equation} where $\nabla \mathcal{L}_{task}(\theta_{Net}^p)$ represents the update gradient of $\theta_{Net}^P$ from loss $\mathcal{L}_{task}(\cdot)$ at time $p$. The proof is as follows:
\begin{align}
    & \mathcal{L}(\theta_{Net}^{p+1})-\mathcal{L}(\theta_{Net}^{p}) \notag\\
    =&\mathcal{L}(\theta_{Net}^{p}-\alpha\cdot  g)-\mathcal{L}(\theta_{Net}^{p}) \notag\\
    &{\small (Second-order\;Taylor \; approximation)} \notag\\
    \approx& - \alpha  g\cdot \nabla \mathcal{L}(\theta_{Net}^{p})+\frac{\nabla^2 \mathcal{L}(\theta_{Net}^{p})}{2}[(-\alpha\cdot  g)]^2 \notag\\
    &{\small (Lipschitz \; continuity)} \notag\\
    \le& - \alpha  g\cdot \nabla \mathcal{L}(\theta_{Net}^{p})+\frac{H}{2}[(-\alpha\cdot  g)]^2 \notag\\
    \le&- \alpha  g\cdot \nabla \mathcal{L}(\theta_{Net}^{p})+\frac{\alpha}{2}\left \| g \right \| ^2 \notag\\
    =&\frac{\alpha}{2}\left \|\nabla \mathcal{L}(\theta_{Net}^{p})-g\right \|^2-\frac{\alpha}{2}\left \| \nabla \mathcal{L}(\theta_{Net}^{p}) \right \|^2 \notag\\
    =&-\frac{\alpha}{2}\left ( \left \| \nabla \mathcal{L}(\theta_{Net}^{p}) \right \|^2- \left \|\nabla \mathcal{L}(\theta_{Net}^{p})-g\right \|^2\right ),
    \label{equ_proof}
\end{align} 
when $ \left \|\nabla \mathcal{L}_{task}(\theta_{Net}^{p})-g\right \|< \left \| \nabla \mathcal{L}_{task}(\theta_{Net}^{p}) \right \|$, Eqn.~\ref{equ_proof}  is strictly negative. Hence, we have a strictly decreasing sequence ${\mathcal{L}_{task}^p(\theta_{Net})}_{p\in T}$, $T$ represents the training period. Then loss $\mathcal{L}_{task}(\theta_{Net})$ must converge because it has a lower bound.

In a typical knowledge distillation scenario, the loss function during the training process can be expressed as: $\mathcal{L}= \mathcal{L}_{task}(\theta_{Net} )+ {\textstyle \sum_{i=1}^{R}}\mathcal{L}_{kd}^i(\theta_{Net}, \theta_{Map}^i)$, $R$ is the number of distillation losses from multi-layer teacher knowledge, which can be simplified to Eqn.~\ref{equ_loss_kd+ts} as $\left \| \theta_{Net} \right \| \gg \left \| \theta_{Map} \right \|$. 
\begin{equation}
    \mathcal{L}(\theta)= \mathcal{L}_{task}(\theta)+  \sum_{i=1}^{R}   \mathcal{L}_{kd}^i(\theta).
    \label{equ_loss_kd+ts}
\end{equation} 
$\theta$ is optimized via the update $\theta^{p+1}=\theta ^p-\alpha \cdot g$, where $\alpha$ is the update step and $g$ is the unified gradient and $g=g_{task} + {\textstyle \sum_{i=1}^{R}} g_{kd}^i$ in the knowledge distillation scenario, $g_{task}$ and $g_{kd}$ represent $\nabla \mathcal{L}_{task}(\theta)$ and $\nabla \mathcal{L}_{kd}(\theta)$ respectively. Based on \textit{Theorem3.1}, we will demonstrate that an increased number of distillation gradient directions $R$ from the teacher results in slower convergence of $\mathcal{L}(\theta)$.

Assuming that all loss function gradients ($\nabla \mathcal{L}_{task}(\theta), \{ \nabla \mathcal{L}_{kd}^i(\theta)\}_{i}^R$) satisfy $H^i$-Lipschitz continuity, i.e. $\left \| \nabla \mathcal{L}^i(x) -\nabla \mathcal{L}^i(y) \right \| \le H^i\left \| x-y \right \| $, where $0\le H\le \infty$. The largest Lipschitz constant among them denoted as $H^{max}$, and all losses updated by the same step $\alpha$. We can apply the Triangle Inequality theorem to conclude that the combined loss gradient $\nabla \mathcal{L}(\theta)$ is also Lipschitz continuous, with a Lipschitz constant equal to:
\begin{equation}
    H^*= (R+1)\cdot H^{max}.
\end{equation} The detailed derivations of this equation are as follows:
\begin{align}
&\left \| \nabla \mathcal{L}(x )-\nabla \mathcal{L}(y) \right \| \notag\\
=&\left \| \sum_{i=1}^{R}  (\nabla \mathcal{L}^i(x )-\nabla \mathcal{L}^i(y )) \right \| \notag\\
=&\left \| \nabla \mathcal{L}^1(x )-\nabla \mathcal{L}^1(y )+\cdots+\nabla \mathcal{L}^R(x )-\nabla \mathcal{L}^R(y ) \right \| \notag\\
\le &\left \| \nabla \mathcal{L}^1(x )-\nabla \mathcal{L}^1(y ) \right \| +\cdots+\left \| \nabla \mathcal{L}^R(x )-\nabla \mathcal{L}^R(y ) \right \| \notag\\
\le &(H^1+\cdots+H^R)\left \| x-y \right \|  \notag\\
\le & R\cdot H^{max}\left \| x-y \right \|.
\label{equ_derivation}
\end{align} The Lipschitz constant of $\nabla \mathcal{L}(\theta)$ is $R\cdot H^{max}$.

{Based on Eqn.~\ref{equ_proof} and Eqn.~\ref{equ_derivation}, we have Eqn.~\ref{equ_L_derivation} to show the distillation convergence constraints. 
\begin{equation}
    \left \| g-g_{task} \right \| <\left \| g_{task} \right \| ;\left \| g-g_{kd}^i \right \| <\left \| g_{kd}^i \right \|, i\in [1,R],
    \label{equ_L_derivation}
\end{equation}
Further details are in the Appendix.}

Now we discuss $m$-to-$n$ (Fig.~\ref{fig_Adapter}(b)) and $1$-to-$n$ (Fig.~\ref{fig_Adapter}(c)) adapter designs that will result in slower convergence than $1$-to-$1$ teacher-side adapter (Fig.~\ref{fig_Adapter}(d)). {The distillation gradient directions of these adapter designs satisfy $R_{m-to-n}>R_{1-to-n}>R_{1-to-1}$.}
\begin{enumerate}
    \item The convergence complexity. From \textit{Theorem3.1} and Eqn.~\ref{equ_L_derivation}, all loss functions ($\mathcal{L}_{task}(\theta), \{ \mathcal{L}_{kd}^i(\theta)\}_{i}^R$) will convergence when $\left \| g-g_{task} \right \| $ $ \le \left \| g_{task} \right \|$ and $\{\| g-g_{kd}^i \| \le  \| g_{kd}^i \| \}_i^R $. If the joint gradient $g$ fails to satisfy any of the $R+1$ inequalities, there will be convergence problems in $\mathcal{L}(\theta)$; consequently, a larger $R$ makes the overall loss function increasingly difficult to converge. Let $C$ denote the optimization complexity, leading to the following relationship:
    \begin{equation}
        C_{m-to-n}>C_{1-to-n}>C_{1-to-1}.
    \end{equation}. Our design could reduce the convergence complexity by having fewer constraints on the gradient magnitudes.
    
    \item {Regarding the update step, the derivation in Eqn.~\ref{equ_derivation} demonstrates that the convergence of the loss function $\mathcal{L}(\theta)$ is bounded by the Lipschitz constant $R\cdot H^{max}$, which restricts the allowable step size $\alpha$ such that $0 < \alpha \le \frac{1}{R\cdot H^{max}}$. {Geometrically, the forced dimensionality reduction required to bridge a massive ID gap steepens the loss landscape, which naturally inflates the Lipschitz constant $H^{max}$.} Consequently, as the number of update directions {and landscape steepness} increases, the interference among the gradients of distinct functions intensifies, which yields the following relation for the permissible step sizes:
    \begin{equation}
        \alpha _{m-to-n}<\alpha _{1-to-n}<\alpha _{1-to-1}.
    \end{equation} Our $1$-to-$1$ teacher-side adapter {could accelerate convergence} by a larger update step.}
\end{enumerate}

The above relationship suggests that our $1$-to-$1$ teacher-side adapter achieves faster convergence and reduces training complexity. 

Given the difficulty in directly quantifying model convergence stability, efficient and stable convergence is empirically manifested as convergence speed, training loss and generalization capability (test performance). While we establish a conceptual connection between ID reduction and optimization difficulty, forging a precise closed-form mathematical binding remains challenging. {Because our mathematical framework relies on idealized assumptions, including perfect smoothness and simplified gradient aggregation, these derivations serve strictly as explanatory support for our empirical behavior rather than a formal convergence proof.} Consequently, we focus on validating these observable metrics and the instability from Section~\ref{sec:instability} in the subsequent experiments.

\begin{table*}[t]
\centering
\renewcommand{\arraystretch}{1.3}
\begin{tabular}{c|cc|ccccc|cc}
\toprule
    \multicolumn{1}{c|}{\multirow{2}{*}{}} & \multirow{2}{*}{\textbf{Model}} & \textbf{Distillation} & \multicolumn{5}{c|}{\textbf{\!Large gap($T_{25\%}$): Student$=$25\% Teacher}} & \multicolumn{2}{c}{\textbf{Huge gap($T_{6.3\%}$)}} \\
    \multicolumn{1}{c|}{} &  & \textbf{layers} & Train time & mIoU & Pixel-acc & \!\! QoS$^{0.5}$\!\!\!\! & \!\!\!\! QoS$^{0.2}$\!\!\! & mIoU & Pixel-acc\\
    \hline
    No KD & \multicolumn{1}{c}{STL-t} & \multicolumn{1}{c|}{-}  & 3.3h & 59.07 & 92.95 & - & - & - & -\\
    \hline

    \multirow{10}{*}{} & \multicolumn{1}{c}{STL-p} & \multicolumn{1}{c|}{-} & 2.4h (5.4h) & 57.99$\pm$0.14 (59.0) & 92.85$\pm$0.10 (92.9)
     & {0.63} & {0.41} & 54.42$\pm$0.40 & 91.67$\pm$0.13\\
     & \multicolumn{1}{c}{MKD} & 1-layer & 3.0h (5.4h) & 57.84$\pm$0.21 (58.8) & 92.89$\pm$0.06 (93.0) & {0.51} & {0.36} & 54.40$\pm$0.23 & 91.77$\pm$0.10 \\
     & ESKD & 1-layer & 2.8h (5.2h)  & \underline{58.53$\pm$0.19} (59.1) & \underline{93.05$\pm$0.01} (93.3) & {0.61} & {0.47} & 54.62$\pm$0.24 & 91.75$\pm$0.12\\ 
    Single- & MGD & 1-layer & 2.9h (5.7h) & 58.11$\pm$0.08 (59.0) & 92.89$\pm$0.06 (93.1) & {0.54} & {0.39} & 54.56$\pm$0.42 & 91.77$\pm$0.10\\
    \cline{2-10}
    
    stage KD & SemCKD & N-layers & 3.8h (9.0h) & 58.24$\pm$0.10 (59.2) & 92.86$\pm$0.07 (93.3) & {0.42} & {0.34} & 54.53$\pm$0.47 & 91.76$\pm$0.05\\
     & KR & N-layers & 3.2h (6.4h) & 56.76$\pm$0.46 (58.0) & 92.62$\pm$0.03 (92.8) & {0.38} & {0.18} & 53.75$\pm$0.17 & 91.67$\pm$0.12\\
    \cline{2-10}
    
     & VanillaKD & soft label & 2.4h (5.4h) & 58.09$\pm$0.05 (59.3) & 92.94$\pm$0.09 (93.3) & {0.65} & {0.44} & 54.57$\pm$0.47 & 91.83$\pm$0.02\\
     & BAM & soft label & 2.4h (5.3h) & 58.17$\pm$0.13 (59.2) & 92.96$\pm$0.08 (93.2) & {0.66} & {0.45} & 54.89$\pm$0.23 & 91.88$\pm$0.05\\
     & \multicolumn{1}{c}{DIST} & soft label & 2.4h (5.6h) & 58.37$\pm$0.09 (59.0) & 92.98$\pm$0.05(93.3) & \textbf{0.67} & {0.47} & \underline{55.53$\pm$0.08} & \underline{91.99$\pm$0.13}\\
    \hline
    
    Multi- & IterDE & soft label & 15.1h & 58.67$\pm$0.09 & 93.02$\pm$0.07 & {0.19} & {0.31} & 55.53$\pm$0.13 & 92.12$\pm$0.09\\
    stage KD  &TAKD+DIST & soft label & 15.5h & 59.60$\pm$0.11 & 93.32$\pm$0.08 & {0.29} & {0.47} & 55.68$\pm$0.14 & 91.91$\pm$0.11\\
    \hline
    
    \multirow{2}{*}{Ours:} & S-stage:$Prog^2$-$t$ & 1-N layers & 3.3h & \textbf{59.73$\pm$0.16} & \textbf{93.36$\pm$0.06} & {\textbf{0.67}} & {\textbf{0.66}} & \textbf{55.86$\pm$0.18} & \textbf{92.15$\pm$0.06}\\
    & M-stage:$Prog^2$-$ts$ & 1-N layers & 12.2h & \textbf{61.72$\pm$0.05} & \textbf{93.82$\pm$0.04} & {0.53} & {\textbf{0.81}} & \textbf{58.39$\pm$0.08} & \textbf{92.88$\pm$0.05}\\
\bottomrule
\end{tabular}
\caption{Performance on Cityscapes on two different teacher-student gap settings (\%). The $Prog^2$-$t$ (single-stage) is our progressively stronger teacher-based
distillation framework and $Prog^2$-$ts$ (multi-stage) is the progressively smaller student co-evolved distillation enhancement. 1 and N layers represent single-layer and multi-layer distillation methods. The data within parentheses '$( )$' represents the time spent and the achieved performance when distilling the single-stage method over the same number of epochs as the two-stage method. The numerals underscored '$\underline{58.53}$' indicate the suboptimal outcomes of the single-stage approach. The parameter size of the full-size teacher model is 24.98M. The two students are obtained by channel compression (pruning 50\% and 75\% shared-blocks channels) from the full-size model, which reduces their parameter size to 25\%(6.25M, $T_{25\%}$) and 6.3\%(1.57M,$T_{6.3\%}$) of the teacher model's. 'QoS$^{0.5}$' represents $\alpha=0.5$ in Eqn.~\ref{equ_qos}.}
\vspace{-15pt}
\label{Tbl1:cityscape-all}
\end{table*}
\section{Experiments}
\label{sec:exp}  
 
\vspace{3pt}\noindent\textbf{Dataset} We evaluate our method on semantic segmentation, depth estimation, and image classification tasks. For segmentation, the benchmark dataset \textbf{Cityscapes}~\cite{cordts2016cityscapes} is used which covers over 50 cities' street scenes with 2975 images for training and 500 for testing. For classification, we used \textbf{{CIFAR-100}}~\cite{krizhevsky2009learning}, which comprises 100 classes containing 600 images each, split into 500 training images and 100 testing images per class. \textbf{{Tiny-ImageNet}}~\cite{deng2009imagenet} is a subset of the larger ImageNet collection and includes 200 classes. Each class consists of 500 training images, 50 validation images, and 50 test images. The images are downscaled to 64x64 pixels, providing a more computationally manageable yet challenging dataset than the full ImageNet dataset. For the depth estimation task, we selected the \textbf{NYU-V2 dataset}~\cite{silberman2012indoor}, which consists of 3 cities, 464 scenes, and 1449 images (654 images for testing). All datasets underwent random cropping and flipping to accomplish image augmentation. We follow prior work {MTAN~\cite{liu2019end}, InvPT~\cite{ye2022inverted}, SemCKD~\cite{chen2021cross}} for preprocessing these datasets to ensure fairness. 
 
{\vspace{3pt}\noindent\textbf {Implementation} On the Cityscapes dataset, we used Adam optimization with a learning rate of 1e-4 for the student, teacher (for our $Prog^2$-$ts$ setting), and the teacher-side adapter. The single-layer convolution for teacher-student alignment in terms of channel dimension follows prior work\cite{li2020knowledge} with a learning rate of 0.1. We train the model with 200 epochs (600 epochs for multi-stage setting), with no learning rate decay, the batch size is set at 8. For {CIFAR-100}, we used SGD optimization with a learning rate of 0.05 for the student and the teacher side adapter, with the Nesterov momentum of 0.9 for 240 epochs and a learning rate decay of 0.1 for 150, 180, and 210 epochs (followed by~\cite{tiancontrastive}), and with the batch size of 128. More setting details on other datasets are in the Appendix.}

\vspace{3pt}\noindent\textbf{Mobile platforms} {For on-device evaluation, we utilize high-end edge devices: Raspberry Pi 4 Model B and the Jetson Nano P3450, as detailed in Fig.~\ref{fig_devices}.}
 
\vspace{3pt}\noindent\textbf{Baselines}
We compare with three types of method: 1) the original teacher without distillation (namely \underline{no-KD}): a full-sized teacher (STL) which adopts SegNet~\cite{badrinarayanan2017segnet} for segmentation task or VGG-16~\cite{simonyan2014very} and ResNet~\cite{he2016deep} for classification task. 2) The classic distillation methods whose training is non-iterative (namely \underline{single-stage KD}) including \textbf{MKD}~\cite{li2020knowledge}, 
\textbf{BAM}~\cite{clark2019bam}, \textbf{ESKD}~\cite{cho2019efficacy}, \textbf{VanillaKD}~\cite{hinton2015distilling}, \textbf{DKD}~\cite{zhao2022decoupled} for classification task only, ~\textbf{DIST}~\cite{huang2022knowledge} and recent strong methods with multi-layer distillation design including~\textbf{SemCKD}~\cite{chen2021cross}, \textbf{KR}~\cite{chen2021distilling}, \textbf{MGD}~\cite{yang2022masked}, and the classic distillation methods with progressive distillation settings (namely \underline{multi-stage KD}) including \textbf{IterDE}~\cite{liu2023iterde} and \textbf{TAKD}~\cite{mirzadeh2020improved}+\textbf{DIST}~\cite{huang2022knowledge}.

\vspace{3pt}\noindent\textbf{Metrics} In alignment with established studies~\cite{liu2019end,chen2021distilling,ye2022inverted}, we employ widely recognized evaluation metrics in our research. We use the mean Intersection over Union (mIoU) and pixel accuracy for semantic segmentation. For depth estimation, we measure absolute error and relative error. Additionally, we determine performance for classification tasks using top-1 and top-5 accuracy metrics. 

{QoS is the service's consistent capability to be effective, available, and reliable. Following the fundamental definitions outlined by~\cite{zhang2023service}, we characterize the QoS value of KD from server to client based on metrics of time and performance.
\begin{equation}
    \mathrm{QoS}(KD) = \alpha \cdot \mathcal{N}(T^{-1}) + (1-\alpha) \cdot \overline{\mathcal{N}(\mathrm{Acc})},
    \label{equ_qos}
\end{equation} where $\mathcal{N}(\cdot)$ denotes the Min-Max normalization over all evaluated methods, $\overline{\mathcal{N}(\mathrm{Acc})}$ denotes the average of all performance metrics.} $\alpha$ signifies individual user preferences for QoS and can assume any value within the specified range $\left [ 0,1 \right ] $.

\begin{figure}[!t]
\centerline{\includegraphics[width=3.5in]{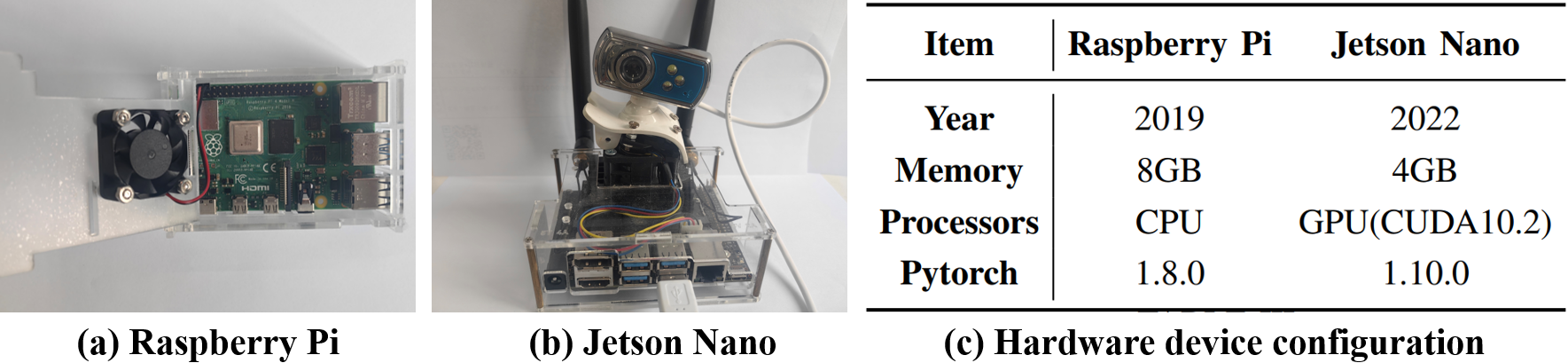}}
\vspace{-3mm}
\caption{Figure of hardware devices.} 
\label{fig_devices}
\end{figure}

\subsection{Quantitative performance comparison} 
\noindent\textbf{Cityscapes} In Table~\ref{Tbl1:cityscape-all}, our method outperforms baselines in both single-stage and multi-stage settings. Performance-wise, $Prog^2$-$ts$ outperforms $Prog^2$-$t$, as the progressively smaller student ensures the teacher-student gap is not excessively large at each stage. However, this improvement incurs a higher training cost, a common challenge for iterative methods. Despite this, our training time remains considerably shorter than other multi-stage baselines. Given that training is conducted offline and the inference model remains identical, this cost is not particularly sensitive for most applications.


In a single-stage, large teacher-student gap setting, $Prog^2$-$t$ outperforms suboptimal ESKD~\cite{cho2019efficacy} by nearly 1.2\% in mIoU and 0.3\% in pixel accuracy within a limited time frame. Similarly, under a huge teacher-student gap, our approach outperforms suboptimal DIST~\cite{huang2022knowledge} with nearly a 0.4\% advantage in mIoU and a 0.06\% advantage in pixel accuracy, demonstrating effectiveness in substantial gap scenarios. Conventional approaches primarily focus on students around 25\% of the teacher size (Resnet-18 and Resnet-101)~\cite{zhao2022decoupled,yang2022cross,huang2022knowledge}, similar to our large gap setting.

Overall, traditional methods statically utilizing multi-teacher layers (SemCKD~\cite{chen2021cross}, KR~\cite{chen2021distilling}) may perform poorly due to insufficient learning in early training. As the teacher-student gap increases (6.3\% teacher), recent strong methods become ineffective (MKD~\cite{li2020knowledge}, ESKD~\cite{cho2019efficacy}, VanillaKD~\cite{hinton2015distilling}). In such cases, our multi-stage method $Prog^2$-$ts$ demonstrates clear improvements by transferring knowledge more finely and stably. Our $Prog^2$-$ts$ method outperforms suboptimal TAKD+DIST~\cite{huang2022knowledge}, exhibiting improvements of 2.6\% and 2.8\% in mIoU and 0.7\% and 1.0\% in pixel accuracy across two student sizes, respectively. Furthermore, our method achieves the most stable results by minimizing deviation across experimental outcomes.


\begin{table}[!t]
\begin{center}
\renewcommand{\arraystretch}{1.3}
\begin{tabular}{cc|cc}
\toprule
{\textbf{ Methods }} & {\textbf{Distill layers}} & {\textbf{Top-1 Acc}} & {\textbf{Top-5 Acc}}\\
\cline{1-4}
{Full Teacher} & {-} & {74.10} & {91.87}\\
\cline{1-4}
{STL-p} & {-} & {69.94 $\pm$ 0.08} & {89.73 $\pm$ 0.15}\\
{MGD} & {single-layer} & {70.98 $\pm$ 0.17} & {90.67 $\pm$ 0.14}\\
{ESKD} & {single-layer} & {71.95 $\pm$ 0.31} & {91.19 $\pm$ 0.03}\\
{SemCKD} & {multi-layer} & {72.30 $\pm$ 0.05} & {91.46 $\pm$ 0.01}\\
{KR} & {multi-layer} & {\underline{72.49 $\pm$ 0.10}} & {\underline{91.60 $\pm$ 0.11}}\\
{DIST} & {logits (soft label)} & {70.78 $\pm$ 0.00} & {90.39 $\pm$ 0.01}\\
{DKD} & {logits (soft label)} & {72.19 $\pm$ 0.04} & {91.37 $\pm$ 0.01}\\
{Ours: $Prog^2$-$t$} & {single to multi-layer} & {\textbf{73.49 $\pm$ 0.05}} & {\textbf{91.76 $\pm$ 0.07}}\\
\cline{1-4}
{TAKD + DIST} & {logits (soft label)} & {72.75 $\pm$ 0.10} & {91.69 $\pm$ 0.06}\\
{Ours: $Prog^2$-$ts$} & {single to multi-layer} & {\textbf{74.44} $\pm$ 0.03} & {\textbf{92.05} $\pm$ 0.03}\\
\bottomrule
\end{tabular}
\caption{{Classification Testing Result on {CIFAR-100} Dataset (\%). The parameter sizes of the teacher and student models are 14.78M and 3.71M of the VGG backbone, respectively.}} 
\label{tab_cifar}
\end{center}
\end{table}

\begin{figure}[!t]
    \vspace{-5mm}
    \centerline{\includegraphics[width=3.3in]{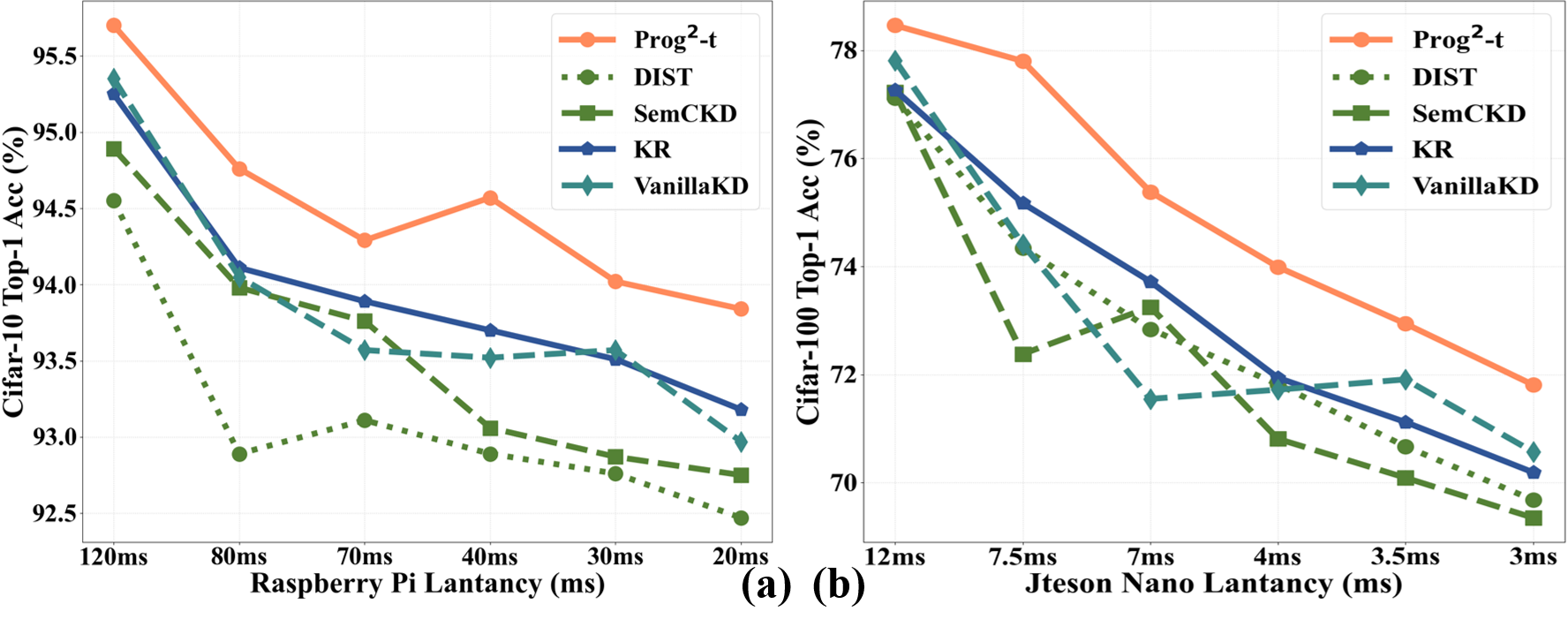}}
    \vspace{-3mm}
    \caption{{Performance evaluation on mobile devices. Six distinct ResNet model sizes are used on the {CIFAR-10 and CIFAR-100} dataset.}}
    \label{fig:device_results}   
\end{figure}

\vspace{3pt}\noindent\textbf{{CIFAR-100}} In this experiment (Table~\ref{tab_cifar}), the parameter sizes of the teacher and student models are 14.78M and 3.71M, respectively. In addition to the KR~\cite{chen2021distilling} method in specific generalization directions in classification tasks, our $Prog^2$-$t$ approach surpasses all single-stage baseline models. The experimental results on the {CIFAR-100} dataset once again demonstrate the effectiveness of progressive$^2$ distillation in compression scenarios. Our method $Prog^2$-$ts$ surpasses the suboptimal baseline TAKD~\cite{mirzadeh2020improved}+DIST~\cite{huang2022knowledge} by 1.8\% in the top-1 accuracy.

{More results on depth estimation task on the NYU-V2 dataset and BERT-based models on the GLUE benchmark are shown in the Appendix.}

\begin{table}[!t]
    \centering
    \renewcommand{\arraystretch}{1.3}
    \begin{tabular}{c|cc|ccc}
    \toprule
         {\textbf{Inference}} & {\multirow{3}{*}{\textbf{Method}}} & {\textbf{Training}}  & {\textbf{Training}} & {\textbf{Per-iter}} & {\textbf{GPU}}\\
          {\textbf{Model}} & & {\textbf{Model}} & {\textbf{Time}} & {\textbf{FLOPs}} & {\textbf{Memory}}\\
         {\textbf{Size (M)}} & & {\textbf{Size (M)}} & {\textbf{(h)}} & {\textbf{(G)}} & {\textbf{(GiB)}}\\
         \hline
         {\multirow{5}{*}{6.25}} & {MKD} & {9.00} & {2.87} & {532.09} & {3.08}\\
         &{KR} & {9.84} & {3.23} & {618.49} & {3.81}\\
         &{SemCKD} & {33.63} & {3.80} & {1022.54} & {4.75}\\
         &{$Prog^2$-$t_{128}$} & {6.34} & {2.96} & {564.28} & {3.62}\\
         &{$Prog^2$-$t$} & {6.53} & {3.48} & {743.25} & {4.25}\\
    \bottomrule
    \end{tabular}
    \caption{{Results of floating point operations (FLOPs) and memory footprint during knowledge distillation training. $Prog^2$-$t_{128}$ represents a smaller number of average channels in the teacher-side multi-layer adapter.}}
    \label{res_tab:flops}
    \vspace{-6mm}
\end{table}


{\subsection{Analysis of computational Efficiency and Mobile Deployment}}

To explicitly evaluate the computational overhead, we measured the Floating Point Operations (FLOPs) per image (using $128 \times 128$ resolution from the Cityscapes dataset) and the peak GPU memory footprint during training in Table~\ref{res_tab:flops}. As established in Section 3, our comparative analysis focuses specifically on multi-layer feature distillation methods (MKD~\cite{li2020knowledge}, KR~\cite{chen2021distilling}, SemCKD~\cite{wang2021knowledge}) to ensure alignment with our theoretical insights. The results indicate that by selecting a reduced number of average channels (denoted as $Prog^2$-$t_{128}$), our method achieves a distinct advantage in computational complexity over state-of-the-art multi-layer feature distillation methods (e.g., KR~\cite{chen2021distilling}, SemCKD~\cite{wang2021knowledge}). This change makes a marginal performance sacrifice (mIoU: $59.5 \to 58.8$). A detailed demonstration of the associated training and convergence stability is presented in the subsequent section.

To fairly simulate mobile constraints, we evaluate six ResNet architectures (0.47M\-7.43M parameters) trained under a standardized server-side GPU budget, abstracting away communication overheads. The models were deployed on Jetson Nano (on CIFAR-100) and Raspberry Pi (on CIFAR-10) to assess inference latency and accuracy. As shown in Fig.~\ref{fig:device_results}, $Prog^2$-$t$ demonstrates the most stable knowledge transfer and yields superior accuracy compared to competing methods under identical resource constraints.


\begin{table}[!t]
\begin{center}
\renewcommand{\arraystretch}{1.15}
\begin{tabular}{cccc|cc}
\toprule
\multicolumn{4}{c|}{{\textbf{Ablation settings}}} & \multicolumn{2}{c}{{\textbf{Performance}}}\\
{\textbf{P-ss}} & {\textbf{P-st}} & {\textbf{T-loc}} & {\textbf{\#TL}} & {\textbf{mIoU(\%)}} & {\textbf{Pixel-acc(\%)}}\\
\cline{1-6}
{--} & {--} & {Enc} & {5} & {58.76} & {93.37} \\
{--} & {--} & {Dec} & {5} & {58.46} & {93.04} \\
{--} & {--} & {Enc} & {$5 \rightarrow 1$} & {59.09} & {93.07} \\
{--} & {--} & {Dec} & {$5 \rightarrow 1$} & {58.58} & {93.04} \\
{--} & {\ding{51}} & {Enc} & {$1 \rightarrow 5$} & {\textbf{59.69}} & {\textbf{93.36}} \\
{--} & {\ding{51}} & {Enc} & {$1 \rightarrow 5$ (R)} & {58.95} & {93.08} \\
{--} & {\ding{51}} & {Dec} & {$1 \rightarrow 5$} & {58.81} & {93.01} \\
{--} & {\ding{51}} & {Dec} & {$1 \rightarrow 5$ (R)} & {\textbf{59.71}} & {\textbf{93.36}} \\
{\ding{51}} & {--} & {Enc} & {--} & {59.78} & {93.26} \\
{\ding{51}} & {--} & {Enc} & {5} & {60.13} & {93.29} \\
{\ding{51}} & {--} & {Enc} & {$5 \rightarrow 1$} & {60.22} & {93.34} \\
{\ding{51}} & {\ding{51}} & {Dec} & {$1 \rightarrow 5$} & {61.00} & {93.60} \\
{\ding{51}} & {\ding{51}} & {Dec} & {$1 \rightarrow 5$ (R)} & {\textbf{62.00}} & {\textbf{93.83}} \\
{\ding{51}} & {\ding{51}} & {Enc} & {$1 \rightarrow 5$} & {\textbf{61.77}} & {\textbf{93.80}} \\
{\ding{51}} & {\ding{51}} & {Enc} & {$1 \rightarrow 5$ (R)} & {61.43} & {93.77} \\
\bottomrule
\end{tabular}
\caption{{Ablation of progressively smaller students (\textbf{P-ss}), progressively stronger teacher (\textbf{P-st}), and the number of progressively (or statically) involved teacher layers (\textbf{\#TL}) on the Cityscapes dataset. \textbf{T-loc} represents whether use SegNet teacher's encoder or decoder knowledge. $1 \rightarrow 5$ (R) means distillation from {back-to-front} in the feature locations. The student and teacher models are the same Table \uppercase\expandafter{\romannumeral1}.}}
\label{tab_ablation}
\vspace{-2mm}
\end{center}
\end{table}

\begin{table}[!t]
\begin{center}
\renewcommand{\arraystretch}{1.3}
\setlength{\tabcolsep}{1.3mm}{
\begin{tabular}{c|ccc|cc}
\toprule
\textbf{Method}& \textbf{Setting} & \textbf{mIoU(\%)} & \textbf{Pixel-acc(\%)} & \textbf{Train Loss} & \textbf{Time}\\
\cline{1-6}
\multirow{3}{*}{$prog^2$-$t$} & {F-ml} & 59.33$\pm$0.24 & 93.20$\pm$0.07 & 0.7910 & 3.7h\\
{} & {F-sf} & 59.29$\pm$0.22 & 93.24$\pm$0.11 & 0.8019 & 4.0h\\
{} & F-sl & \textbf{59.53$\pm$0.16} & \textbf{93.30$\pm$0.06} & 0.7797 & 3.5h\\
\bottomrule
\end{tabular}}
\caption{{Ablation of single-layer student feature (\textbf{F-sl}), multi-layer student features (\textbf{F-ml}), and single-layer student fusion feature which fuses multi-layer features (\textbf{F-sf}) on the Cityscapes dataset. The student and teacher models are the same as Tabel~\ref{Tbl1:cityscape-all}.}} 
\label{tab_abl_uncertainty}
\vspace{-5mm}
\end{center}
\end{table}

\subsection{Ablation study and discussion of progressive settings} 
{To explicitly align empirical results with the proposed theoretical mechanisms, we map ablation settings to our three core contributions: 1) Intrinsic Dimension Control and 2) Progressive Scheduling are evaluated via the P-st and P-ss modules (Fig~\ref{fig:crossKD}, Table~\ref{tab_ablation}, more details are in the Appendix), demonstrating that progressive feature dimension manipulation mitigates the capacity gap. 3) Table \ref{tab_abl_uncertainty} isolates the adapter architecture necessity, multi-layer feature fusion on the student side (F-sf) yields inferior performance compared to both single-layer distillation (F-sl) and independent multi-layer distillation  (F-ml), demonstrating that positioning the adapter on the teacher side is the optimal configuration. Although additional adapter capacity may implicitly provide minor regularization, model inference during deployment remains unaffected (Fig.~\ref{fig:device_results}), and progressive topological matching primarily drives the significant performance leaps.}

\noindent\textbf{Ablation study of different progressive settings} We conduct experiments of different combinations of the teacher and the student. For the number of involved teacher layers, $a\rightarrow b$ means the number of the included layer is gradually expanded from $a$ to $b$ and a single number indicates it is a static layer involvement. The results in Table~\ref{tab_ablation} show clearly that the progressively evolved student model and enhanced teacher model achieve better results. In SegNet-based encoder-decoder network, the depth-semantic monotonicity lacks strict. Results of $a$ to $b$ (R) demonstrate that the proposed method remains effective by adopting a {back-to-front} reverse progressive strategy even when semantics are diluted with depth.

\noindent\textbf{Discussion of progressive time} We now examine the accuracy variations and training latency across different progressive stages of the multi-stage model in Table~\ref{tab_multi_stage}. The "1 stage" configuration denotes the exclusive deployment of the progressively enhanced teacher model, corresponding to our $Prog^2$-$t$ strategy. In contrast, "5 stages" iteratively compress the student model parameters to 6.25M (progressively distilled student), aligning with the $Prog^2$-$ts$ framework. Here, each time window $T_i$ progressively incorporates a larger subset of teacher layers $L_i^t$. The results demonstrate that increasing the number of progressive stages improves knowledge transfer robustness but imposes substantial temporal overhead. Users may thus calibrate stage counts based on application-specific latency-accuracy tradeoffs during deployment.

\noindent\textbf{Effect of the position of teacher-side adapter} To evaluate the impact of varying student model mapping parameters, and the number and position of layers involved in the distillation process as mentioned in Sec~\ref{sec:tea_adapter}. In Table~\ref{tab_abl_uncertainty}, we compared the outcomes, training loss, and training time of the knowledge distillation process in the $Prog^2$-$t$ experimental setup by evaluating single-layer student feature, multi-layer student features, and single-layer student fused feature with the adapter module (Fig.~\ref{fig_FlowFPN}) integrated across multiple layers. The results demonstrate that our method outperforms the multi-layer student feature distillation algorithms in terms of training time, training loss, standard deviations, and testing results. This underscores the critical importance of reducing uncertainty in the distillation process.

\begin{figure}[!t]
\vspace{-1mm}
\centerline{\includegraphics[width=3.1in]{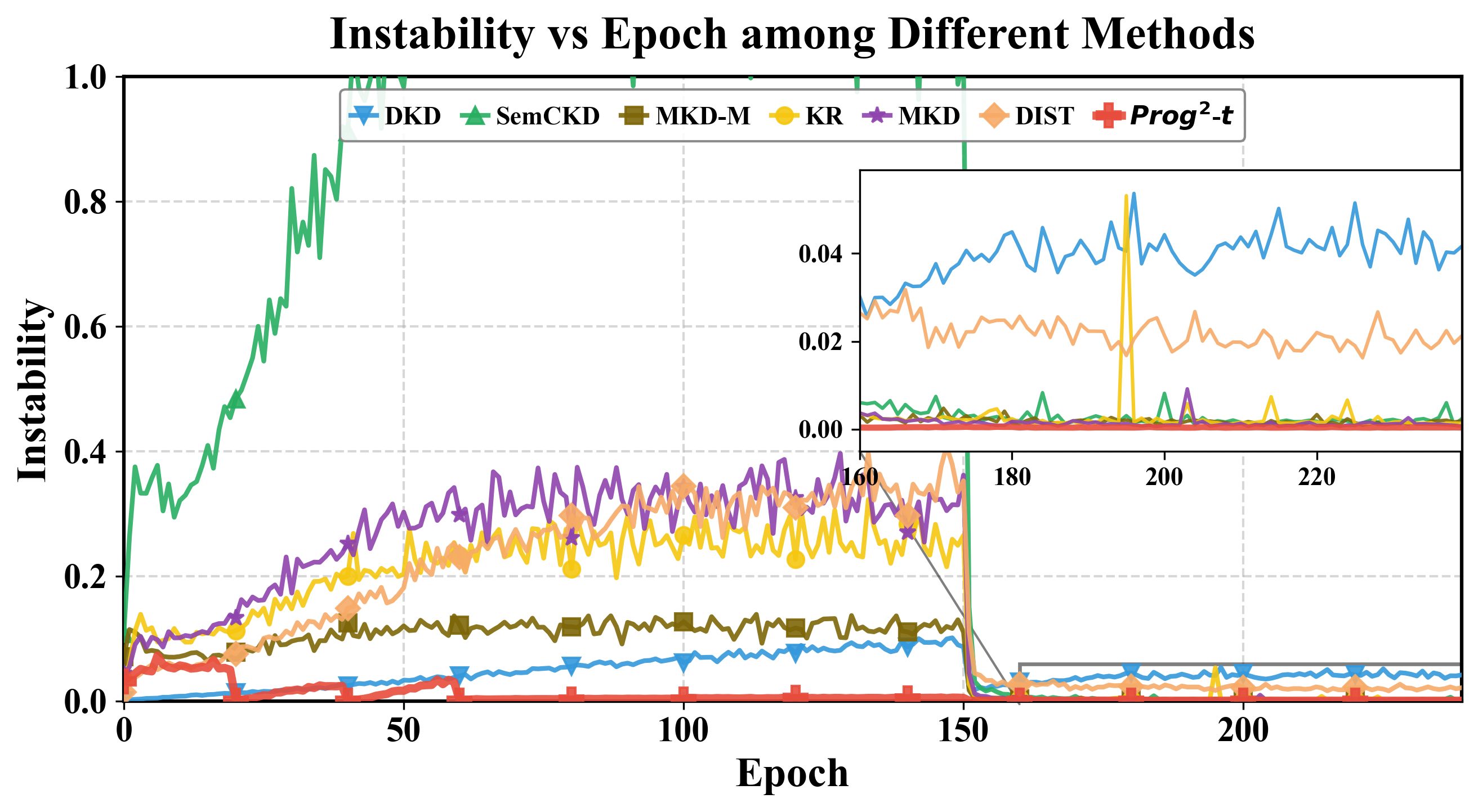}}
\vspace{-4mm}
\caption{{Training instability on the CIFAR-100 dataset under an experimental setup consistent with Table~\ref{tab_cifar}, where the instability is quantified using the previously formulated metric~\ref{eqn:instable}. A learning rate decay of 0.1 was applied at the $150_{th}$ round to ensure convergence. 'MKD-M' represents distillation more student layers than 'MKD'.}}
\label{fig_instability}
\vspace{-2mm}
\end{figure}

\begin{table*}[!t]
    \centering
    \renewcommand{\arraystretch}{1.3}
    \begin{tabular}{cc|ccccc}
    \toprule
         \textbf{Progress Stages Number} & \textbf{Training Time} & \textbf{mIoU (24.98M)} & \textbf{mIoU (19.13M)} & \textbf{mIoU (14.06M)} & \textbf{mIoU (9.76M)} & \textbf{mIoU (6.25M)}\\
         \hline
         5 stages & 12.22h & 60.70 & 61.65 & 61.98 & 61.88 & 61.72\\
         3 stages & 8.34h & 60.69 & -& 61.41 & -& 60.96\\
         1 stage & 3.98h & -& -& -& -& 59.66\\
    \bottomrule
    \end{tabular}
    \caption{Results of different stage numbers within the $Prog^2$-$ts$ method.}
    \label{tab_multi_stage}
    \vspace{-5mm}
\end{table*}

\begin{figure}[!t]
\centerline{\includegraphics[width=3.2in]{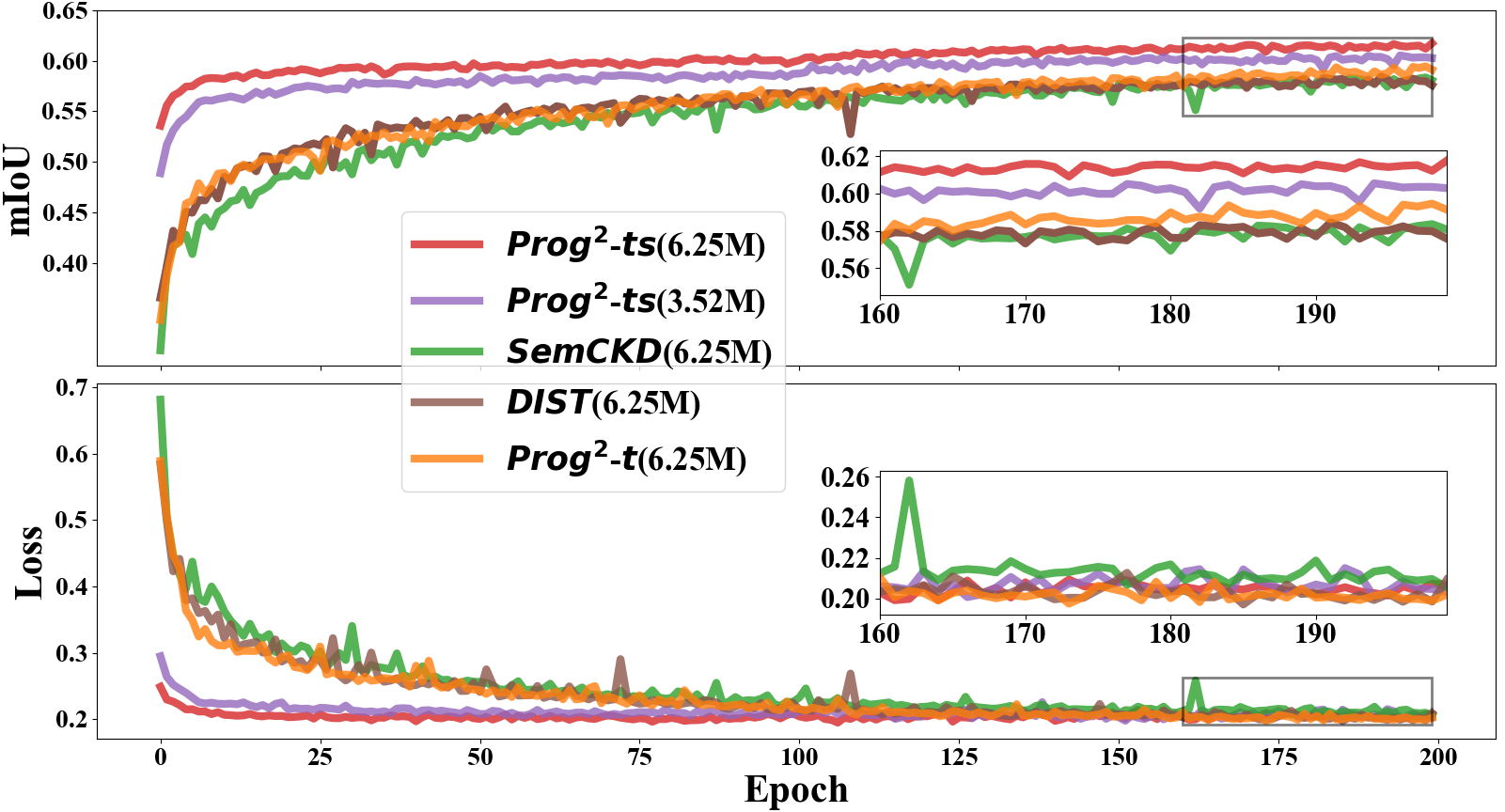}} 
\vspace{-3mm}
\caption{{The training mIoU and loss curves for models of different sizes during the training process in the $Prog^2$-$ts$, and two single-stage models: $Prog^2$-$t$ and SemCKD. All models were trained for 200 epochs.}} 
\label{fig_progressive_mIoU}
\end{figure}


\vspace{-6pt}

\subsection{Discussion of distillation instability and convergence speed}
To address the concern regarding our internally constructed metric ($S_{instability}$), we benchmarked it against an established optimization diagnostic: Gradient Norm Variance~\cite{faghri2020study,mccandlish2018empirical} (a standard measure of optimization landscape stability). As illustrated in Appendix Fig. 8 and Fig. 9, our $S_{instability}$ exhibits a positive correlation with the gradient variance across epochs. {This empirical alignment suppports that $S_{instability}$ is broadly consistent with established optimization dynamics. Consequently, this positive correlation empirically supports the use of our mathematically derived metric as an observational proxy for convergence landscape corrugation.}


\noindent{\textbf{Distillation Instability} The corresponding Fig.~\ref{fig_instability} visualizes the magnitude of instability throughout the training process of the single-stage model. The results demonstrate that multi-layer feature distillation (MKD, MKD-M, KR, SemCKD)inherently amplifies the optimization uncertainty during training. Conversely, by modifying the intrinsic dimensions of the teacher features (Fig.~\ref{fig:crossKD}) to systematically adjust the learning difficulty, our method consistently maintains a low level of instability. This empirical observation strictly aligns with the conclusions derived from our theoretical proxy analysis based on Lipschitz continuity.}

\begin{figure}[!t]
\centerline{\includegraphics[width=3.1in]{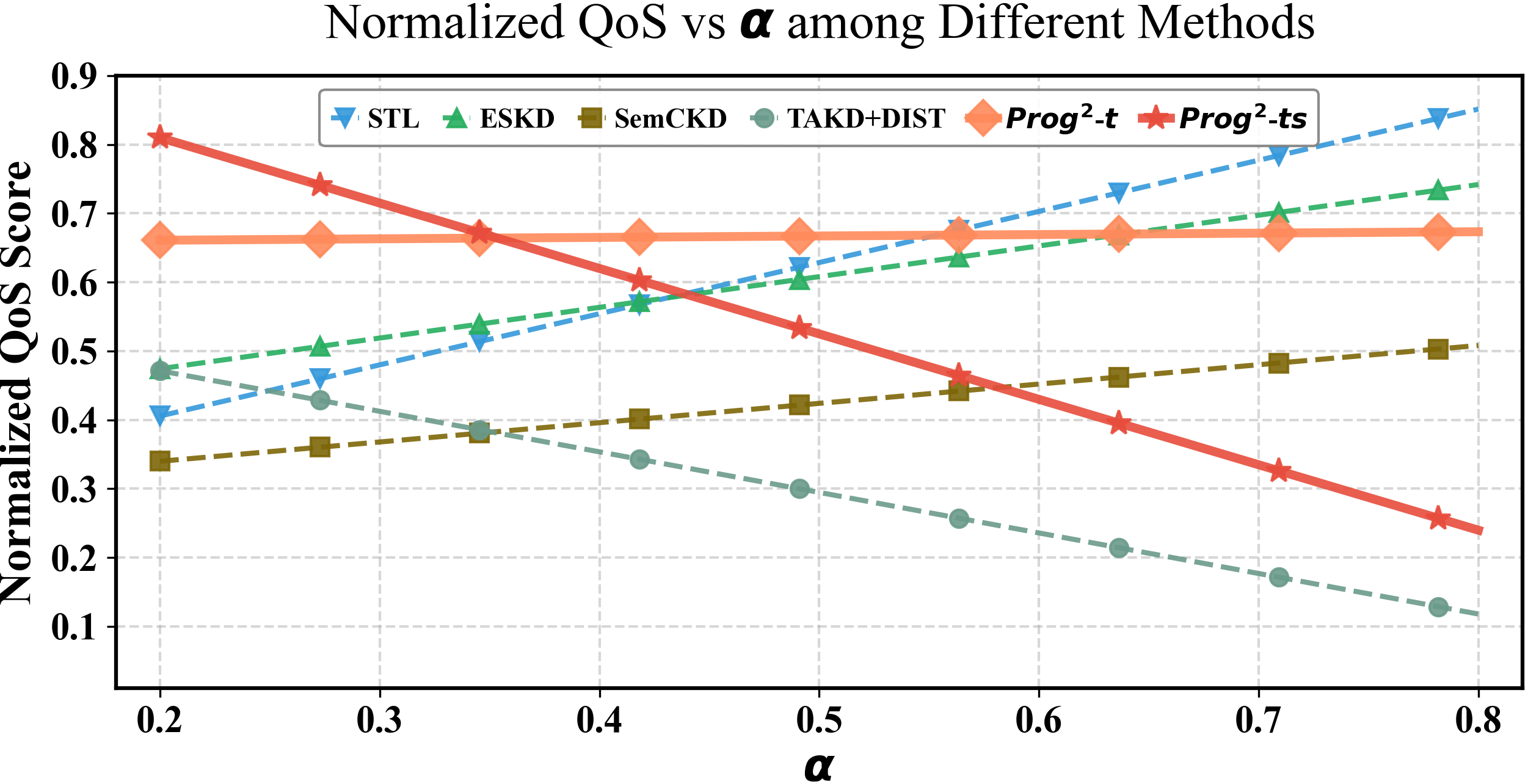}}
\vspace{-6pt}
\caption{{Quality-of-Services values with different user preferences of knowledge distillation from server to client based on time and performance metrics. Results on the segmentation task on Cityscapes dataset with large gap teacher-student pairs.}}
\label{fig_qos}
\vspace{-3mm}
\end{figure}

\noindent\textbf{Distillation Stability and convergence Speed} To compare the training stability, we use the results on Cityscapes as an illustration and plot the mIoU and loss curves along the 200 training epochs for our models (with different compression levels) and the best multi-feature based baseline (SemCKD) in Fig.~\ref{fig_progressive_mIoU}. The plots reveal that 1) Compared to SemCKD, our method (both $prog^2$-$t$ and $prog^2$-$ts$) achieves faster convergence, less uncertainty and higher accuracy with the help of our proposed teacher-side multi-layer adapter, which aligns with our conclusions drawn from the Lipschitz continuity conditions. 2) $Prog^2$-$ts$ works better than $Prog^2$-$t$, which indicates the effectiveness of the teacher-student co-evolving design. A higher compression ratio ($prog^2$-$ts$  (3.52 M)) can still yield satisfactory performance. 3) For $prog^2$-$ts$, a heavier compression comes {with a performance drop} ($prog^2$-$ts$ (6.25 M) vs. $prog^2$-$ts$  (3.52 M)).

\begin{figure}[!t]
    \vspace{-14pt}
    \centering
	\addtocounter{figure}{0} 
	\centering  
	\subfloat[\footnotesize Student: ResNet 8×4 double(4.88M)]{
		\begin{minipage}[t]{0.45\linewidth}
        \includegraphics[width=1\linewidth,height=0.6\linewidth]{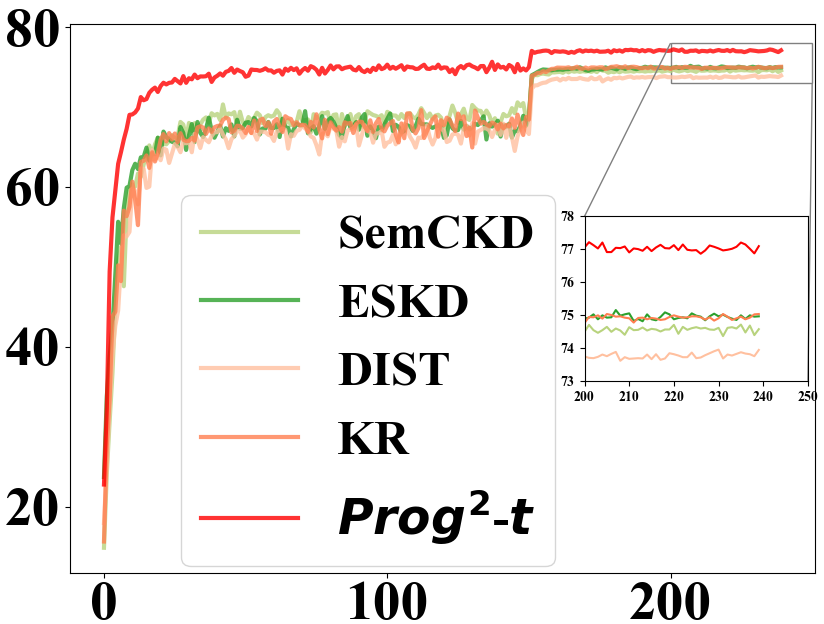}
		\end{minipage}%
	}%
	\subfloat[\footnotesize Student: ResNet 8×4(1.23M)]{
		\begin{minipage}[t]{0.45\linewidth}
        \includegraphics[width=1\linewidth,height=0.6\linewidth]{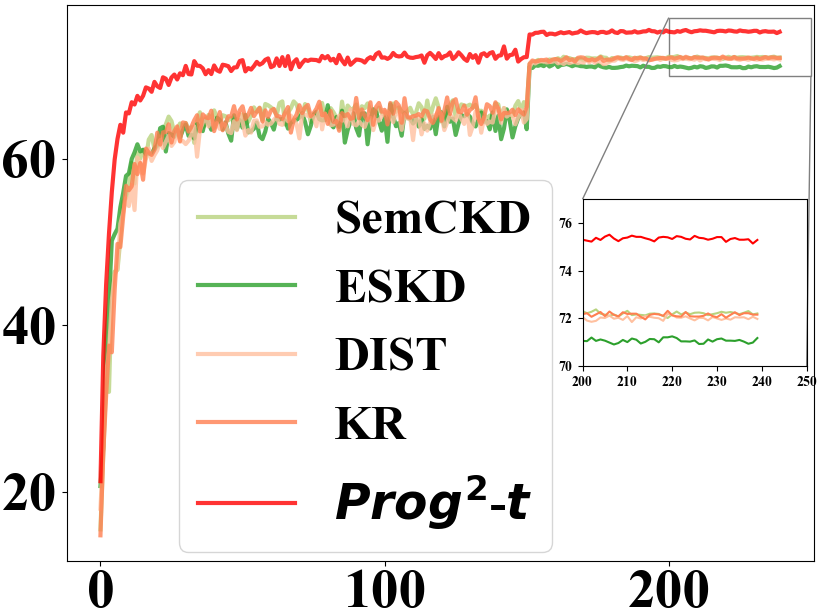}
		\end{minipage}%
	}%
    \\
	\subfloat[\footnotesize Student: ResNet 110(1.74M)]{
		\begin{minipage}[t]{0.45\linewidth}
        \includegraphics[width=1\linewidth,height=0.6\linewidth]{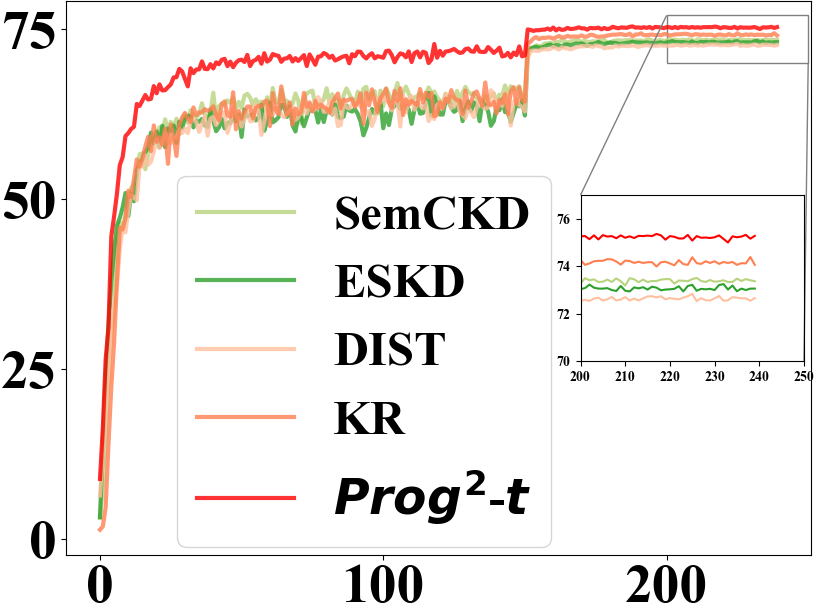}
		\end{minipage}%
	}%
	\subfloat[\footnotesize Student: ResNet 56(0.86M)]{
		\begin{minipage}[t]{0.45\linewidth}
        \includegraphics[width=1\linewidth,height=0.6\linewidth]{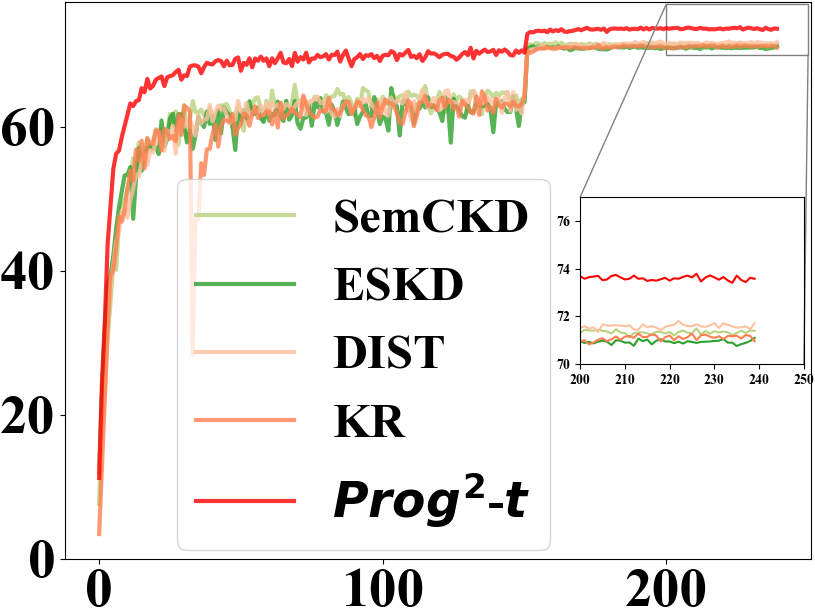}
		\end{minipage}%
	}%
    \\
	\subfloat[\footnotesize Student: ResNet 44(0.67M)]{
		\begin{minipage}[t]{0.45\linewidth}
        \includegraphics[width=1\linewidth,height=0.6\linewidth]{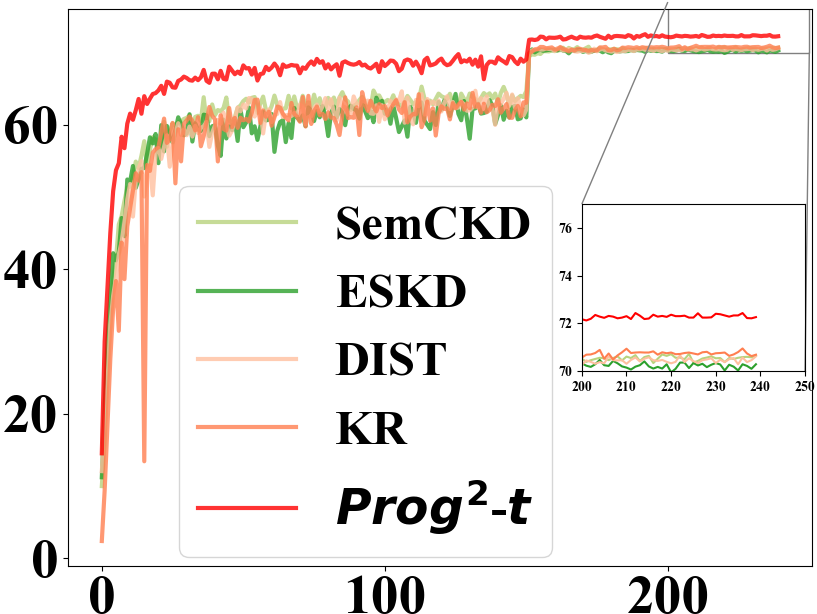}
		\end{minipage}%
	}%
	\subfloat[\footnotesize Student: ResNet 32(0.47M)]{
		\begin{minipage}[t]{0.45\linewidth}
        \includegraphics[width=1\linewidth,height=0.6\linewidth]{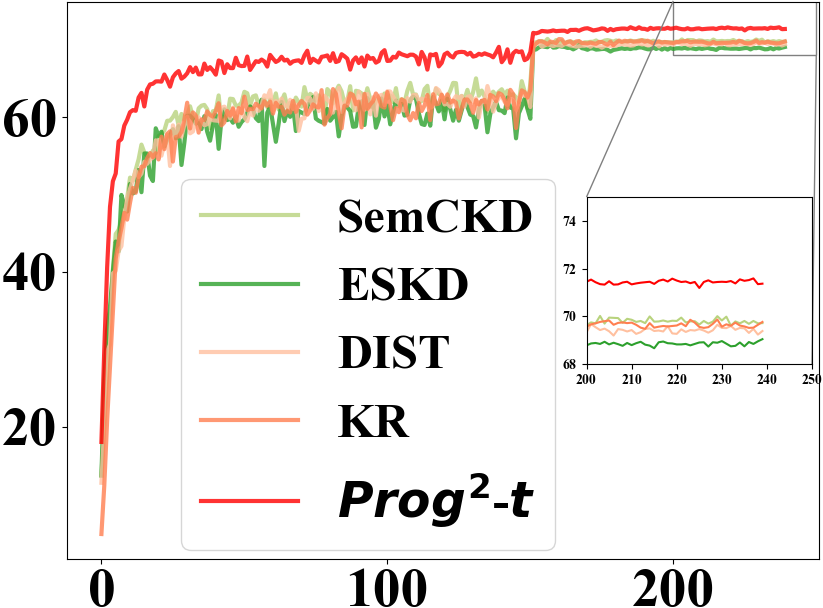}
		\end{minipage}%
	}%
    \caption{{The Top-1 accuracy curves for different-sized student models, using the same teacher: ResNet 32$\times$4 (7.43M) on the CIFAR-100 dataset.}}
    \label{fig:cifar-acc-lines}
    \vspace{-6pt}
\end{figure}

\subsection{Systematic analysis of user preference, task type, hyperparameters, and model architectures}

\noindent\textbf{Effect of user preference $\alpha$}
{In Fig.~\ref{fig_qos}, we present the Quality of Service (QoS) values across different settings of the hyperparameter $\alpha$, reflecting varying user preferences. For rigorous evaluation, we exclude the extreme $\alpha$ endpoints of 0 and 1, focusing on the $[0.2, 0.8]$ interval. Results indicate that while the feature distillation method SemCKD is restricted by significant time constraints, our $Prog^2$-$t$ approach consistently performs well across diverse preferences. As $\alpha$ approaches $0.2$, the priority shifts toward maximizing the final student accuracy. The proposed $Prog^2$-$t$ and $Prog^2$-$ts$ maintain favorable performance under balanced time and accuracy preferences. When $\alpha > 0.5$, prioritizing distillation execution time, the multi-stage progressive algorithm struggles to satisfy strict latency requirements. Since substantial distillation-based model compression is typically an offline, one-time process with amortizable training overhead, we primarily focus on accuracy-oriented scenarios. Consequently, our methods achieve optimal QoS when $\alpha < 0.5$.}




\noindent\textbf{Effect of knowledge distillation weight $\lambda$}
Next, we investigated the impact of modifying the hyperparameter $\lambda$ used to control the distillation weight size on the experimental results in the {CIFAR-100}, Cityscapes, and NYU-V2 datasets. We set $\lambda=1$ for the segmentation task, $\lambda=10$ for the depth task, and $\lambda=100$ for the classification task.

\noindent\textbf{Training curves of various students}
We present testing accuracy curves for variously sized student models (6 kinds from 4.88M to 0.47M) using the same teacher model (ResNet 32 $\times$ 4, 7.43M) on the CIFAR-100 dataset. The results reveal that conventional distillation methods exhibit minimal differences and do not perform satisfactorily when a significant disparity exists between the teacher and student models (Fig.~\ref{fig:cifar-acc-lines} (d)-(f)). In contrast, our approach significantly improves accuracy across the entire training process.

\begin{table}[!t]
\vspace{-10pt}
\begin{center}
\renewcommand{\arraystretch}{1.3}
\begin{tabular}{cc|cc}
\toprule
\textbf{dataset} & \textbf{$\lambda$} &\textbf{mIoU} & \textbf{Pixel Acc}\\
\cline{1-4}
\multirow{3}{*}{Cityscapes} & 0.1 & 58.81 & 93.03\\
 & 1 & \textbf{59.69} & \textbf{93.36}\\
 & 10 & 59.62 & 93.41\\
\cline{1-4}
\textbf{dataset} & \textbf{$\lambda$} & \textbf{Top-1 Acc} & \textbf{Top-5 Acc}\\
\cline{1-4}
\multirow{3}{*}{{CIFAR-100}} & 0.1 & 70.11 & 90.23\\
 & 50 & \textbf{72.85} & \textbf{90.68}\\
 & 100 & 72.11 & 90.29\\
\cline{1-4}
\textbf{dataset} & \textbf{$\lambda$} & \textbf{Abs Err} & \textbf{Rel Err}\\
\cline{1-4}
\multirow{3}{*}{NYU-V2} & 0.1 & 55.07 & 22.97\\
 & 10 & \textbf{51.98} & \textbf{21.31}\\
 & 50 & 53.24 & 21.24\\
\bottomrule
\end{tabular}
\caption{Different knowledge distillation weight $\lambda$ in Cityscapes, NYU-V2 and {CIFAR-100} dataset (\%).}
\label{tab_ablation2}
\end{center}
\vspace{-5mm}
\end{table} 


\section{CONCLUSION}~\label{sec:con} 
{\noindent In this work, we introduce Progressive$^2$, a novel knowledge distillation strategy designed to mitigate optimization instability via the iterative co-evolution of a progressively stronger teacher and a progressively smaller student. Extensive evaluations across semantic segmentation, depth estimation, and classification tasks demonstrate that our approach consistently facilitates efficient model compression, optimizing lightweight model deployment to resource-constrained mobile devices. While our teacher-side adapter provides theoretical insights into reducing uncertainty and accelerating convergence, these analyses rely on idealized Lipschitz smoothness assumptions. Given the highly non-convex optimization landscape of deep neural networks, these formulations serve as theoretical proxies rather than strict mathematical guarantees. Additionally, the relationship between progressive intrinsic dimension control and observed optimization improvements remains correlational, lacking formal causal validation. Future work will focus on relaxing these smoothness assumptions and exploring non-convex optimization theory to establish rigorous causal guarantees for knowledge transfer dynamics.}

\bibliographystyle{IEEEtran}
\bibliography{TMM}

\begin{IEEEbiography}
[{\includegraphics[width=1in,height=1.25in,clip,keepaspectratio]{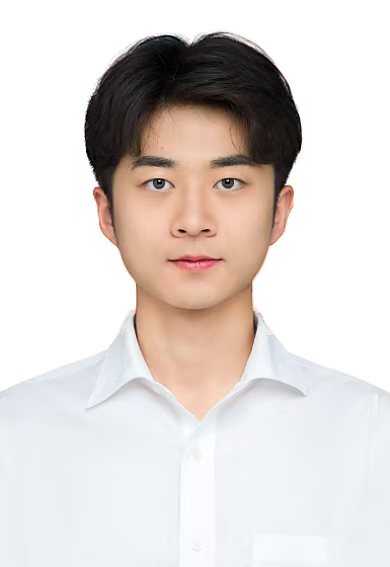}}]{Tiancong Cheng} is currently working toward the PhD degree in computer science and technology in the Northwestern Polytechnical University, Xi'an, China. He received the B.E. degree from Northwestern Polytechnical University, Xi’an, China, in 2022.  His research interests include mobile computing, knowledge transfer, and media intelligence. 
\end{IEEEbiography}
\vspace{-20mm}
\begin{IEEEbiography}
[{\includegraphics[width=1in,height=1.25in,clip,keepaspectratio]{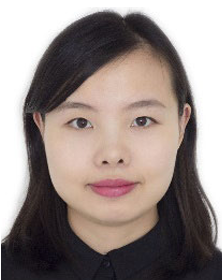}}]{Ying Zhang} (Member, IEEE) received the B.E. degree in computer science from Northwestern Polytechnical University, Xian, China, in 2009, and the Ph.D. degree in computer science from the National University of Singapore (NUS), in 2014. She is currently a professor at Northwestern Polytechnical University, Xi’an, China. Before that, she was the Head of the Data Security Unit and a Research Scientist with the Department of Cyber Security and Intelligence, Institute for Infocomm Research (I2R), A*STAR, Singapore. Her research interests include machine learning, spatiotemporal data mining, and multimedia.
\end{IEEEbiography}
\vspace{-10mm}
\begin{IEEEbiography}
[{\includegraphics[width=1in,height=1.25in,clip,keepaspectratio]{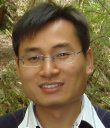}}]{Zhiwen Yu} (Senior Member, IEEE) is currently the Vice President of Harbin Engineering University, Harbin, China, and a Professor at Northwestern Polytechnical University, Xi'an, China. He received the PhD degree in computer science from Northwestern Polytechnical University, Xi’an, China, in 2005. He was an Alexander Von Humboldt fellow with Mannheim University, Germany, and a research fellow with Kyoto University, Kyoto, Japan. His research interests include ubiquitous computing, mobile crowd sensing, and human computer interaction. 
\end{IEEEbiography}
\vspace{-10mm}
\begin{IEEEbiography}
[{\includegraphics[width=1in,height=1.25in,clip,keepaspectratio]{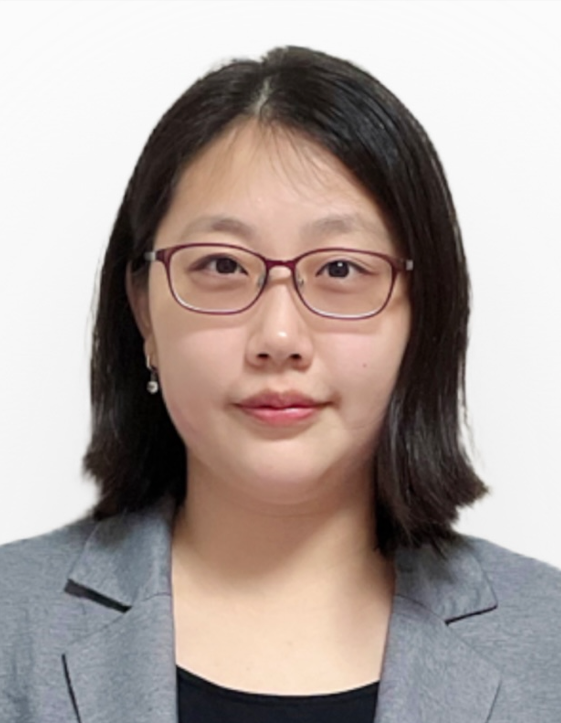}}] {Yifang Yin} received the BE degree from the Department of Computer Science and Technology, Northeastern University, Shenyang, China, in 2011 and the PhD degree from the National University of Singapore, Singapore, in 2016. She is currently a senior scientist with Institute for Infocomm Research (I$^{2}$R) in the Agency for Science, Technology and Research, Singapore (A*STAR). She also holds an adjunct faculty position with IIIT-Delhi. Before joining A*STAR, she worked as a senior research fellow with the Grab-NUS AI Lab with the National University of Singapore. She also worked as a Research Intern with the Incubation Center, Research and Technology Group, Fuji Xerox Co., Ltd., Japan, from October, 2014 to March, 2015. Her research interests include machine learning, weather science, spatiotemporal data mining, and multimodal analysis in multimedia.
\end{IEEEbiography}
\vspace{-10mm}
\begin{IEEEbiography}
[{\includegraphics[width=1in,height=1.25in,clip,keepaspectratio]{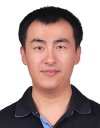}}]{Bin Guo} (Senior Member, IEEE) is currently a professor at Northwestern Polytechnical University, Xi’an, China. He received the PhD degree in computer science from Keio University, Minato, Japan, in 2009. He was a postdoctoral researcher with the Institut TELECOM SudParis, Essonne, France. His research interests include ubiquitous computing, mobile crowd sensing, and HCI.
\end{IEEEbiography}



\end{document}


\title{Appendix-Progressive$^2$: A Teacher-Student Progressive Co-Evolving Knowledge Distillation Method for Substantial Model Compression}

\author{Tiancong Cheng, \IEEEmembership{Student Member,~IEEE}, Ying Zhang\textsuperscript{*}, \IEEEmembership{Member,~IEEE}, Zhiwen Yu\textsuperscript{*}, \IEEEmembership{Senior Member,~IEEE}, Bin Guo, \IEEEmembership{Senior Member,~IEEE}}

\maketitle

\subsection{Detailed of the intrinsics dimension.}

\begin{table*}[t]
\centering
\caption{Intrinsic dimension (ID) of feature representations extracted from the last hidden layer for models with different capacities on Cityscapes and CIFAR-100, comparing pretrained (Pre.), randomly initialized (Rand.), and 5\%-trained (10 epochs) models.}
\label{tab:id_capacity}
\setlength{\tabcolsep}{4pt}
\begin{tabular}{c c c c c c c c c c}
\toprule
\multirow{2}{*}{\textbf{Dataset}} 
& \multicolumn{3}{c}{\textbf{Tiny}} 
& \multicolumn{3}{c}{\textbf{Small}} 
& \multicolumn{3}{c}{\textbf{Large}} \\
\cmidrule(lr){2-4}\cmidrule(lr){5-7}\cmidrule(lr){8-10}
& \textbf{Pre.} & \textbf{5\% (10 ep.)} & \textbf{Rand.}
& \textbf{Pre.} & \textbf{5\% (10 ep.)} & \textbf{Rand.}
& \textbf{Pre.} & \textbf{5\% (10 ep.)} & \textbf{Rand.} \\
\midrule
\multirow{2}{*}{\shortstack[c]{\textbf{Cityscapes}\\ \textbf{(SegNet-based)}}}
& \multirow{2}{*}{13.00 $\pm$ 0.08}
& \multirow{2}{*}{18.26 $\pm$ 0.12}
& \multirow{2}{*}{24.34 $\pm$ 0.45}
& \multirow{2}{*}{12.09 $\pm$ 0.07}
& \multirow{2}{*}{13.43 $\pm$ 0.10}
& \multirow{2}{*}{30.44 $\pm$ 0.34}
& \multirow{2}{*}{9.29 $\pm$ 0.06}
& \multirow{2}{*}{9.76 $\pm$ 0.06}
& \multirow{2}{*}{47.22 $\pm$ 0.46} \\
& & & & & & & & & \\

\multirow{2}{*}{\shortstack[c]{\textbf{CIFAR-100}\\ \textbf{(ResNet-based)}}}
& \multirow{2}{*}{14.50 $\pm$ 0.41}
& \multirow{2}{*}{12.45 $\pm$ 0.27} 
& \multirow{2}{*}{16.33 $\pm$ 0.22}
& \multirow{2}{*}{13.70 $\pm$ 0.32}
& \multirow{2}{*}{11.77 $\pm$ 0.28} 
& \multirow{2}{*}{17.84 $\pm$ 0.30}
& \multirow{2}{*}{13.18 $\pm$ 0.31}
& \multirow{2}{*}{11.28 $\pm$ 0.27} 
& \multirow{2}{*}{14.44 $\pm$ 0.13} \\
& & & & & & & & & \\
\bottomrule
\end{tabular}
\end{table*}

Subsequent sections detail the mathematical computation framework, the sampling protocol, and the layer-wise dynamic analysis of Intrinsic Dimension. This analysis rigorously proves that the representational capacity disparity between the teacher and student models is fundamentally a significant Intrinsic Dimension gap when viewed through the lens of geometric manifold complexity.

We introduce a clearer theoretical quantity to characterize the teacher-student knowledge gap and validate the design mechanism of Progressive$^2$. We employ the \textbf{Intrinsic Dimension (ID)}~\cite{levina2004maximum} of model features to quantify this disparity. The estimation of ID, derived from the TwoNN~\cite{facco2017estimating} estimator that is based on computing the ratio between the distances to the second and first nearest neighbors (NN), has been extended to analyze the training dynamics of deep neural networks~\cite{ansuini2019intrinsic} and subsequently applied to assess the representational differences in feature-based knowledge distillation~\cite{li2023rethinking}. We adopt the calculation framework provided by the '\textit{IntrinsicDimDeep}' repository~\cite{ansuini2019intrinsic} and strictly follow the multi-round sampling strategy established in recent literature~\cite{li2023rethinking}. This metric clarifies the underlying geometric challenge: the capacity gap between the powerful teacher and the compact student manifests mathematically as a significant ID gap. While direct distillation struggles to bridge this discrepancy, Progressive$^2$ constructs intermediate transition states. This strategy effectively decomposes the substantial ID gap into multiple manageable increments, thereby smoothing the transition from high-dimensional manifolds to low-dimensional representations.

Specifically, the student and teacher networks exhibit distinct intrinsic dimensions at identical architectural locations, such as the input features $f^t$ and $f^s$ of the last hidden layer (or the specific task layer). The representational intrinsic dimension of the teacher model is typically smaller, which indicates a more precise abstraction of knowledge and yields more regularized final outputs. A substantial discrepancy in model capacity or performance between the teacher and student networks correspondingly enlarges the gap in their intrinsic dimensions. Under these circumstances, conventional feature distillation methods that rely on aligning feature representations become ineffective. The highly abstract nature of the teacher features prevents the compact student model from approximating them accurately. This optimization difficulty is precisely quantifiable through the intrinsic dimension and is further demonstrated by the toy example in Fig. 5 of our manuscript, which illustrates that an increased representational disparity degrades the stability of the distillation process. We subsequently provide a detailed formulation of the intrinsic dimension methodology.

\subsubsection{A Systematic Explanation of Intrinsic Dimension}
Mapping a sample to an activation or feature vector at a specific layer of a deep network yields an embedding dimension equal to the number of neurons or flattened channels. However, these high-dimensional points typically do not span the entire ambient space but rather concentrate on lower-dimensional structures, such as manifolds or clustered sets~\cite{ansuini2019intrinsic}. The intrinsic dimension defines the minimum degrees of freedom or coordinates required to describe these representation points, representing the minimal parameter count needed to characterize the feature set without significant information loss. This concept emphasizes the geometric complexity of the data representations rather than the explicit dimensionality of the network layer. Because deep networks are universally over-parameterized~\cite{zhang2017understanding} and intra-layer activations exhibit high redundancy~\cite{cogswell2015reducing}, relying solely on the embedding dimension fails to capture the true representational complexity. Previous studies~\cite{ansuini2019intrinsic,liurethinking,cohen2020separability} have shown that the intrinsic dimension of well-trained networks is often several orders of magnitude smaller than the number of units per layer.

Given a dataset $\mathcal{D}=\{(x_i,y_i)\}_{i=1}^{M}$, we denote by $\theta_{\mathrm{Enc}}$ the hidden parameters of the network encoder that produces the last hidden layer representation. For each input sample $x_i$, we extract the last hidden layer feature as:
$$f_i=\theta_{\mathrm{Enc}}(x_i)\in\mathbb{R}^{D},$$
where $D$ is the embedding dimension of the last hidden layer. The resulting feature set is:
$$\mathcal{F}=\{f_i\}_{i=1}^{N}, \quad N\le M,$$
which forms a point cloud in $\mathbb{R}^{D}$. The intrinsic dimension aims to quantify the effective degrees of freedom of $\mathcal{F}$, typically satisfying $d\ll D$. To improve comparability across models and training stages, we apply $L_2$-normalization to each feature vector, yielding
$$\tilde{f}_i=\frac{f_i}{\lVert f_i\rVert_2}.$$
All subsequent computations are performed on the normalized set$$\tilde{\mathcal{F}}=\{\tilde{f}_i\}_{i=1}^{N}.$$

We adopt the Two Nearest Neighbors estimator to compute the intrinsic dimension of $\tilde{\mathcal{F}}$. For each point $\tilde{f}_i$, we compute the Euclidean distance to its nearest and second-nearest neighbors in $\tilde{\mathcal{F}}\setminus\{\tilde{f}_i\}$, denoted respectively as:
$$r_{i,1}=\min_{j\neq i}\lVert \tilde{f}_i-\tilde{f}_j\rVert_2,$$
and
$$r_{i,2}=\min_{j\neq i, j\neq j^{\star}}\lVert \tilde{f}_i-\tilde{f}_j\rVert_2,$$
where $j^{\star}=\arg\min_{j\neq i}\lVert \tilde{f}_i-\tilde{f}_j\rVert_2$. We then define the distance ratio$$\mu_i=\frac{r_{i,2}}{r_{i,1}}, \quad \mu_i\ge 1.$$

Under the standard assumption that points are locally distributed approximately uniformly on a $d$-dimensional manifold and that local neighborhoods are sufficiently small to render curvature effects negligible, the cumulative distribution of $\mu$ satisfies$$F(\mu)=\mathbb{P}(\mu_i\le \mu)=1-\mu^{-d}.$$
Rearranging this equation yields a linear relation:
$$-\log\bigl(1-F(\mu)\bigr)=d\log(\mu).$$

In practice, we sort the sequence $\{\mu_i\}_{i=1}^{N}$ in ascending order to obtain:
$$\mu_{(1)}\le \mu_{(2)}\le \cdots \le \mu_{(N)},$$
and compute the empirical cumulative distribution function:
$$\hat{F}(\mu_{(k)})=\frac{k-0.5}{N}.$$

We define $x_k=\log\bigl(\mu_{(k)}\bigr)$ and $y_k=-\log\bigl(1-\hat{F}(\mu_{(k)})\bigr)$, estimating the intrinsic dimension $\hat{d}$ as the slope of the best linear fit given by:
$$\hat{d}=\arg\min_{d}\sum_{k\in\Omega}\left(y_k-d x_k\right)^2.$$

To mitigate finite-sample and boundary effects, we fit only over a central quantile range $\Omega$, effectively excluding the extreme tails of the distance ratios.

\subsubsection{Sampling protocol and uncertainty quantification}

Because the TwoNN estimator relies on nearest-neighbor statistics, the resulting estimation is inherently sensitive to sampling noise and class imbalance. To mitigate these issues, we implement a comprehensive evaluation protocol. 
\begin{itemize}
    \item First, we maintain a fixed evaluation subset to construct the normalized feature set $\tilde{\mathcal{F}}$ across all evaluated models and training epochs. This consistency ensures that any observed variations in the intrinsic dimension strictly reflect representational shifts rather than random sampling fluctuations. 
    \item Second, we incorporate bootstrap aggregation to stabilize the estimation. We repeat the TwoNN computation $B$ times through subsampling without replacement. During each iteration $b$, we randomly sample a data fraction $\rho$, typically set to $0.9$, to compute the dimension estimate $\hat{d}^{(b)}$. The final intrinsic dimension and its standard deviation are formally defined as:
    $$\mathrm{ID}=\frac{1}{B}\sum_{b=1}^{B}\hat{d}^{(b)},$$
    $$\mathrm{Std}=\sqrt{\frac{1}{B-1}\sum_{b=1}^{B}\left(\hat{d}^{(b)}-\mathrm{ID}\right)^2}.$$
    \item Furthermore, our experiments encompass semantic segmentation tasks on the Cityscapes~\cite{cordts2016cityscapes} dataset, where intermediate feature representations naturally form a union of multiple sub-manifolds corresponding to distinct semantic categories and mixed boundary regions. Applying the TwoNN estimator directly to all pixel features introduces significant biases. Specifically, class boundaries and mixed regions artificially inflate the estimated intrinsic dimension. Concurrently, the severe class imbalance characteristic of the Cityscapes dataset renders the results highly dependent on the underlying sampling distribution, precluding fair comparisons. To establish a reliable metric for segmentation tasks, we employ a class-balanced sampling strategy. We extract a predetermined, equal number of points per class, drawing as many samples as possible for minority classes, and subsequently truncate all categories to a uniform sample size. These balanced subsets are then aggregated to compute the overall intrinsic dimension. This mechanism isolates the calculation from the influence of dataset class proportions and prevents long-tail categories from dominating the estimation process.
\end{itemize}

\begin{table*}[t]
\centering
\caption{Layer-wise intrinsic dimension (ID) of the trained SegNet-based model on Cityscapes. IDs are estimated from the feature representations of each layer.}
\label{tab:segnet_layerwise_id}
\setlength{\tabcolsep}{4pt}
\renewcommand{\arraystretch}{1.05}
\resizebox{\textwidth}{!}{%
\begin{tabular}{c c c c c c c c c c c}
\toprule
\textbf{Model} & \textbf{L1} & \textbf{L2} & \textbf{L3} & \textbf{L4} & \textbf{L5} & \textbf{L6} & \textbf{L7} & \textbf{L8} & \textbf{L9} & \textbf{L10} \\
\midrule
\textbf{Tiny} & 5.61 $\pm$ 0.07 & 8.90 $\pm$ 0.10 & 11.11 $\pm$ 0.12 & 11.60 $\pm$ 0.08 & 12.70 $\pm$ 0.09 & 8.05 $\pm$ 0.06 & 6.30 $\pm$ 0.07 & 12.30 $\pm$ 0.10 & 17.54 $\pm$ 0.09 & 13.00 $\pm$ 0.08 \\
\textbf{Small} & 6.40 $\pm$ 0.07 & 9.05 $\pm$ 0.13 & 10.37 $\pm$ 0.08 & 11.35 $\pm$ 0.07 & 12.32 $\pm$ 0.10 & 12.68 $\pm$ 0.06 & 12.37 $\pm$ 0.07 & 16.94 $\pm$ 0.10 & 21.92 $\pm$ 0.10 & 12.09 $\pm$ 0.07 \\
\textbf{Large} & 6.72 $\pm$ 0.08 & 9.40 $\pm$ 0.14 & 10.02 $\pm$ 0.07 & 9.32 $\pm$ 0.08 & 9.70 $\pm$ 0.07 & 12.69 $\pm$ 0.10 & 14.87 $\pm$ 0.09 & 16.70 $\pm$ 0.10 & 15.93 $\pm$ 0.10 & 9.29 $\pm$ 0.06 \\
\bottomrule
\end{tabular}%
}
\end{table*}

\subsubsection{Results and Analysis of Intrinsic Dimension}
Following the above analysis, we compute the intrinsic dimension for models of varying capacities. Specifically, we evaluate SegNet-based~\cite{badrinarayanan2017segnet} architectures for semantic segmentation on the Cityscapes~\cite{cordts2016cityscapes} dataset and ResNet-based~\cite{he2016deep} architectures for image classification on the CIFAR-100~\cite{krizhevsky2009learning} dataset. The feature representations utilized for the intrinsic dimension estimation are extracted directly from the output of the last hidden layer in each respective model (specifically, the feature map immediately before the 'pred\_task' layer in SegNet, and the feature tensor immediately before the 'avgpool' layer in ResNet). The corresponding quantitative results are summarized in the subsequent table Table~\ref{tab:id_capacity}.

We compare three SegNet variants of varying capacities derived from established channel pruning technique~\cite{li2017pruning}. The pruning ratios applied to the shared modules are 0, 50\%, and 75\%, which yield model sizes of 24.98M, 6.25M, and 1.57M parameters, respectively. Similarly, for the Cifar-100 dataset, we evaluate multiple ResNet architectures with parameter counts of 1.74M, 0.67M, and 0.18M. Analysis of the results is as follows:
\begin{itemize}
    \item 1) The quantitative results indicate that the Intrinsic Dimension (ID) of randomly initialized models consistently exceeds that of their trained counterparts.
    \item 2) The ID of models trained on the target datasets corroborates findings from prior studies~\cite{ansuini2019intrinsic}, demonstrating that improvements in test performance are accompanied by a progressive reduction in the intrinsic dimension. Specifically, as model capacity decreases from large to small, the task performance degrades while the ID increases (e.g., SegNet-Large: mIoU is 59.15 and ID is 9.29, while SegNet-Small: mIoU is 57.89 and ID is 13.09).
    \item 3) Experimental results on the CIFAR-100 dataset indicate that the ID of the features extracted from the final layer initially experiences a sharp decline before gradually increasing, which aligns with observations in prior literature. Conversely, the ID of the SegNet architecture on the Cityscapes dataset exhibits a continuous downward trend.
    \item 4) Notably, although the large model exhibits a high ID at random initialization, this metric converges rapidly after a minimal number of training epochs, specifically within the first 5\% of the total training schedule. Consequently, analyzing the capacity gap between the teacher and student networks throughout the entire optimization trajectory reveals that the overall distillation difficulty depends on the cumulative ID disparity. The distillation process proves effective when the representational dimension (ID) of the student network rapidly approaches that of the teacher. However, an excessive capacity gap results in a persistently large deviation in the ID of their feature representations. For instance, on the Cityscapes dataset, the trained SegNet-Large model exhibits an intrinsic dimension of 9.29, whereas the untrained SegNet-Small model presents a significantly higher value of 24.34. Furthermore, the intrinsic dimension discrepancy between the two models remains substantial throughout the entire training process, plateauing at a notable gap of 3.80, which represents the difference between their respective final converged values of 13.09 and 9.29.
    \item (Optional) 5) We present the layer-wise ID values for pre-trained models of varying capacities on the Cityscapes dataset in Table~\ref{tab:segnet_layerwise_id}. The layer-wise dynamics within individual models generally align with established findings in the literature~\cite{ansuini2019intrinsic}, demonstrating that the ID initially increases in the early layers before subsequently decreasing in the deeper layers.
\end{itemize}
Consequently, investigating the final high-level semantic representation is essential (in the SegNet model, this feature is the output of layer 'L10' in Table~\ref{tab:segnet_layerwise_id}). This representation has undergone multiple convolutional operations, non-linear transformations, and receptive field integrations, yet remains uncompressed by the fully connected classification head into the label space. Functionally, it closely approximates the intrinsic dimension variations of the universal representation carrier following the knowledge extraction process, reflecting the manifold complexity of the final model representations.

The preceding analysis indicates that the difficulty of distillation extends beyond a simple capacity mismatch. Instead, it constitutes a profound optimization challenge driven by the geometric structural discrepancies between the trained teacher and student representations, particularly the intrinsic dimension gap, coupled with inter-layer misalignments. Therefore, attempting to directly distill knowledge from a converged large teacher model to a randomly initialized small student model at the onset of training encounters a substantial intrinsic dimension gap, which is further exacerbated by their capacity differences. This representational disparity persists even after the student model converges, since the fully trained SegNet-Small model achieves an intrinsic dimension of 13.09, whereas the SegNet-Large model maintains a value of 9.29.

To address this optimization bottleneck, our progressive approach shifts the operational target from the conventional student-side feature map fusion to a teacher-side feature map fusion strategy. This modification provides enhanced control over the intrinsic complexity of the manifold representations output by the teacher model. Consequently, the student model can acquire knowledge across a reduced intrinsic dimension gap, leading to superior knowledge transfer.

\begin{figure*}[htbp]
    \centering
    \begin{subfigure}[b]{0.49\linewidth}
        \centering
        \includegraphics[width=\linewidth]{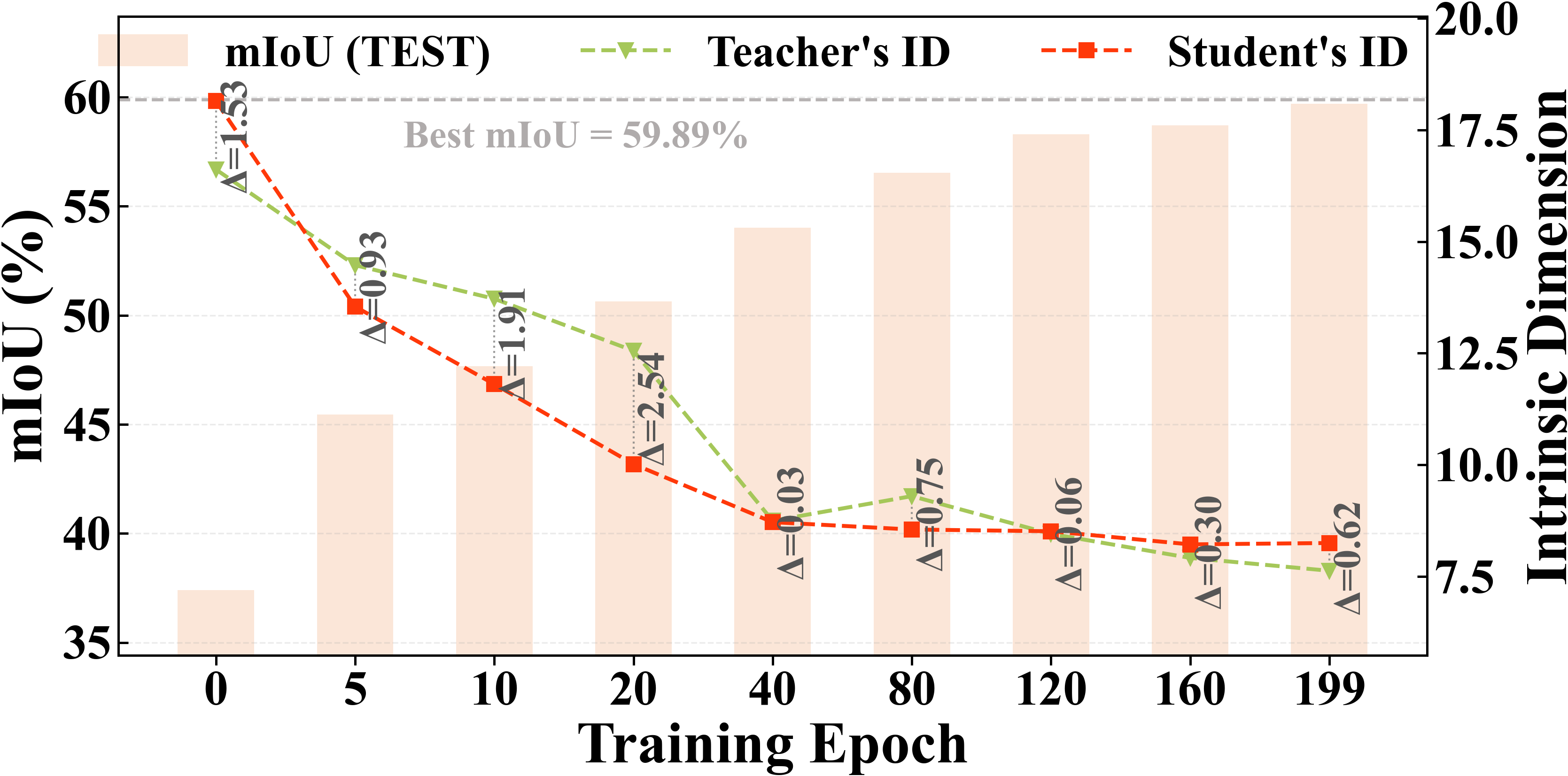}
        \caption{Prog$^2$-t}
    \end{subfigure}
    \hfill
    \begin{subfigure}[b]{0.49\linewidth}
        \centering
        \includegraphics[width=\linewidth]{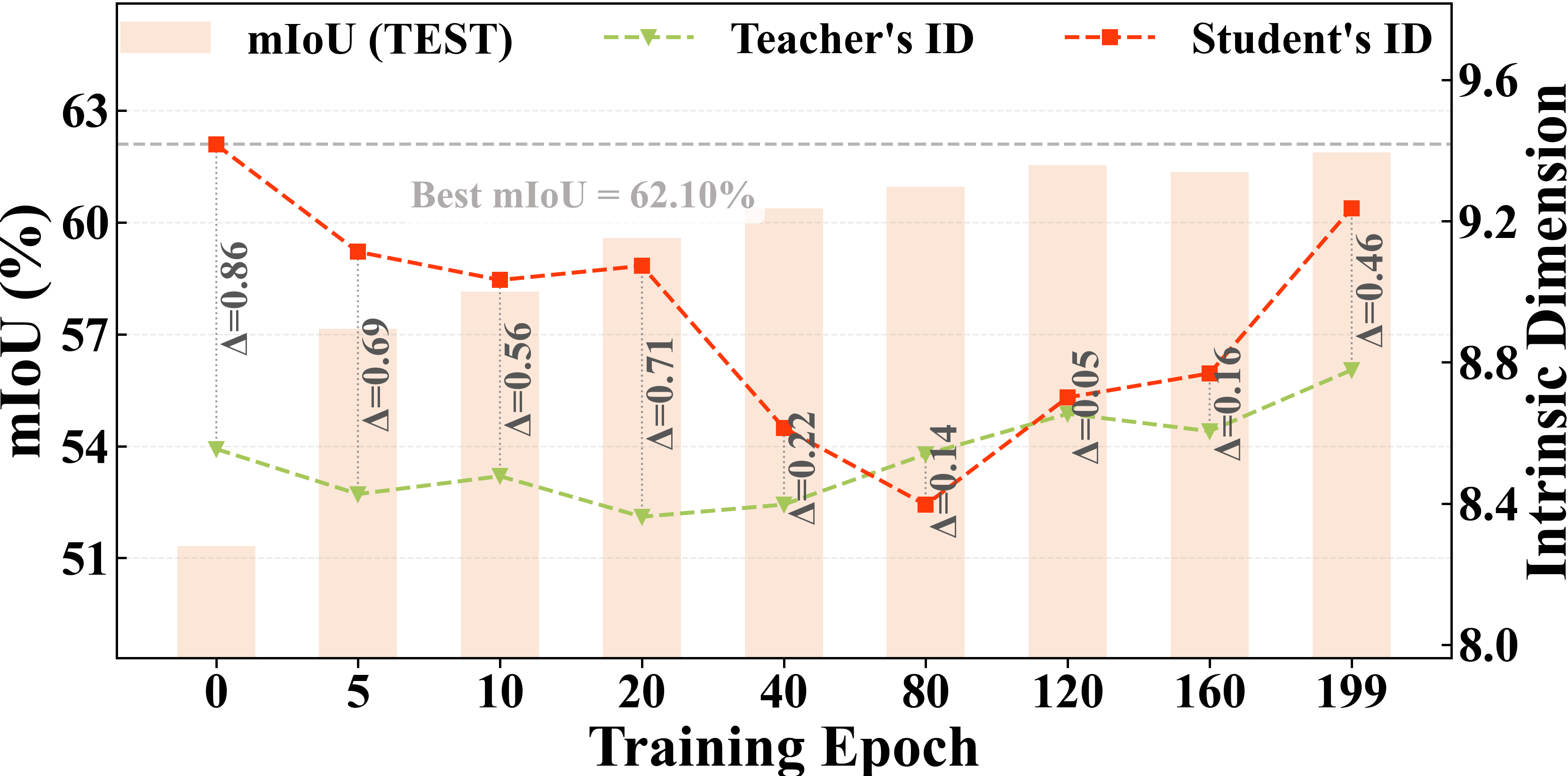}
        \caption{Prog$^2$-ts}
    \end{subfigure}

    \vspace{0.5em} 

    \begin{subfigure}[b]{0.49\linewidth}
        \centering
        \includegraphics[width=\linewidth]{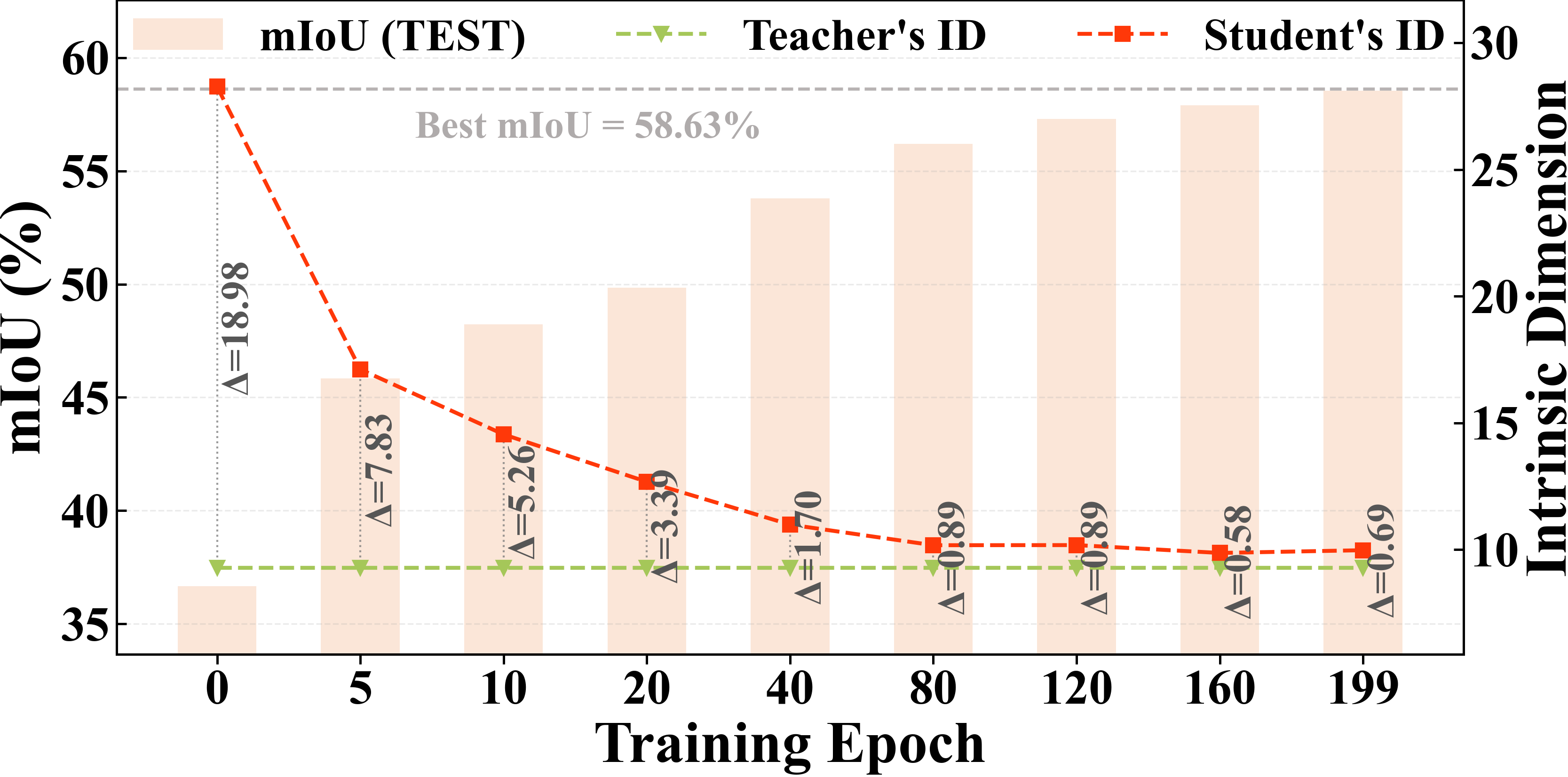}
        \caption{DIST}
    \end{subfigure}
    \hfill
    \begin{subfigure}[b]{0.49\linewidth}
        \centering
        \includegraphics[width=\linewidth]{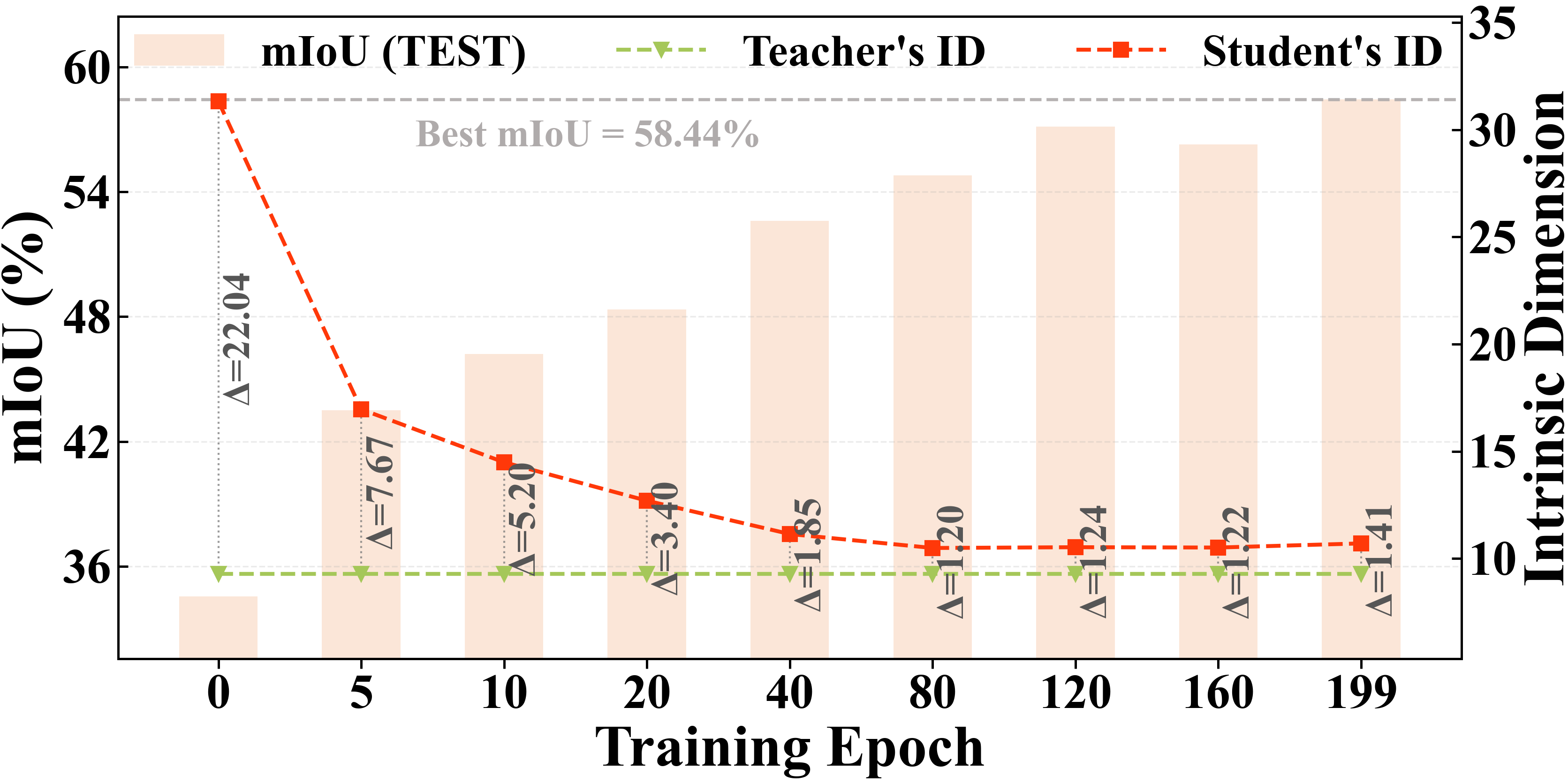}
        \caption{SemCKD}
    \end{subfigure}

    \vspace{0.5em}

    \begin{subfigure}[b]{0.49\linewidth}
        \centering
        \includegraphics[width=\linewidth]{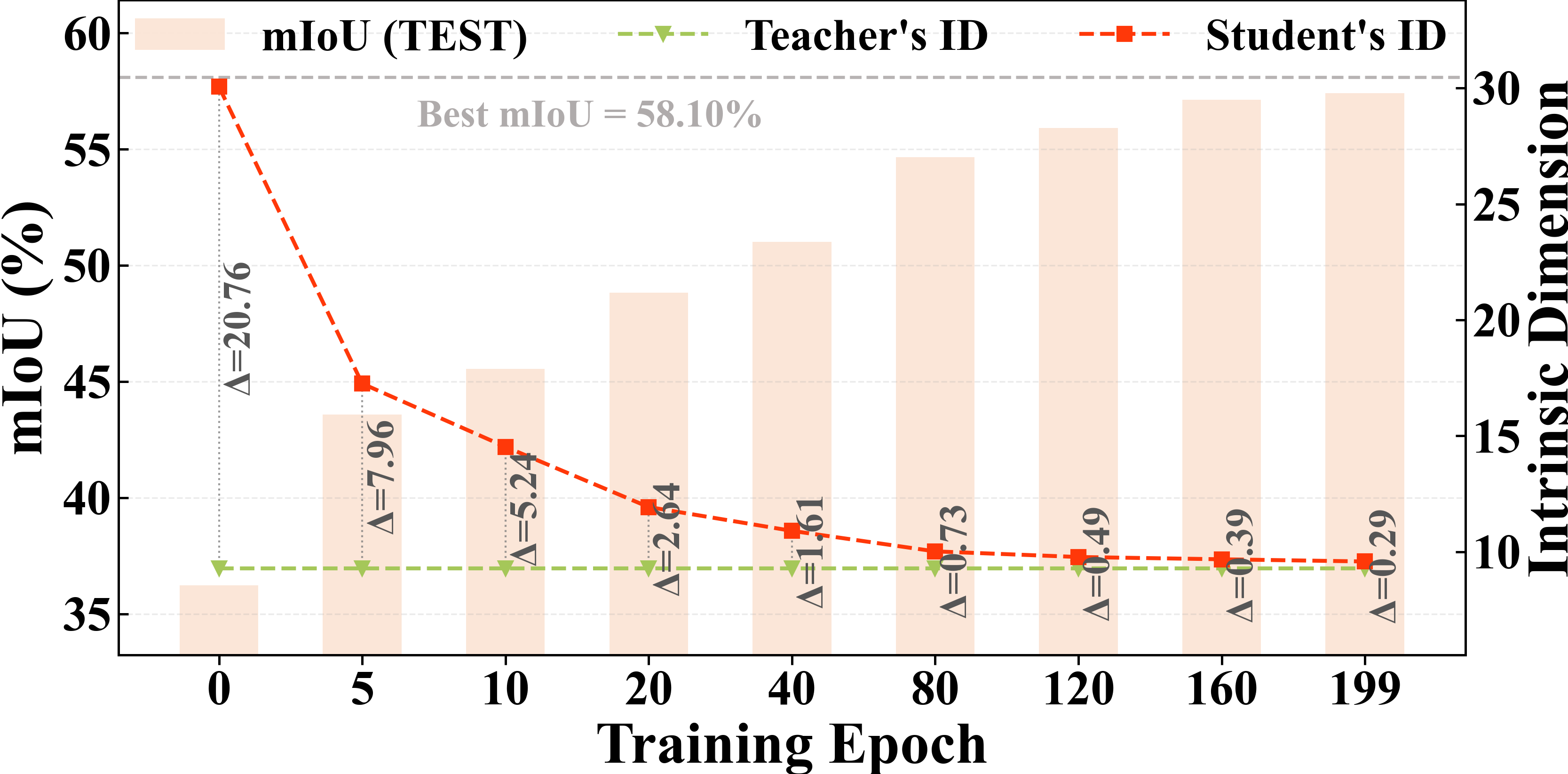}
        \caption{VanillaKD}
    \end{subfigure}
    \hfill
    \begin{subfigure}[b]{0.49\linewidth}
        \centering
        \includegraphics[width=\linewidth]{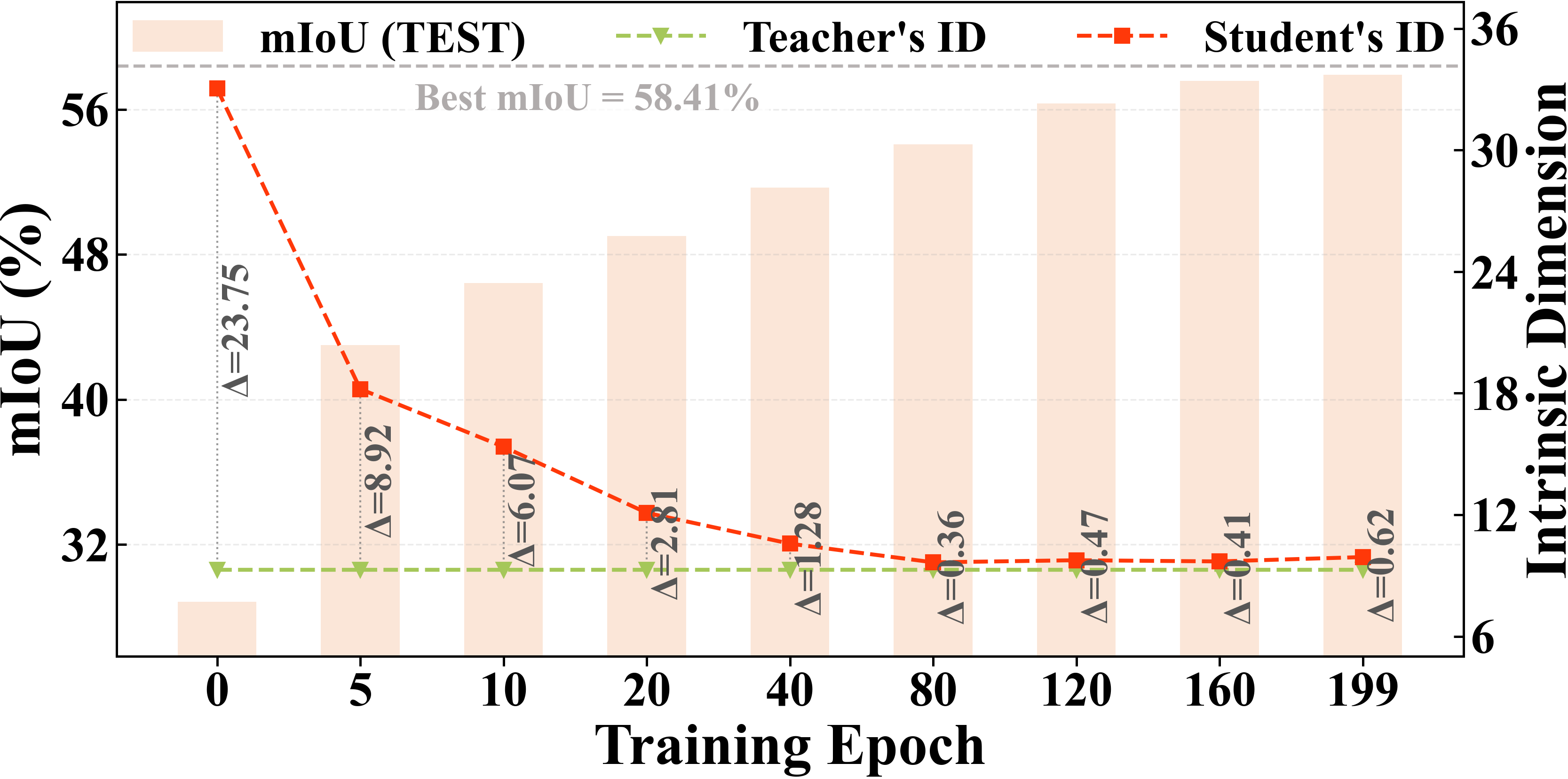}
        \caption{BAMKD}
    \end{subfigure}

    \vspace{0.5em}

    \caption{Teacher, student intrinsic dimension and mIoU at typical epochs.}
    \label{fig:tea_stu_ID}
\end{figure*}

\subsubsection{Ours Solution}
Our progressive frameworks $Prog^2$-$t$ and $Prog^2$-$ts$ facilitate a co-evolutionary optimization process between the teacher and the student (Fig.~\ref{fig:tea_stu_ID} (a,b)). Both the progressive teacher adaptation and the joint progressive scheme necessitate a remapping of the teacher features. Although this remapping introduces a marginal parameter overhead during training, it yields superior performance upon deployment and across downstream tasks. By systematically remapping the teacher features, the intrinsic dimension gap between the two models is consistently constrained within a narrow and manageable range (Fig.~\ref{fig:tea_stu_ID} (a,b)). This alignment allows the student network to assimilate the hierarchical progression of teacher features from simple to complex. Compared to mimicking static logit distributions~\cite{hinton2015distilling,mirzadeh2020improved,huang2022knowledge}, this dynamic alignment equips the student with the ability to learn the derivation of knowledge step by step rather than merely memorizing the final output.

\subsection{Theoretical support of accelerating convergence.}

This proof demonstrates that the training process of our method converges, and its convergence speed is faster compared to other methods. First, we give the sufficient condition for the loss to decrease ($\mathcal{L}(\theta^{p+1}) - \mathcal{L}(\theta^p) < 0$). Then, we define the relationship between adapter complexity $|\theta|$ and the update step size $\alpha$ as: $|\theta_{m-to-n}| > |\theta_{1-to-1 (ours)}| \implies \alpha_{m-to-n} < \alpha_{1-to-1 (ours)}$

The goal of training deep neural networks lies in finding an optimal set of parameters $\theta^{*}$ that achieves low loss $\mathcal{L}_{task}(\theta_{Net})$ where the independent variable is $\theta_{Net}$ on the target task. This implies that the task loss must be continually minimized $\mathcal{L}_{task} (\theta_{Net}^p )\downarrow$ over time $p\uparrow$ and parameters $\theta_{Net}$ are continually optimized to $\theta^*$. 

\textit{Theorem 3.1} \textit{Based on~\cite{liu2021conflict}, assume that the gradient of the loss function satisfies H-Lipschitz continuity, i.e. $\left \| \bigtriangledown \mathcal{L}_{task}(x) -\bigtriangledown \mathcal{L}_{task}(y) \right \| \le H\left \| x-y \right \| $, where $0\le H\le \infty$. Deep neural networks $\mathcal{L}_{task}(\theta_{Net})$ can converge to a Pareto stationary point with the update step size $\alpha$ meets $0< \alpha \le \frac{1}{H}$ and the searched gradient $g$ satisfies:}
\begin{equation}
    \left \|\bigtriangledown \mathcal{L}_{task}(\theta_{Net}^{p})-g\right \|< \left \| \bigtriangledown \mathcal{L}_{task}(\theta_{Net}^{p}) \right \|,
\end{equation} where $\bigtriangledown \mathcal{L}_{task}(\theta_{Net}^p)$ represents the update gradient of $\theta_{Net}^P$ from loss $\mathcal{L}_{task}(\cdot)$ at time $p$. The proof is as follows:
\begin{align}
    & \mathcal{L}(\theta_{Net}^{p+1})-\mathcal{L}(\theta_{Net}^{p})\\
    =&\mathcal{L}(\theta_{Net}^{p}-\alpha\cdot  g)-\mathcal{L}(\theta_{Net}^{p}) \\
    &{\small (Second-order\;Taylor \; approximation)}\\
    \approx& - \alpha  g\cdot \bigtriangledown \mathcal{L}(\theta_{Net}^{p})+\frac{\bigtriangledown^2 \mathcal{L}(\theta_{Net}^{p})}{2}[(-\alpha\cdot  g)]^2\\
    &{\small (Lipschitz \; continuity)}\\
    \le& - \alpha  g\cdot \bigtriangledown \mathcal{L}(\theta_{Net}^{p})+\frac{H}{2}[(-\alpha\cdot  g)]^2\\
    \le&- \alpha  g\cdot \bigtriangledown \mathcal{L}(\theta_{Net}^{p})+\frac{\alpha}{2}\left \| g \right \| ^2\\
    =&\frac{\alpha}{2}\left \|\bigtriangledown \mathcal{L}(\theta_{Net}^{p})-g\right \|^2-\frac{\alpha}{2}\left \| \bigtriangledown \mathcal{L}(\theta_{Net}^{p}) \right \|^2 \\
    =&-\frac{\alpha}{2}\left ( \left \| \bigtriangledown \mathcal{L}(\theta_{Net}^{p}) \right \|^2- \left \|\bigtriangledown \mathcal{L}(\theta_{Net}^{p})-g\right \|^2\right ),
    \label{equ_proof}
\end{align} when $ \left \|\bigtriangledown \mathcal{L}_{task}(\theta_{Net}^{p})-g\right \|< \left \| \bigtriangledown \mathcal{L}_{task}(\theta_{Net}^{p}) \right \|$, Eqn.~\ref{equ_proof}  is strictly negative. Hence, we have a strictly decreasing sequence ${\mathcal{L}_{task}^p(\theta_{Net})}_{p\in T}$, $T$ represents the training period. Then loss $\mathcal{L}_{task}(\theta_{Net})$ must converge because it has a lower bound.

In a typical knowledge distillation scenario, the loss function during the training process can be expressed as: $\mathcal{L}= \mathcal{L}_{task}(\theta_{Net} )+ {\textstyle \sum_{i=1}^{R}}\mathcal{L}_{kd}^i(\theta_{Net}, \theta_{Map}^i)$, $R$ is the number of learning directions from multi-layer teacher knowledge, which can be simplified to Eqn.~\ref{equ_loss_kd+ts} as $\left \| \theta_{Net} \right \| \gg \left \| \theta_{Map} \right \|$. 
\begin{equation}
    \mathcal{L}(\theta)= \mathcal{L}_{task}(\theta)+  \sum_{i=1}^{R}   \mathcal{L}_{kd}^i(\theta).
    \label{equ_loss_kd+ts}
\end{equation} The parameter $\theta$ is subject to optimization and can be updated according to $\theta^{p+1}=\theta ^p-\alpha \cdot g$, where $\alpha$ is the update step and $g$ is the unified gradient and $g=g_{ts} + {\textstyle \sum_{i=1}^{R}} g_{kd}^i$ in the knowledge distillation scenario, $g_{ts}$ and $g_{kd}$ represent $\bigtriangledown \mathcal{L}_{task}(\theta)$ and $\bigtriangledown \mathcal{L}_{kd}(\theta)$ respectively. Based on \textit{Theorem3.1}, we will demonstrate that an increased number of distillation gradient directions $R$ from the teacher results in slower convergence of $\mathcal{L}(\theta)$.

Assuming that all loss function gradients ($\bigtriangledown \mathcal{L}_{task}(\theta), \{ \bigtriangledown \mathcal{L}_{kd}^i(\theta)\}_{i}^R$) satisfy $H^i$-Lipschitz continuity, i.e. $\left \| \bigtriangledown \mathcal{L}^i(x) -\bigtriangledown \mathcal{L}^i(y) \right \| \le H^i\left \| x-y \right \| $, where $0\le H\le \infty$. The largest Lipschitz constant among them denoted as $H^{max}$, and all losses updated by the same step $\alpha$. We can apply the Triangle Inequality theorem to conclude that the combined loss gradient $\bigtriangledown \mathcal{L}(\theta)$ is also Lipschitz continuous, with a Lipschitz constant equal to:
\begin{equation}
    H^*= (R+1)\cdot H^{max}.
\end{equation} The detailed derivations of this equation are as follows:
\begin{align}
&\left \| \bigtriangledown \mathcal{L}(x )-\bigtriangledown \mathcal{L}(y) \right \| \\
=&\left \| \sum_{i=1}^{R}  (\bigtriangledown \mathcal{L}^i(x )-\bigtriangledown \mathcal{L}^i(y )) \right \| \\
=&\left \| \bigtriangledown \mathcal{L}^1(x )-\bigtriangledown \mathcal{L}^1(y )+\cdots+\bigtriangledown \mathcal{L}^R(x )-\bigtriangledown \mathcal{L}^R(y ) \right \| \\
\end{align}
\begin{align}
\le &\left \| \bigtriangledown \mathcal{L}^1(x )-\bigtriangledown \mathcal{L}^1(y ) \right \| +\cdots+\left \| \bigtriangledown \mathcal{L}^R(x )-\bigtriangledown \mathcal{L}^R(y ) \right \|\\
\le &(H^1+\cdots+H^R)\left \| x-y \right \| \\
\le & R\cdot H^{max}\left \| x-y \right \|.
\label{equ_derivation}
\end{align} The Lipschitz constant of $\bigtriangledown \mathcal{L}(\theta)$ is $R\cdot H^{max}$.

{Then, it can be derived that, in a typical knowledge distillation scenario}, the loss function during the training process can be expressed as: $\mathcal{L}= \mathcal{L}_{ts}(\theta_{Net} )+ {\textstyle \sum_{i=1}^{R}}\mathcal{L}_{kd}^i(\theta_{Net},\theta_{Map}^i)$, $R$ is the number of learning directions from multi-layer teacher knowledge, which can be simplified to Eqn.~\ref{equ_L} as $\left \| \theta_{Net} \right \| \gg \left \| \theta_{Map} \right \|$. 
\begin{equation}
    \mathcal{L}(\theta)= \mathcal{L}_{ts}(\theta)+  \sum_{i=1}^{R}   \mathcal{L}_{kd}^i(\theta).
    \label{equ_L}
\end{equation} Based on Eqn.~\ref{equ_proof} and Eqn.~\ref{equ_derivation}, we have Eqn.~\ref{equ_L_derivation} and the derivations. 

\begin{figure}[t]
\centerline{\includegraphics[width=3.4in]{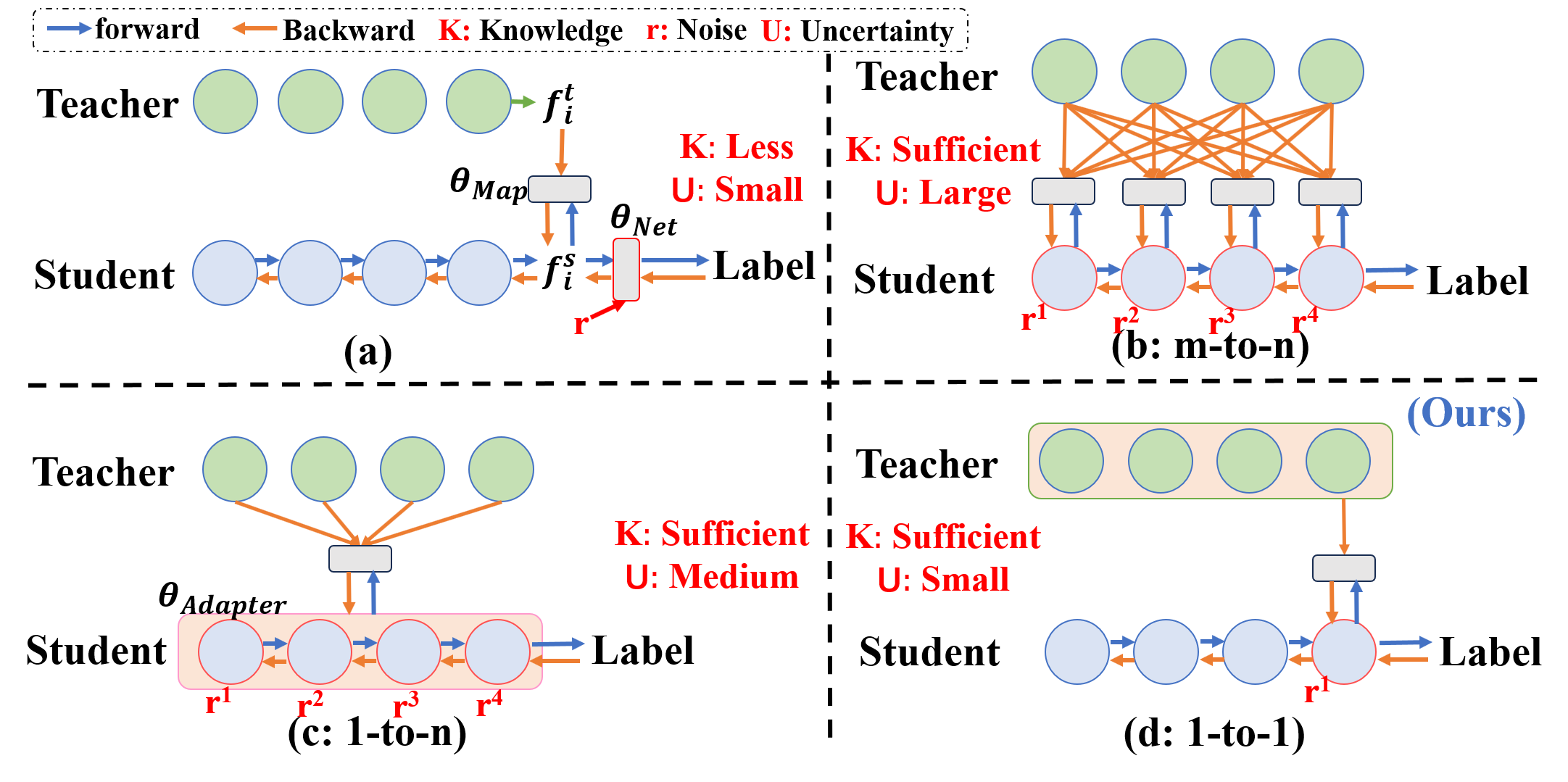}}
\vspace{-3mm}
\caption{{The backpropagation gradient conflict between teacher-student distillation and student-supervised learning.} 'U' and 'K' represent whether the uncertainty is large and the knowledge transfer is sufficient.} 
\label{appen_fig_Adapter}
\vspace{-6pt}
\end{figure}

Now we discuss $m$-to-$n$ (Fig.~\ref{appen_fig_Adapter}(b)) and $1$-to-$n$ (Fig.~\ref{appen_fig_Adapter}(c)) adapter designs that will result in slower convergence than $1$-to-$1$ teacher-side adapter (Fig.~\ref{appen_fig_Adapter}(d)). These adapter designs satisfy $R_{m-to-n}>R_{1-to-n}>R_{1-to-1}$.
\begin{enumerate}
    \item {The convergence complexity. From \textit{Theorem3.1} and Eqn.~\ref{equ_L_derivation}, all loss functions ($\mathcal{L}_{task}(\theta), \{ \mathcal{L}_{kd}^i(\theta)\}_{i}^R$) will convergence when $\left \| g-g_{ts} \right \| $ $ \le \left \| g_{ts} \right \|$ and $\{\| g-g_{kd}^i \| \le  \| g_{kd}^i \| \}_i^R $. If the joint gradient $g$ fails to satisfy any of the $R+1$ inequalities, there will be convergence problems in $\mathcal{L}(\theta)$; consequently, a larger $R$ makes the overall loss function increasingly difficult to converge. Let $C$ denote the optimization complexity, leading to the following relationship:
    \begin{equation}
        C_{m-to-n}>C_{1-to-n}>C_{1-to-1}.
    \end{equation}. Our design could reduce the convergence complexity by having fewer constraints on the gradient magnitudes. }
    
    \item {The update step. From the formula derivation (Eqn. 32 -~\ref{equ_L_derivation}), it can be concluded that the convergence of the loss function $\mathcal{L}(\theta)$ is influenced by the maximum Lipschitz constant $H^{max}$ required for the step size $\alpha$, $0 < \alpha \le \frac{1}{H^{max}}$.}
    
    {From the above conclusion, it can be inferred that as the number of update directions increases, the mutual influence among the gradients of different functions intensifies, resulting in greater difficulty in converging the overall loss function. Therefore, it is evident that that $H^{max}_{m-to-n}>H^{max}_{1-to-n}$ $>H^{max}_{1-to-1}$, and derive the following relation:
    \begin{equation}
        \alpha _{m-to-n}<\alpha _{1-to-n}<\alpha _{1-to-1}.
    \end{equation} Our $1$-to-$1$ teacher-side adapter could fasten the convergence by a larger update step.}

    \item {The student-side parameters. From Fig.~\ref{appen_fig_Adapter}, it is evident that the amount of additional parameters $\theta_{add}$ ($\theta_{Adapter}$, $\theta_{Map}$) involved during the student training satisfies $\theta_{add}^{m-to-n}>\theta_{add}^{1-to-1}$, $\theta_{add}^{1-to-n}>\theta_{add}^{1-to-1}$. Let $D$ denote the student optimization difficulty, leading to the following relationship:    
    \begin{equation}
        D_{m-to-n},D_{1-to-n}>D_{1-to-1}.
    \end{equation} Our design could reduce the student training difficulty by fewer parameters involved in the optimization.}
\end{enumerate}

The above relationship suggests that our $1$-to-$1$ teacher-side adapter achieves faster convergence and reduces training complexity. Further details are as follows:
\begin{align}
&\mathcal{L}(\theta^{p+1})-\mathcal{L}(\theta^{p})\\
\approx& - \alpha  g\cdot \bigtriangledown \mathcal{L}(\theta^{p})+\frac{\bigtriangledown^2 \mathcal{L}(\theta^{p})}{2}(-\alpha\cdot  g)^2\\
&{\small (\left \| \bigtriangledown \mathcal{L}(x )-\bigtriangledown \mathcal{L}(y) \right \| \le (R+1)\cdot H^{max}\cdot\left \| x-y \right \| )}\\
\le & - \alpha  g\cdot \bigtriangledown \mathcal{L}(\theta^{p} )+\frac{(R+1) H^{max}}{2}(-\alpha\cdot  g)^2\\
&{\small (\mathcal{L}(\theta^{p})=\mathcal{L}_{ts}(\theta^p)+  \sum_{i=1}^{R}   \mathcal{L}_{kd}^i(\theta^p), 0<\alpha \le \frac{1}{H^{max}})}\\
& - \sum_{i=1}^{R} \left [ \frac{\alpha}{2}\left ( \left \| \bigtriangledown \mathcal{L}_{kd}^i(\theta^{p}) \right \|^2- \left \|\bigtriangledown \mathcal{L}_{kd}^i(\theta^{p})-g\right \|^2\right ) \right ] \\ 
&{\small (g=g_{ts}+\sum_{i=1}^{R}g_{kd}^i= \bigtriangledown \mathcal{L}(\theta^{p}))}\\
\Rightarrow & \left \| g-g_{ts} \right \| <\left \| g_{ts} \right \| ;\left \| g-g_{kd}^i \right \| <\left \| g_{kd}^i \right \|, i\in [1,R]
\label{equ_L_derivation}
\end{align}  

{Given the difficulty in directly quantifying model uncertainty and convergence stability, efficient and stable convergence is empirically manifested as convergence speed, training loss and generalization capability (test performance). Consequently, we focus on validating these observable metrics in the subsequent experiments.}

\subsection{Intuitive and quantitative illustration of training uncertainty with knowledge distillation}

\begin{figure}[!t]
    \centering
    \includegraphics[width=3.3in]{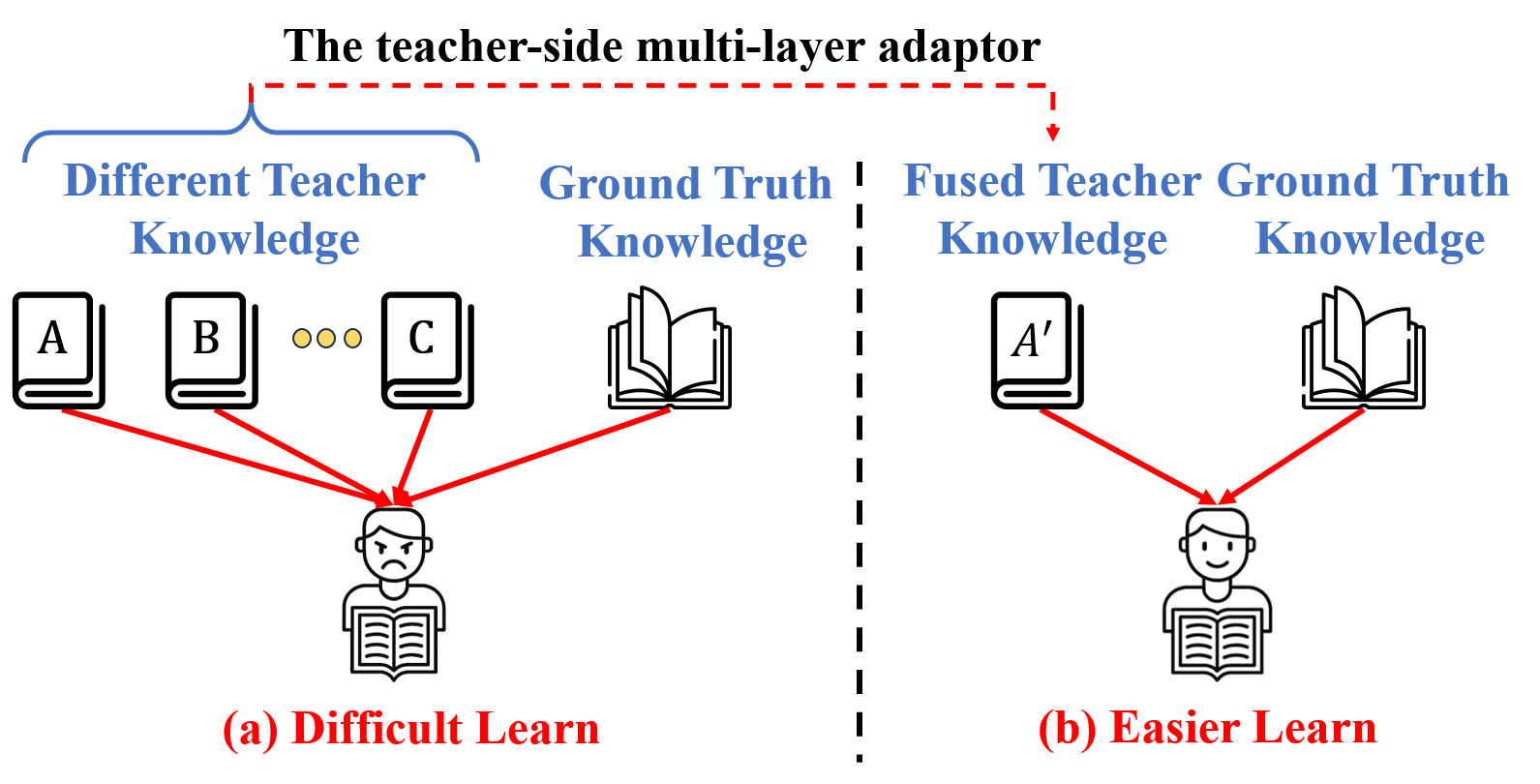}
    \caption{Intuitive visualization about the teacher-side multi-layer adaptor.}
    \label{supp_fig:1}
\end{figure}

To comprehend the uncertainty reduction effect of the teacher-side multi-layer adaptor from an intuitive perspective, we can regard multi-layer teacher features as distinct categories of teacher knowledge (as illustrated in Fig.~\ref{supp_fig:1}). For less-experienced student models, the simultaneous acquisition of diverse teacher-derived knowledge alongside knowledge from ground-truth labels presents inherent challenges. A straightforward and intuitive approach involves leveraging the teacher model's generalization capacity to integrate diverse teacher feature knowledge. This mechanism permits student models to concentrate on assimilating a unified category of fused teacher knowledge. 

We employ a toy optimization example shown in the Fig.~\ref{supp_fig_toy} to better substantiate this claim. This example demonstrates a 3D optimization plot of the student model's task loss (Fig.~\ref{supp_fig_toy} (a)) and the corresponding 2D convergence graphs of specific task loss and distillation loss function combinations. As shown in Fig.~\ref{supp_fig_toy} (d), ill-chosen distillation losses (e.g., excessive distillation strength or misguided distillation direction) prevent student models from converging to optimal solutions. While these results confirm the inherent difficulty in finding optimal distillation loss functions, they also suggest that moderately incorporating the teacher’s generalization capacity benefits student model training (Fig.~\ref{supp_fig_toy} (b), as supported by prior studies~\cite{sun2019patient,huang2022knowledge,li2023curriculum}). Therefore, we aim to enhance the absorption of teacher knowledge while minimizing the optimization challenges posed by distillation loss. This provides another intuitive demonstration of the operational efficacy of the teacher-side multi-layer adaptor.

\begin{table}[ht]
    \centering
    \renewcommand{\arraystretch}{1.18}
    \begin{tabular}{cc|cc}
    \toprule
    \multicolumn{2}{c|}{\textbf{Distillation}} & {\textbf{$\theta_{Net}$}}\\
    Position & Directions & (in distillation uncertainty $U\left ( \theta _{Net},r \right )$)\\
    \hline
    1 & 1 & 24.94M\\
    3 & 1 & 23.24M\\
    1 & 5 & 124.7M ($1$-to-$n$)\\
    3 & 5 & 116.2M ($1$-to-$n$)\\
    5 & 5 & 51.3M ($1$-to-$n$)\\
    1,3,5 & 5 & 292.2M ($m$-to-$n$)\\
    5 & 1 & \textbf{10.26M} (Ours: 1-to-1)\\
    \bottomrule
    \end{tabular}
    \caption{The variations in $\theta_{Net}$ across different distillation combinations.}
    \label{supp_tab:var}
\end{table}

\begin{figure}[ht]
\centerline{\includegraphics[width=3.4in]{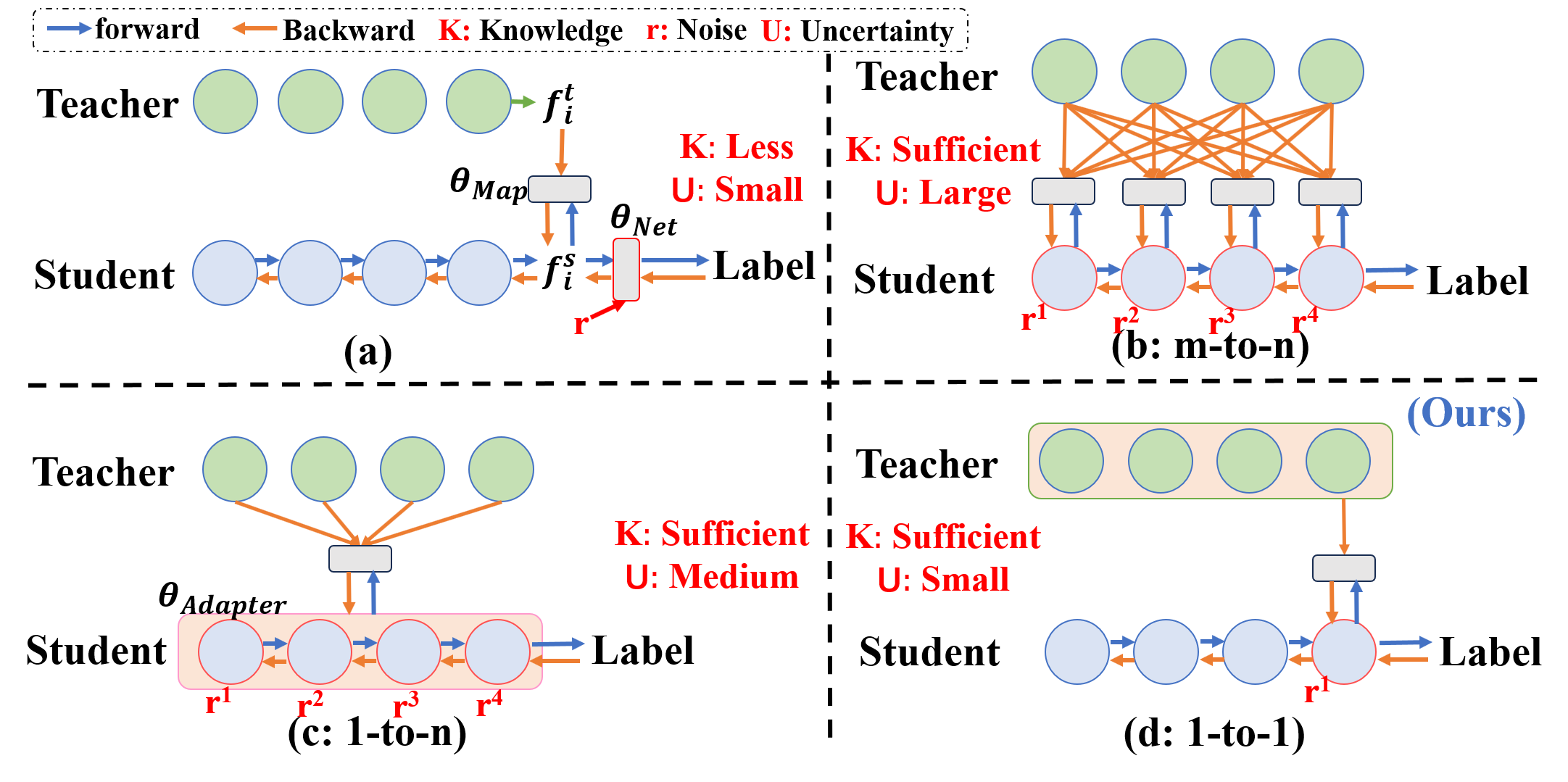}}
\caption{The conflict between teacher-student distillation and student-supervised learning. 'U' and 'K' represent whether the uncertainty is large and the knowledge transfer is sufficient.} 
\label{supp_fig_Adapter}
\end{figure}

As demonstrated above, identifying the optimal distillation loss function for student model training remains non-trivial, with persistent conflicts observed between task-specific loss and distillation loss. To address this, we model such conflicts as training noise governed by Lipschitz continuity properties. Crucially, this noise is amplified through both the positional allocation (where distillation terms are applied) and the quantitative scale (number of distillation directions) within the distillation framework (illustrated in Fig.~\ref{supp_fig_Adapter}). Taking SegNet which comprises 5 blocks as an example (parameters of SegNet is 24.98M), distillation can be implemented in any block. Assuming different distillation methods introduce an equivalent noise level (i.e., $r^1=r^2=\cdots=r^N$),\footnote{For simplicity, we adopt this assumption here. In practice, noise estimation should be determined by how distillation affects the variance of the model's final logits~\cite{zhou2021rethinking}, which will be thoroughly investigated in our future work.} we can estimate the ultimate scaling magnitude induced by the distillation loss by $U(\theta _{Net},r)\propto \left ( \left \| \theta _{Net}^1+\theta _{Net}^2+\cdots+\theta _{Net}^N \right \|  \right ) \cdot r$, where $\theta _{Net}^i$ represents the parameters which amplify the noise caused by distillation loss $\mathcal{L}_{kd}^i $. The variations in $\theta_{Net}$ across different distillation combinations can be approximated in Table~\ref{supp_tab:var}. The distillation position indicates the hierarchical level of features in the student model (1-5 in SegNet student) that require distillation training, while the distillation direction reflects the number of distillation loss terms being trained concurrently. The results demonstrate that our teacher-side multi-layer adaptor introduces the smallest magnitude of uncertainty amplification.

\subsection{Implementation of the toy example}
In the toy example, the training process of the student model is obtained through the smoothed summation of logarithmic and quadratic functions via the $\tanh(\cdot)$ activation. In Fig.~\ref{supp_fig_toy} (a), the specific task loss function for the student model used in the figure is presented below.

\begin{equation*}
\begin{aligned}
    &f_1=\log_{}{\left \{ max(\left |   0.5\!\times \!(x_1+1.5)\!+\!\tanh(x_2)\right |,\;0.00005) \right \} } +6\\
    &f_2=(x_1^2\!+\!0.1\!\times \!(x_2+8)^2)/10-20\\
    &f_{stu} = f_1\times max(\tanh (x_2\!*\!0.5),0)+ f_2\times \\
    & \qquad \qquad \qquad \qquad \qquad \qquad \qquad max(\tanh (-x_2\!*\!0.5),0)\\
\end{aligned}
\end{equation*}

To better demonstrate the effectiveness of the knowledge distillation method, the teacher selects the same convergence function as the Student. Fig.~\ref{supp_fig_toy} (c) demonstrates a distillation direction perfectly aligned with the student model’s optimization path, which can accelerate convergence.

In Fig.~\ref{supp_fig_toy} (d), when the direction of the knowledge distillation loss function aligns with the convergence direction of the student model, but the distillation strength is excessive, the distillation loss function is:

\begin{equation*}
\begin{aligned}
&f_1=\log_{}{\left \{ max(\left |   0.5\!\times \!(x_1+1.5)\!+\!\tanh(x_2)\right |,\;0.00005) \right \} }\times5 +30\\
&f_2=(x_1^2\!+\!0.1\!\times \!(x_2+8)^2)/5-20\\
&f_{tea} = f_1\times max(\tanh (x_2\!*\!2),0)+f_2\times max(\tanh (-x_2\!*\!2),0)\\
\end{aligned}
\end{equation*}.

When the distillation direction conflicts with the student's convergence direction, the distillation loss function is represented in the following:

\begin{equation*}
\begin{aligned}
&f_1=\log_{}{\left \{ max(\left |   0.5\!\times \!(x_1-1.5)\!-\!\tanh(x_2)+2\right |,\;0.00005) \right \} } +6\\
&f_2=((x_1-3)^2\!+\!0.1\!\times \!(x_2+8)^2)/10-20\\
&f_{tea} = f_1\times max(\tanh (x_2\!*\!2),0)+f_2\times max(\tanh (-x_2\!*\!2),0)\\
\end{aligned}
\end{equation*}.

\begin{figure*}[!h]
\centerline{\includegraphics[width=6.5in]{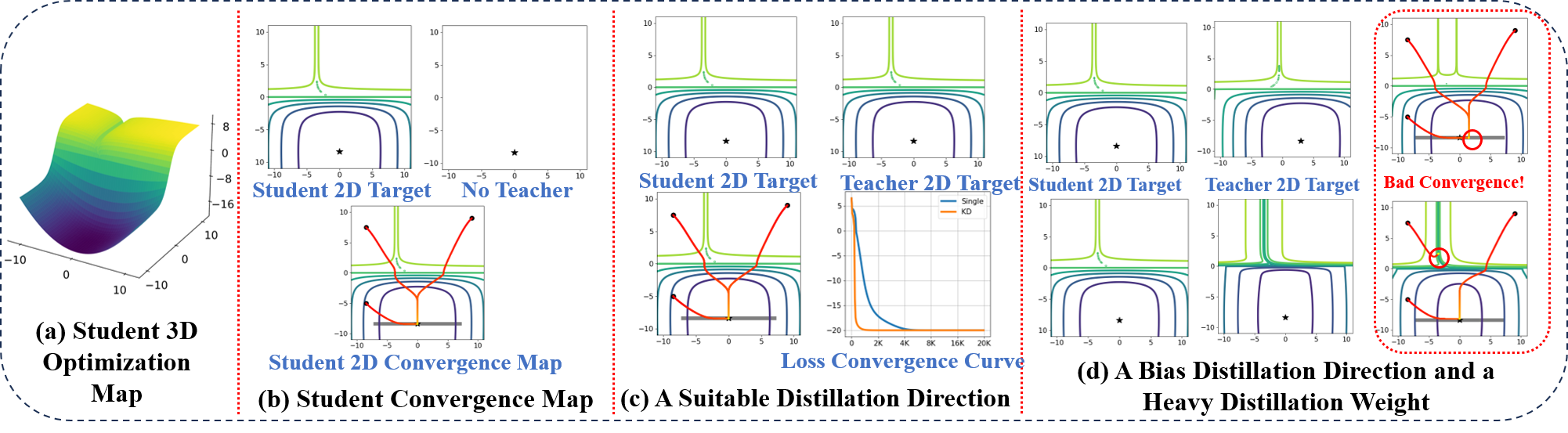}}
\caption{Illustration of why getting a suitable knowledge distillation direction is hard.} 
\label{supp_fig_toy}
\end{figure*}

\subsection{Approach of automatically determine the progressive stages}

Given a required accuracy level, it is necessary to automatically determine the progressive rounds in specific applications. We propose two feasible solutions by first formulating the problem as follows:

\begin{equation*}
\begin{aligned}
\min_{L^{t_k}} \,\, time&(L^{t_k})\\
{\small s.t.\quad  parameter(\theta _{stu})} &{\small \le parameter^{goal}} \\
{\small accuracy(\theta _{stu})} &{\small \ge  accuracy^{goal}}\\
\end{aligned}
\end{equation*}.

The first solution follows prior work~\cite{raychaudhuri2022controllable} by empirically determining the optimal teacher-student co-evolving rounds $L^{t_k}$ under parameter and accuracy constraints within the student model, and records the corresponding results: $\{ parameter^{goal}_i,accuracy_i^{goal},L_i^{t_k} \}_i^{record_{num}}$. Subsequently, these record data are utilized with $parameter^{goal}_i$ and $accuracy_i^{goal}$ as input to output $L_i^{t_k}$. A compact hypernet can be trained to achieve the automated selection of teacher-student co-evolving rounds.

The second automated approach employs large-scale model (LLM)-assisted frameworks, such as GPT-4o~\cite{achiam2023gpt} and Deepseek-R1~\cite{liu2024deepseek}. This method leverages the logical reasoning capabilities of LLMs, employs suitable prompts to align the models with current objectives, and integrates pre-input partial data $\{ parameter^{goal}_i,accuracy_i^{goal},L_i^{t_k} \}_i^{record_{num}}$ to enhance contextual awareness in LLMs.

\subsubsection{Clearer Explanation of Why Feature Map Contains Richer Knowledge} \label{sec:fea_richer}
To clearly explain that intermediate feature layers of deep learning models encapsulate richer knowledge than the final logit outputs, we analyze this phenomenon from the perspective of underdetermined linear systems and null spaces. Considering a standard classification model, the final output is formulated as $p = \theta_{head} \cdot f$. Here, $p \in \mathbb{R}^{P}$ represents the logit vector, where $P$ denotes the number of classes, such as $P=100$ for the CIFAR-100 dataset. The vector $f \in \mathbb{R}^{C}$ denotes the feature map following global average pooling (or $f \in \mathbb{R}^{CHW}$), with $C$ representing the channel dimension, which is typically $C=512$ for a standard ResNet architecture. The weight matrix of the fully connected classification head is denoted by $\theta_{head} \in \mathbb{R}^{P \times C}$. 

Attempting to reconstruct the intermediate feature $f$ from the final prediction $p$ requires solving a linear system comprising $P$ equations and $C$ unknowns. Given that the number of unknowns significantly exceeds the number of equations, the system is strictly underdetermined, implying insufficient information to uniquely identify $f$. Assuming a particular solution $f_{particular}$ exists, the general solution is expressed as: 
$$f = f_{particular} + v,$$ 
where $v \in \mathrm{Null}(\theta_{head})$ and $\mathrm{Null}(\theta_{head})$ represents the null space of the weight matrix containing all vectors satisfying: $$\theta_{head} \cdot v = 0.$$ 

According to the rank-nullity theorem, the dimension of this null space is at least $C - P$, equating to a 412-dimensional subspace in this specific configuration. Consequently, a single logit output $p$ corresponds to an infinitely large set of distinct features residing within this expansive subspace. Because identifying the exact original feature from this infinite set of possibilities is mathematically impossible, the projection process is fundamentally irreversible. A similar information discarding mechanism occurs during the downsampling operation preceding the final output. The limited capacity of compact student models constrains their ability to assimilate the highly abstracted knowledge produced by the teacher network. As evidenced by the quantitative results, the notable reduction in the intrinsic dimension of the model during the final stages indicates that the network deliberately discards specific dimensional information within the feature space to achieve a more concentrated and task-specific representation (Table~\ref{tab:segnet_layerwise_id}).

\subsection{Formal definitions of training instability with knowledge distillation}

To further substantiate the strength of our method in addressing multi-layer feature inconsistencies and instability, we first emphasize the prerequisite of thorough knowledge distillation: the student model should acquire knowledge as comprehensively as possible from all hierarchical features of a teacher model containing multi-layer representations (e.g., a full-size SegNet teacher model with 5 learnable feature layers partitioned by blocks), rather than selectively learning from isolated layers. The student model's learning of teacher features across different layers can be conceptualized as distinct distillation directions. Next, we will proceed to elucidate and quantitatively characterize training instability and multi-layer feature discrepancies.

As visualized in our toy example, the parameter space can be conceptualized as a topological landscape where the loss function $\mathcal{L}(\theta)$ represents the potential energy. In physical terms, force is defined as the negative gradient of potential energy, $F = -\nabla U$. Consequently, the first-order derivative (gradient) of the loss function acts as a force exerted on the model parameters, propelling them in the direction of the steepest descent in energy. The gradient norm $\lVert g \rVert$ signifies the magnitude of this force. A substantial intrinsic gap between the teacher and student models engenders a highly steep potential energy surface, resulting in excessively large gradients. Physically, this is analogous to an object subjected to an overwhelming force, inducing massive acceleration. Under a discrete optimization step size (learning rate), this excessive force causes the parameters to overshoot the optimal trajectory or oscillate violently across the valley walls. This phenomenon represents the direct physical manifestation of training instability. Furthermore, the second-order derivative, or the Hessian matrix $\nabla^2 \mathcal{L}$, describes the curvature of this landscape, indicating the narrowness of the valley and the overall difficulty of the optimization process. A larger gap between the teacher and student correlates with a narrower valley, where an identical force induces a more severe deviation from the optimal path.

Fundamentally, teacher-student distillation stability is dictated by gradient discrepancies~\cite{zhu2021student}. As demonstrated in prior literature (toy example in our manuscript), variations in teacher feature representations and model capacities yield distillation gradients with disparate magnitudes and directions. Drawing upon insights from multi-task learning research~\cite{yu2020gradient,liu2021conflict,liu2023famo}, ensuring distillation stability requires that the gradients derived from the distillation objective do not conflict with the gradients originating from the supervised ground-truth loss. Instead, the distillation gradients should constructively align with and accelerate the convergence of the supervised learning task.

Initially, we define the curvature of the student model updates during the distillation process as follows:
\begin{align}
    \mathbb{H} (\mathcal{L}; & \theta^p,\theta^{p+1}) = \\ 
    &\int_{0}^{1} \nabla \mathcal{L}(\theta^p)^T \nabla^2 \mathcal{L}(\theta^p+x\cdot (\theta^{p+1}-\theta^p)) \nabla \mathcal{L}(\theta^p) dx \\
    &= \int_{0}^{1} \nabla \mathcal{L}(\theta^p)^T \nabla^2 \mathcal{L}(\theta^p+x\eta \cdot \nabla \mathcal{L}(\theta^p)) \nabla \mathcal{L}(\theta^p) dx \cdot \eta^2 \\
    & \quad \text{(Assuming the update step $\eta$ is sufficiently small)} \\
    &\approx \eta^2 \nabla \mathcal{L}(\theta^p)^T \nabla^2 \mathcal{L}(\theta^p) \nabla \mathcal{L}(\theta^p) \\
    & \quad \text{(Second-order Taylor approximation)} \\
    &\approx 2 \cdot \left[ \mathcal{L}(\theta^{p+1}) - \mathcal{L}(\theta^p) - \nabla \mathcal{L}(\theta^p)^T (\theta^{p+1}-\theta^p) \right],
\end{align}
where the parameter update is given by $\theta^{p+1}-\theta^p = \eta \cdot (\nabla \mathcal{L}_{ts}(\theta^p) + \nabla \mathcal{L}_{kd}(\theta^p))$, $\eta$ denotes the learning rate, and $\theta^p$ represents the student parameters at iteration $p$. As established in existing literature, a smooth curvature is indicative of a stable neural network training process. Conversely, if the curvature of the combined loss landscape, comprising the distillation loss $\mathcal{L}_{kd}(\theta^p)$ and the supervised task loss $\mathcal{L}_{ts}(\theta^p)$, is excessively sharp, and their respective gradient directions conflict, applying a large learning rate forces the student optimization to disproportionately favor the dominant objective. For instance, as illustrated in Fig.5 (d) in our manuscript, if the distillation loss is heavily weighted and its gradient opposes the supervised gradient, the student model may overfit to the teacher representations while neglecting the ground-truth signals. Under such high-curvature conditions, the summation of these conflicting losses induces severe oscillatory updates. To precisely quantify the training instability arising from the confluence of sharp curvature and misaligned gradients, we define an instability metric based on the product of the normalized curvature and a gradient alignment penalty:
\begin{align}
    S_{\mathrm{instability}}
    &= \sum_{x_i \in \mathcal{D}} \left[ 1 - \cos\!\left(\nabla_{\theta^p_{Net}} \mathcal{L}_{task},\; \nabla_{\theta^p_{Net}} \mathcal{L}_{kd}\right) \right] \notag\\
    &\qquad \qquad\qquad\cdot \mathcal{N}\left( \mathbb{H} (\mathcal{L};\theta^p_{Net},\theta^{p+1}_{Net}) \right),
    \label{eqn:instable}
\end{align}
where $\mathcal{N}(\cdot)$ denotes a normalization function. The gradient inconsistency is measured using cosine similarity, consistent with established methodologies. The behavior of this metric is mathematically robust: 
\begin{itemize}
    \item When $\nabla \mathcal{L}_{ts}$ and $\nabla \mathcal{L}_{kd}$ are perfectly aligned, $\cos(\cdot) = 1$. The penalty term $[1 - 1]$ becomes zero. Consequently, regardless of the magnitude of the curvature $\mathbb{H}$, the synergistic forces prevent oscillatory conflict, yielding $S_{\mathrm{instability}} = 0$.
    \item When the gradients are orthogonal, $\cos(\cdot) = 0$. The penalty multiplier is $1$, meaning the instability is dictated solely by the local curvature $\mathbb{H}$.
    \item When the gradients are diametrically opposed, $\cos(\cdot) = -1$. The penalty term evaluates to $[1 - (-1)] = 2$. This acts as a severe amplification mechanism: in regions of extreme gradient conflict, the instability induced by the curvature $\mathbb{H}$ is doubled, accurately modeling the physical phenomenon where the optimization process is highly susceptible to collapse.
\end{itemize}
Similarly, by employing the Gauss-Newton approximation to estimate the curvature, we derive the following relation:
\begin{align}
    \mathbb{H} (\mathcal{L}; & \theta^p,\theta^{p+1}) = \\
    & \int_{0}^{1} \nabla \mathcal{L}(\theta^p)^T \nabla^2 \mathcal{L}(\theta^p+x\cdot (\theta^{p+1}-\theta^p)) \nabla \mathcal{L}(\theta^p) dx \\
    & \quad \text{(Gauss-Newton approximation)} \\
    &\approx \nabla \mathcal{L}(\theta^p)^T \left[ \nabla \mathcal{L}(\theta^{p+1}) \cdot \nabla \mathcal{L}(\theta^{p+1})^T \right] \nabla \mathcal{L}(\theta^p).
\end{align}
A massive representational gap between the teacher and student models implies a profound divergence between the distillation gradient $\nabla \mathcal{L}_{kd}(\theta^p)$ and the task gradient $\nabla \mathcal{L}_{ts}(\theta^p)$. This divergence subsequently causes a significant disparity between the total consecutive gradients $\nabla \mathcal{L}(\theta^p)$ and $\nabla \mathcal{L}(\theta^{p+1})$, which precludes the formation of a smooth curvature $\mathbb{H}(\mathcal{L};\theta^p,\theta^{p+1})$ and fundamentally destabilizes the training process. Therefore, while the teacher knowledge must strictly guide the student, the representational gap between them must be carefully bounded. Conventional methods largely neglect this critical constraint. To resolve this, we formalize the "progressively stronger teacher" paradigm, ensuring that the complexity of the distilled knowledge escalates gradually to maintain optimization stability throughout the training trajectory.

We refrain from directly utilizing the raw curvature metric. As indicated by the approximation formula:
$$\mathbb{H} \approx \left( \nabla \mathcal{L}(\theta^p)^T \nabla \mathcal{L}(\theta^{p+1}) \right)^2,$$
the magnitude of the calculated curvature is intrinsically coupled with the scale of the loss gradients. Consequently, larger loss values naturally generate greater gradients, which this specific formulation amplifies to a quadratic scale. Therefore, we define the normalized curvature $\mathcal{N}(\mathbb{H})$ as:
$$\mathcal{N}(\mathbb{H}) = \left( \tilde{\nabla} \mathcal{L}(\theta^p)^T \tilde{\nabla} \mathcal{L}(\theta^{p+1}) \right)^2, where \quad \tilde{\nabla} \mathcal{L} = \frac{\nabla \mathcal{L}}{\mathcal{L}+1}.$$

\begin{figure}[!t]
\vspace{-2mm}
\centerline{\includegraphics[width=3.3in]{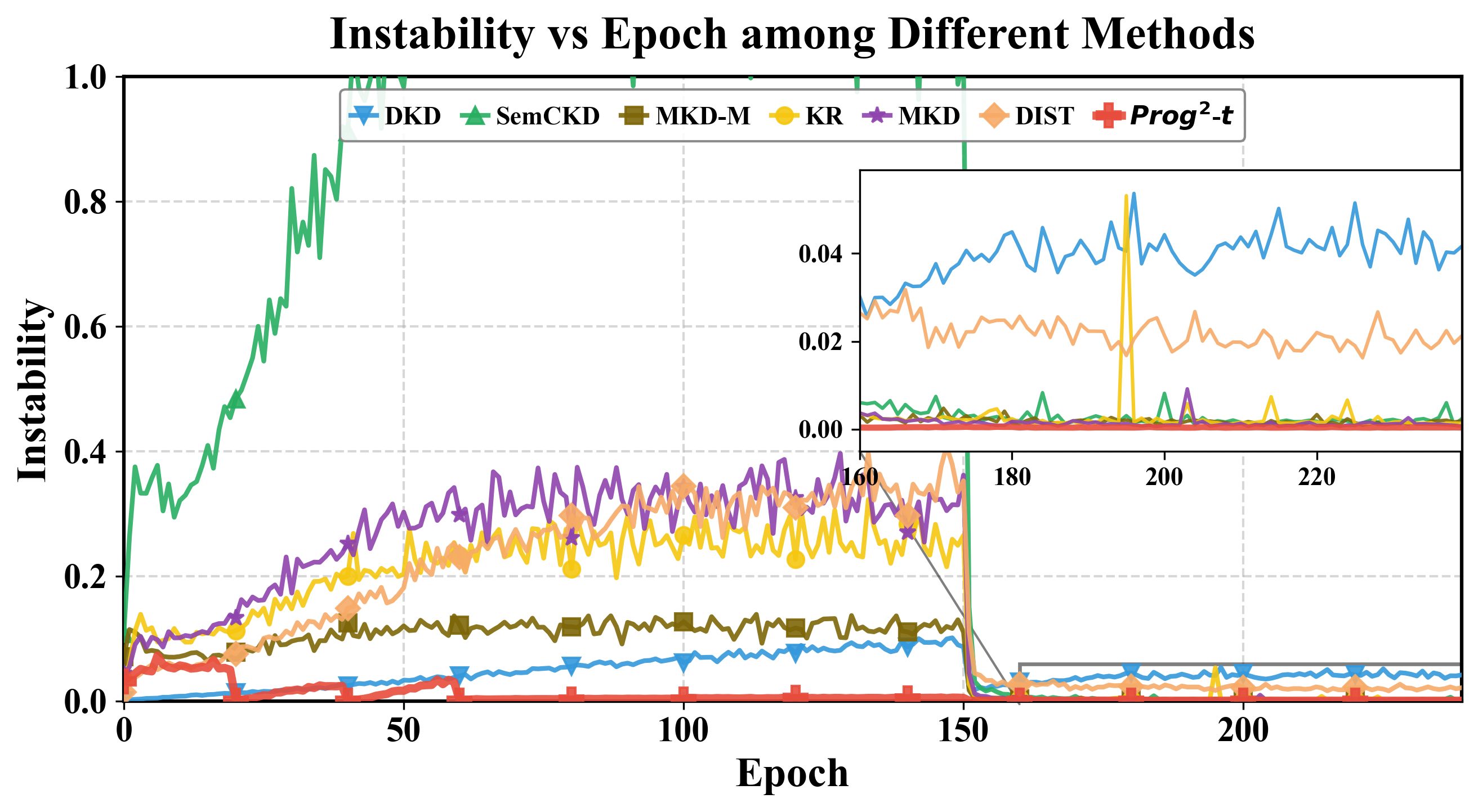}}
\vspace{-3mm}
\caption{Training instability on the CIFAR-100 dataset under an experimental setup consistent with Table \uppercase\expandafter{\romannumeral2} in our manuscript, where the instability is quantified using the previously formulated metric~\ref{eqn:instable}. A learning rate decay of 0.1 was applied at the $150_{th}$ round to ensure convergence. 'MKD-M' represents distillation more student layers than 'MKD'.}
\label{res:fig_instability}
\end{figure}

$\mathcal{L}+1$ prevents the denominator from being too small, which would result in meaningless values. The corresponding Fig.~\ref{res:fig_instability} visualizes the magnitude of instability throughout the training process of the single-stage model. The results demonstrate that multi-layer feature distillation (MKD, MKD-M, KR, SemCKD)inherently amplifies the optimization uncertainty during training. Conversely, by modifying the intrinsic dimensions of the teacher features (Fig.~\ref{fig:tea_stu_ID}) to systematically adjust the learning difficulty, our method consistently maintains a low level of instability. This empirical observation strictly aligns with the conclusions derived from our theoretical proxy analysis based on Lipschitz continuity.

\begin{table}[!t]
\begin{center}
\renewcommand{\arraystretch}{1.3}
\begin{tabular}{cc|c}
\toprule
\multirow{2}{*}{\textbf{ Methods}} & \multirow{2}{*}{\textbf{Distill Layers}} & \textbf{Training Instability}\\
 & & (Result Transition)\\
\hline
DIST & logits (soft label) & 54 $\to$ 11\\
ESKD & single-layer & 29 $\to$ 9\\
SemCKD & multi-layer & 130 $\to$ 13\\
Ours: $Prog^2$-$t$ & single to multi-layer & \textbf{10 $\to$ 7}\\
\bottomrule
\end{tabular}
\caption{Training instability on Cityscapes dataset with SegNet backbones.}
\label{app_tab:instable}
\end{center}
\end{table}

\subsection{Implementation Details} 

On the Cityscapes dataset, we used Adam optimization with a learning rate of 1e-4 for the student, teacher (for our $Prog^2$-$ts$ setting), and the teacher side adaptor. The single-layer convolution for teacher-student alignment in terms of channel dimension follows prior work\cite{li2020knowledge} with a learning rate of 0.1. We train the model with 200 epochs (600 epochs for multi-stage setting), with no learning rate decay, the batch size is set at 8. 

For Cifar-100, we used SGD optimization with a learning rate of 0.05 for the student and the teacher side adaptor, with the Nesterov momentum of 0.9 for 240 epochs and a learning rate decay of 0.1 for 150, 180, and 210 epochs (followed by~\cite{tiancontrastive}), and with the batch size of 128. A learning rate of 0.1 is used for optimizing the single-layer convolution for the distillation dimension alignment model. 

On the NYU-V2 dataset (for depth estimation task), the batch size was set to 4. The student model was optimized using Stochastic Gradient Descent (SGD) with a learning rate of $0.02$, momentum of $0.9$, and weight decay of $1e-4$. In the multi-stage distillation setting, the teacher model and the teacher-side adapter utilized the same SGD optimizer configurations (learning rate: $0.02$, momentum: $0.9$, weight decay: $1e-4$). For the single-layer convolution used in teacher-student feature alignment, we employed the Adam optimizer with a learning rate of $0.1$, consistent with the settings used for the previous datasets.

On the Tiny-ImageNet-200 dataset, the learning rate for both the student model and the teacher-side adapter was set to $0.1$. All other hyperparameters and configurations remained identical to those used for the CIFAR-100 dataset.

Across all experiments, baseline methods were reproduced within our training pipeline based on their official open-source implementations. To ensure fair comparison, all hyperparameters and weights for these competing methods were strictly maintained according to their original open-source configurations.

The experiments are trained on 6 GPUS (i.e.NVIDIARTX3090). We performed all baseline measurements in isolation on a single GPU to accurately determine training times on all the datasets.

we now summarize the dataset splits, model architectures, optimizer settings, learning rates, batch sizes, training epochs, number of independent runs, random-seed protocol, baseline implementation protocol, hardware environment, software environment, and timing-measurement conditions. We also clarify that the reported mean and standard deviation values are computed over three independent runs. Since no additional statistical significance results beyond the reported mean and standard deviation are available from the provided experimental records, we do not claim additional statistical significance results. The corresponding reproducibility details are summarized in the following tables (Table~\ref{tab:dataset_settings}, Table~\ref{tab:general_protocol}, and Table~\ref{tab:ts_settings})

\subsection{More Results on Distillation Stability}
Specifically, we employ SegNet as the backbone of the Cityscapes dataset, adjusting the channel numbers of shared modules to the compression levels ($p$) defined in Algorithm 1 for cases when $p$ is set to $25\%$ (14.06M), $50\%$ (6.25M) and $75\%$ (1.57M). These student models have undergone 44\%(\textbf{large student}), 75\%(\textbf{moderate student}), and 94\%(\textbf{small student}) compression rates in comparison to the teacher model, which measures 24.98M in size. The remaining experimental settings are consistent with those described in the paper.

\begin{figure}[!t]
\vspace{-6pt}
\centerline{\includegraphics[width=3.in]{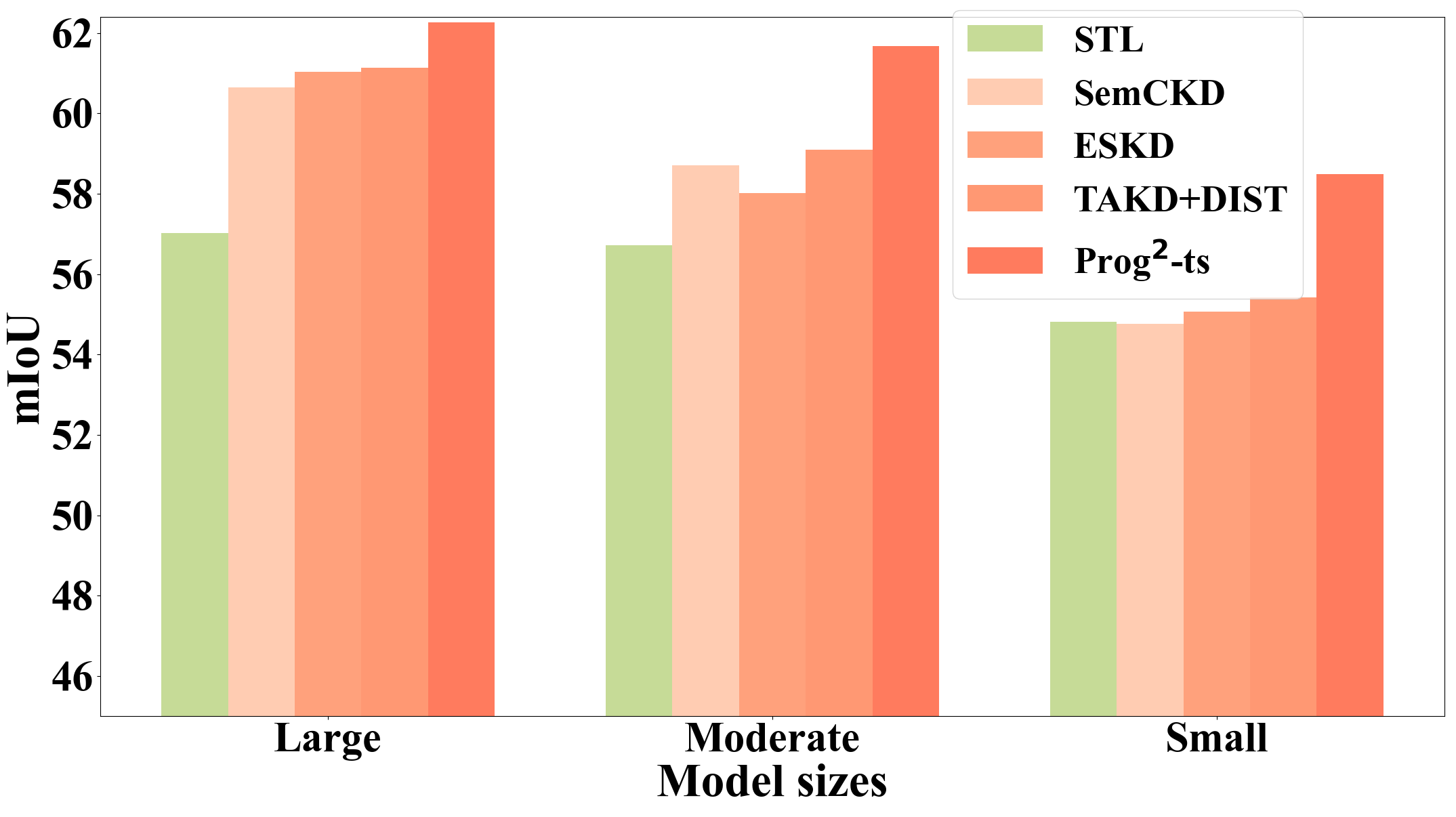}}
\vspace{-10pt}
\caption{The standardized mIoU results of models under various compression rates with different distillation methods are compared. We employ SegNet as the backbone of the Cityscapes dataset, adjusting the channel numbers of shared modules to the compression levels ($p$) defined in Algorithm 1 for cases when $p$ is set to $25\%$ (14.06M), $50\%$ (6.25M) and $75\%$ (1.57M). These student models have undergone 44\%(large student), 75\%(moderate student), and 94\%(small student) compression rates in comparison to the teacher model, which measures 24.98M in size.} 
\label{fig_diff_compress}
\end{figure}

It can be seen from figure ~\ref{fig_diff_compress} that as the model is continuously compressed, various types of distillation methods (SemCKD~\cite{chen2021cross}, ESKD~\cite{cho2019efficacy}, TAKD~\cite{mirzadeh2020improved}+DIST~\cite{huang2022knowledge}) are approaching failure (they are represented in the figure as a comparison to the single-task method(STL)). However, our progressive distillation method greatly ensures the effectiveness and stability of the model, which gradually manifests with the increase of compression rate.

\subsection{Ablation study of various learning rates of the teacher-side multi-layer adaptor}

In Table~\ref{app_tab:lr_adaptor}, an extended ablation study was conducted to analyze the impact of learning rates for the teacher-side multi-layer adaptor. Experimental results demonstrate that employing either excessively high (learning rate=1e-1) or low (learning rate=1e-6) learning rates consistently yields suboptimal performance. Therefore, to maintain consistency with the student model's training protocol, we ultimately set the learning rate of the teacher-side multi-layer adaptor to 1e-4.

\begin{table}[!t]
    \centering
    \renewcommand{\arraystretch}{1.3}
    \begin{tabular}{c|c|cc}
    \toprule
        \textbf{Method} & \textbf{Learning Rate} &\textbf{mIoU} & \textbf{Pixel Acc}\\
         \hline
         \multirow{6}{*}{$Prog^2$-$t$}& 1e-1 & 0.5904 & 0.9321\\
         & 1e-2 & 0.5956 & 0.9327\\
         & 1e-3 & 0.5971 & 0.9331\\
         & 1e-4 & \textbf{0.5977} & 0.9337\\
         & 1e-5 & 0.5976 & \textbf{0.9342}\\
         & 1e-6 & 0.5952 & 0.9323\\
    \bottomrule
    \end{tabular}
    \caption{Results among different learning rates of the teacher-side multi-layer adaptor, experiments conducted based on $Prog^2$-$t$ method on Cityscapes dataset.}
    \label{app_tab:lr_adaptor}
\end{table}

\begin{table}[!t]
\begin{center}
\renewcommand{\arraystretch}{1.3}
\begin{tabular}{cc|cc}
\toprule
\textbf{Methods} & \textbf{Distill Layers} & \textbf{Top-1 Acc} & \textbf{Top-5 Acc}\\
\cline{1-4}
Full Teacher & - & 57.30 & 80.43\\
\cline{1-4}
STL-p & - & 47.47 & 74.14\\
MGD & single-layer & 47.86 & 74.24\\
ESKD & single-layer & 48.20 & 74.71\\
SemCKD & multi-layer & 48.32 & 74.73\\
KR & multi-layer & \underline{48.66} & \underline{74.91}\\
DIST & logits (soft label) & 48.37 & 74.76\\
DKD & logits (soft label) & 48.22 & 74.27\\
Ours: $Prog^2$-$t$ & single to multi-layer & \textbf{49.75} & \textbf{75.64}\\
\cline{1-4}
TAKD + DIST & logits (soft label) & 49.18 & 75.55\\
Ours: $Prog^2$-$ts$ & single to multi-layer & \textbf{51.25} & \textbf{76.81}\\
\bottomrule
\end{tabular}
\caption{Classification Testing Result on Tiny-Imagenet-200 Dataset (\%).} 
\label{tab_imagenet}
\end{center}
\end{table}

\begin{table}[!t]
\begin{center}
\renewcommand{\arraystretch}{1.3}
\begin{tabular}{cc|cc}
\toprule
\textbf{Methods} & \textbf{Distill Layers}$\quad$ & $\quad$\textbf{mIoU}$\quad$ & \textbf{Pixel-acc}\\
\cline{1-4}
Full Teacher & - & 0.8312 & 0.9565\\
\cline{1-4}
STL-p & - & 0.8017 & 0.9487\\
MGD & single-layer & 0.7887 & 0.9478\\
ESKD & single-layer & \underline{0.8135} & \underline{0.9523}\\
SemCKD & multi-layer & 0.8132 & 0.9520\\
KR & multi-layer & 0.7754 & 0.9369\\
DIST & logits (soft label) & 0.8101 & 0.9512\\
Ours:$Prog^2$-$t$ & single to multi-layer & \textbf{0.8173} & \textbf{0.9528}\\
\cline{1-4}
TAKD+DIST & logits (soft label) & 0.8157 & 0.9526\\
Ours:$Prog^2$-$ts$ & single to multi-layer & \textbf{0.8242} & \textbf{0.9555}\\
\bottomrule
\end{tabular}
\caption{Semantic segmentation with Segformer backbone results on Cityscapes dataset.}
\label{tab:segformer}
\end{center}
\vspace{-10pt}
\end{table}

\subsection{Results of Transformer-based Models}
We utilize SegFormer~\cite{xie2021segformer}, an enhanced adaptation of the Vision Transformer (ViT) framework on the Cityscapes dataset to validate the scalability of our distillation method further. We use pretrained SegFormer with Mit-B1 backbone (13.68M) as the teacher model and SegFormer with Mit-B0 backbone (3.72M) as the student model. The results are showed on Table~\ref{tab:segformer}.

\subsection{Implementation of image classification on Tiny-Imagenet-200 dataset}  We further investigate the comparative results of various methods on Tiny-ImageNet 200 datasets (Table~\ref{tab_imagenet}). Our $Prog^2$-$t$ still achieves optimal results.

\begin{table}[!t]
    \centering
    \renewcommand{\arraystretch}{1.3}
    \begin{tabular}{c|cccccc}
    \toprule
        \multirow{2}{*}{\textbf{Method}} & \textbf{MRPC} & \textbf{RTE} & \textbf{SST-2} & \textbf{QNLI} & \textbf{MNLI} & \textbf{QQP}\\
        & \textbf{(3.7k)} & (2.5k) & (67k) & (105k) & (393k) & (364k) \\
        \cline{1-7}
        BERT$_{Base}$ & 88.70 & 68.95 & 91.97 & 91.32 & 83.18 & 90.91\\
        \cline{1-7}
        BERT$_{Small}$ & 81.12 & 64.26 & 87.84 & 86.51 & 77.02 & 89.01\\
        VanillaKD & 83.82 & 64.98 & 88.76 & 87.20 & 77.69 & 89.23\\
        PKD & 84.31 & 64.98 & 89.68 & 87.77 & 80.09 & 90.12\\
        ESKD & 83.91 & 65.24 & 88.71 & 88.10 & 77.77 & 88.89\\
        Ours: $Prog^2$-$t$ & \textbf{85.05} & \textbf{66.43} & \textbf{90.73} & \textbf{88.55} & \textbf{81.97} & \textbf{90.89}\\
        \bottomrule
    \end{tabular}
    \caption{Accuracy results of BERT-Based models on GLUE benchmark.}
    \label{app_tab:bert}
\end{table}

\subsection{Implementation of depth estimation on NYU-V2 dataset}
In this experiment (Table ~\ref{tab_nyuv2}), the parameter sizes of the teacher and student models are 24.98M and 6.25M, respectively. Based on the experimental results of the depth estimation task on the NYU-V2 dataset shown in Table ~\ref{tab_nyuv2}, our method $Prog^2$-$t$ achieved a reduction of 1.6\% in absolute error and 0.2\% in relative error compared to the ESKD~\cite{cho2019efficacy} method. The $Prog^2$-$ts$ method achieved a reduction of 4.6\% in absolute error and 1.5\% in relative error compared to the second-best result.
\begin{table}[!t]
\begin{center}
\renewcommand{\arraystretch}{1.3}
\begin{tabular}{cc|cc}
\toprule
\textbf{Methods} & \textbf{Distill layers}$\quad$ & $\quad$\textbf{Abs Err}$\quad$ & \textbf{Rel Err}\\
\cline{1-4}
Full Teacher & - & 0.4920 & 0.2110\\
\cline{1-4}
STL-p & - & 0.6008 & 0.2503\\
MGD & single-layer & 0.5397 & 0.2196\\
ESKD & single-layer & \underline{0.5355} & \underline{0.2126}\\
SemCKD & multi-layer & 0.5576 & 0.2305\\
KR & multi-layer & 0.5878 & 0.2462\\
DIST & logits (soft label) & 0.5543 & 0.2377\\
Ours:$Prog^2$-$t$ & single to multi-layer & \textbf{0.5198} & \textbf{0.2146}\\
\cline{1-4}
TAKD+DIST & logits (soft label) & 0.5265 & 0.2163\\
Ours:$Prog^2$-$ts$ & single to multi-layer & \textbf{0.4810} & \textbf{0.1997}\\
\bottomrule
\end{tabular}
\caption{Depth Estimation Testing Result on NYU-V2 Dataset.}
\label{tab_nyuv2}
\end{center}
\end{table}

\subsection{Implementation of BERT-based models on GLUE benchmark}
We analyze the effectiveness of knowledge distillation in BERT-based methods~\cite{devlin2018bert,sanh2019distilbert,jiao2020tinybert} within the natural language processing (NLP) task. These experiments build upon foundational work~\cite{sun2019patient} examining previous applications of knowledge distillation to BERT models. We conducted experiments on the GLUE benchmark using six datasets (MRPC, RTE, SST-2, QNLI, MNLI, QQP). Dataset details are as follows:

\begin{itemize}
    \item MRPC (Microsoft Research Paraphrase Corpus): A semantic equivalence task containing sentence pairs automatically extracted from online news sources and manually annotated for paraphrase identification.
    \item RTE (Recognizing Textual Entailment): A natural language inference task combining datasets from annual textual entailment challenges, with samples constructed from news articles and Wikipedia.
    \item SST-2 (Semantic Textual Similarity Benchmark): A similarity/paraphrase task aggregating sentence pairs from news headlines, video captions, image captions, and natural language inference data.
    \item QNLI (Question-answering NLI): A natural language inference task.
    \item MNLI (Multi-Genre Natural Language Inference): A crowdsourced NLI dataset with annotations for textual entailment across diverse genres.
    \item QQP (Quora Question Pairs): A similarity/paraphrase task comprising question pairs from the Quora platform.
\end{itemize}

We employed BERT-base-uncased~\cite{devlin2018bert} as the large teacher model (contains 12 features, $5_{th},7_{th}$, and $9_{th}$ features are used) and BERT-small-uncased model (contains 4 features) as the compact student model. Since the features of both student and teacher models in natural language processing (NLP) tasks are two-dimensional, our progressive method avoids attention fusion between feature maps and instead focuses on feature vector attention fusion. Thus, our progressive framework employs the Multi-Head Attention mechanism from~\cite{vaswani2017attention} as the feature fusion strategy. All other algorithmic components align with those described in the manuscript. For experimental implementation, key parameters: learning rate, batch size, model optimizers, distillation weights, and max sequence length, were all configured according to established practices in prior studies~\cite{sun2019patient}. We report accuracy scores for all baselines, with MNLI results showing matched accuracy. The experimental outcomes are delineated in Table~\ref{app_tab:bert}. 

Our progressive teacher-side feature fusion distillation approach $Prog^2$-$t$ has achieved notable effectiveness in NLP tasks, underscoring the robust scalability of knowledge distillation techniques. 

\begin{table}[!t]
\begin{center}
\renewcommand{\arraystretch}{1.3}
\begin{tabular}{c|c|cc}
\toprule
\textbf{ Model } & \textbf{Methods} & \textbf{ Abs Err } & \textbf{ Rel Err }\\
\cline{1-4}
\textit{Teacher} & ResNet Depth STL & 0.0161 & 17.42\\
\textit{Student} & Semantic Pruned STL & 58.00 & 92.88\\
\hline
\multirow{3}{*}{\textit{Cross-task}} & MKD & 56.55 & 92.46\\
& CKD & 57.74 & 92.90\\
& Ours: $Prog^2$-$t$ & \textbf{59.95} & \textbf{93.38}\\
\bottomrule
\end{tabular}
\captionsetup{font=small}
\caption{Testing Result on Cityscapes Dataset. Knowledge transfer from depth task to semantic task, from ResNet101 trained on NYU-V2 dataset model to SegNet model on Cityscapes dataset. The MKD method is from~\cite{li2020knowledge}, and the CKD method~\cite{li2022learning} is a widely used cross-task distillation method.}
\label{app_tab:cross}
\end{center}
\end{table}

\subsection{Cross-domain results of our methods}
We utilize a ResNet101-based teacher model (designed for depth estimation tasks) to distill knowledge into a Pruned SegNet-based student model (targeting semantic segmentation tasks), thereby validating the effectiveness of cross-architecture knowledge distillation. Table~\ref{app_tab:cross} shows that our teacher-side adapter naturally maps teacher features and student features to the same distribution, facilitating its easy extension to cross-domain or cross-task scenarios. Our $Prog^2$-$t$ method exhibits architecture-agnostic effectiveness, maintaining robust performance regardless of model architectures, even those custom-tailored for specialized tasks, thus evidencing remarkable generalizability. This is something that previous compression methods, such as model quantization and low-rank decomposition, were unable to achieve.

\subsection{Stability and convergence of the $Prog^2$-$ts$ method}
We acknowledge that our work does not provide a formal proof for the stability and convergence of the progressive compression distillation used in the Prog2-ts method. However, we first need to clarify the motivation for this design. Prior work has already demonstrated that when a significant capacity gap exists between the teacher and student models~\cite{mirzadeh2020improved,huang2022knowledge,stanton2021does}, direct knowledge distillation can actually degrade the student's performance. This is because the representations learned by an overly capable teacher are at a much higher level of abstraction than the student can handle. Consequently, its soft labels become overconfident~\cite{huh2024platonic}, which prevents the student from learning the valuable information from these labels that extends beyond the ground-truth data.

Subsequently, many studies have proposed using intermediate-sized "teacher assistants" to mitigate the large capacity gap in knowledge distillation~\cite{mirzadeh2020improved,son2021densely,liu2025monotakd}. This serves as the primary motivation for our multi-stage progressive distillation approach. We want to clarify that proving the convergence and stability for our multi-stage method $Prog^2$-$ts$ is different from the proof for a single-stage process $Prog^2$-$t$. Due to space constraints, this was not included in the main paper, but we will provide the detailed derivation in the appendix file. The purpose of this proof is to formally establish that our multi-stage progressive distillation method $Prog^2$-$ts$ achieves a smaller error bound and exhibits more stable convergence than single-stage knowledge distillation methods (e.g. the VanillaKD). This provides the theoretical rationale for its adoption in our work. The proof proceeds as follows:

According to the VC theory~\cite{vapnik1999overview} one can decompose the classification error of a classifier $f_s$ as
\begin{equation}
    E(f_s) - E(f_g) \leq O\left( \frac{|\mathcal{F}_s|_\theta}{n^{\beta_{sg}}} \right) + \epsilon_{sg},
    \label{eqn_est}
\end{equation} 
where, the $O(\cdot)$ and $\epsilon_{sg}$ terms are the estimation and approximation error, respectively. The former is related to the statistical procedure for learning given the number of data points, while the latter is characterized by the capacity of the learning machine. Here, $f_g \in \mathcal{F}_g$ is the ground truth target function and $f_s \in \mathcal{F}_s$ is the student function, $E$ is the error, $|\cdot|_\theta$ is some function class capacity measure, $n$ is the number of data point, and finally $\frac{1}{2} \leq \beta_{sg} \leq 1$ is related to the learning rate acquiring small values close to $\frac{1}{2}$ for difficult problems while being close to 1 for easier problems. Note that $\epsilon_{sg}$ is the approximation error of the student function class $\mathcal{F}_s$ with respect to $f_g \in \mathcal{F}_g$. Building on the top of Lopez-Paz~\cite{lopez2015unifying}, we extend their result and investigate why and when introducing progressively intermediate-sized "teacher assistants" improves knowledge distillation. In Eqn.~\ref{eqn_est} student learns from scratch (STL-p). Next we let $f_t \in \mathcal{F}_t$ be the teacher function, then
\begin{equation}
    E(f_t) - E(f_g) \leq O\left( \frac{|\mathcal{F}_t|_\theta}{n^{\beta_{tg}}} \right) + \epsilon_{tg},
    \label{eqn_tg}
\end{equation}
where, $\beta_{tg}$ and $\epsilon_{tg}$ are correspondingly defined for teacher learning from scratch (STL-t). Moreover, we can transfer the knowledge of the teacher 
directly to the student and retrieve the baseline knowledge distillation (VanillaKD)~\cite{hinton2015distilling}. To simplify the argument we assume the training is done via pure distillation ($\lambda=1$):
\begin{equation}
    E(f_s) - E(f_t) \leq O\left( \frac{|\mathcal{F}_s|_\theta}{n^{\beta_{st}}} \right) + \epsilon_{st},
    \label{eqn_st}
\end{equation}
where $\beta_{st}$ and $\epsilon_{st}$ are associated to student learning from teacher. If we combine Eqn.~\ref{eqn_tg} and Eqn.~\ref{eqn_st} we get
\begin{equation}
    O\left( \frac{|\mathcal{F}_t|_\theta}{n^{\beta_{tg}}} + \frac{|\mathcal{F}_s|_\theta}{n^{\beta_{st}}} \right) + \epsilon_{tg} + \epsilon_{st} \leq O\left( \frac{|\mathcal{F}_s|_\theta}{n^{\beta_{sg}}} \right) + \epsilon_{sg}.
\end{equation}
to hold for VanillaKD to be effective. Lopez-Paz~\cite{lopez2015unifying} pointed out $|\mathcal{F}_t|_\theta$ should be small, otherwise the VanillaKD would not outperform STL-p. We acknowledge that similar to Lopez-Paz~\cite{lopez2015unifying}, we work with the upper bounds not the actual performance and also in an asymptotic regime. Here we built on top of their result and put an intermediate-sized teacher assistant $f_m\in \mathcal{F}_t$ between the small student and the large teacher:
\begin{equation}
    E(f_s) - E(f_m) \leq O\left( \frac{|\mathcal{F}_s|_\theta}{n^{\beta_{sm}}} \right) + \epsilon_{sm},
    \label{eqn_sm}
\end{equation}
and, then the intermediate-sized teacher itself learns from the large teacher
\begin{equation}
    E(f_m) - E(f_t) \leq O\left( \frac{|\mathcal{F}_m|_\theta}{n^{\beta_{mt}}} \right) + \epsilon_{mt},
    \label{eqn_mt}
\end{equation}
where, $\beta_{sm}, \epsilon_{sm}, \beta_{mt},$ and $\epsilon_{mt}$ are defined accordingly. Combining Eqn.~\ref{eqn_tg}, Eqn.~\ref{eqn_sm} and Eqn.~\ref{eqn_mt} leads to the following equation that needs to be satisfied in order to $Prog^2$-$ts$ outperforms VanillaKD and STL-p, respectively:
\begin{align}
    O\left( \frac{|\mathcal{F}_t|_\theta}{n^{\beta_{tg}}} + \frac{|\mathcal{F}_m|_\theta}{n^{\beta_{mt}}} + \frac{|\mathcal{F}_s|_\theta}{n^{\beta_{sm}}} \right) + \epsilon_{tg} + \epsilon_{mt} + \epsilon_{sm} \label{eqn_1} \\
    \leq O\left( \frac{|\mathcal{F}_t|_\theta}{n^{\beta_{tg}}} + \frac{|\mathcal{F}_s|_\theta}{n^{\beta_{st}}} \right) + \epsilon_{tg} + \epsilon_{st} \label{eqn_2} \\
    \leq O\left( \frac{|\mathcal{F}_s|_\theta}{n^{\beta_{sg}}} \right) + \epsilon_{sg}. \label{eqn_3}
\end{align}
We now discuss how the first inequality (Eqn.~\ref{eqn_1} $\leq$ Eqn.~\ref{eqn_2}) holds which entails $Prog^2$-$ts$ outperforms VanillaKD. To do so, we first hypothesize that a larger capacity gap between the teacher and student models reduces the student's learning efficiency: $\beta_{st} \leq \beta_{sm}$ and $\beta_{st} \leq \beta_{mt}$. This hypothesis aligns with the theoretical conclusions derived from the Eqn.(32) in our manuscript. Student learning directly from the large teacher is certainly more difficult than either student learning from the intermediate-sized teacher or the intermediate-sized teacher learning from the large teacher. Therefore, asymptotically speaking, $O\left( \frac{|\mathcal{F}_m|_\theta}{n^{\beta_{mt}}} + \frac{|\mathcal{F}_s|_\theta}{n^{\beta_{sm}}} \right) \leq O\left( \frac{|\mathcal{F}_s|_\theta}{n^{\beta_{st}}} \right)$ which in turn leads to $O\left( \frac{|\mathcal{F}_t|_\theta}{n^{\beta_{tg}}} + \frac{|\mathcal{F}_m|_\theta}{n^{\beta_{mt}}} + \frac{|\mathcal{F}_s|_\theta}{n^{\beta_{sm}}} \right) \leq O\left( \frac{|\mathcal{F}_t|_\theta}{n^{\beta_{tg}}} + \frac{|\mathcal{F}_s|_\theta}{n^{\beta_{st}}} \right)$. Moreover, according to assumption of Hinton~\cite{hinton2015distilling} we know $\epsilon_{mt} + \epsilon_{sm} \leq \epsilon_{st}$. These two together establish Eqn.~\ref{eqn_1} $\leq$ Eqn.~\ref{eqn_2}, which means that the upper bound of error in $Prog^2$-$ts$ is smaller than its upper bound in VanillaKD.

Similarly, for the second inequality Eqn.~\ref{eqn_2} $\leq$ Eqn.~\ref{eqn_3} which can use the hypothesis $\beta_{sg} \leq \beta_{st}$ and $\beta_{sg} \leq \beta_{tg}$ and $\epsilon_{tg} + \epsilon_{st} \leq \epsilon_{sg}$. Note that, these are asymptotic equations and hold when $n \rightarrow \infty$. In this hypothesis, when $|\mathcal{F}_t|_\theta$ is very large (infinite), then the inequality Eqn.~\ref{eqn_2} $\leq$ Eqn.~\ref{eqn_3} may not be valid and VanillaKD fails.

A second failure case, which is the central focus of our paper, arises when a large capacity gap exists between the teacher and student models. In this situation, the teacher model becomes highly confident in its own outputs. Consequently, the knowledge transferred from the teacher effectively degenerates into another form of hard, ground-truth labels (which means, e.g. $\beta_{st}$ is very small and close to $\beta_{sg}$). In this case, the error due to transfer from ground truth to teacher outweighs (Eqn.~\ref{eqn_2}) in comparison to (Eqn.~\ref{eqn_3}) and the inequality then becomes invalid (as $O\left( \frac{|\mathcal{F}_s|_\theta}{n^{\beta_{st}}} \right ) + \epsilon_{st} \approx O\left( \frac{|\mathcal{F}_s|_\theta}{n^{\beta_{sg}}} \right ) + \epsilon_{sg}$). In this case multi-stage method $Prog^2$-$ts$ turns out to be the key. By injecting an intermediate-sized "teacher assistant" between student and the large teacher we break the very small $\beta_{st}$ to two larger components $\beta_{sm}$ and $\beta_{mt}$ which makes the inequality (Eqn.~\ref{eqn_1} $\leq$ Eqn.~\ref{eqn_2}) holds for improving knowledge distillation.

\subsection{Gradient Variance Benchmark Metric and Correlation Analysis}
Inspired by prior studies on gradient variance, gradient noise scale, and training dynamics based on gradient norms~\cite{pascanu2013difficulty,mccandlish2018empirical,chen2018gradnorm,yu2020gradient,faghri2020study}, we have now benchmarked $S_{\mathrm{instability}}$ against the \textbf{Total Gradient Norm Variance (T-GNV)} as a reference indicator for optimization fluctuations:
\begin{equation}
    \text{T-GNV}_t = \text{Var}_{k \in [p-w+1, p]} (\|g_{\text{total}}^k\|_2)
\end{equation}
, where $g_{\text{total}}^k = \nabla_\theta (\mathcal{L}_{\text{task}}^k + \lambda \mathcal{L}_{\text{kd}}^k)$, $p$ denotes the current training iteration, and $w$ is the size of the sliding window used to compute the temporal variance. The sliding window size $w$ is empirically set to 10. A smaller window is more sensitive to short-term gradient fluctuations and captures rapid oscillations but introduces more noise. Conversely, a larger window such as $w=5$ or $w=10$ smooths short-term noise and emphasizes long-term trends while potentially introducing lag and attenuating local anomalies. A larger T-GNV indicates stronger temporal fluctuations in the overall optimization force applied to the student model. 

\begin{figure}[hbpt]
    \centering
    \includegraphics[width=0.95\linewidth]{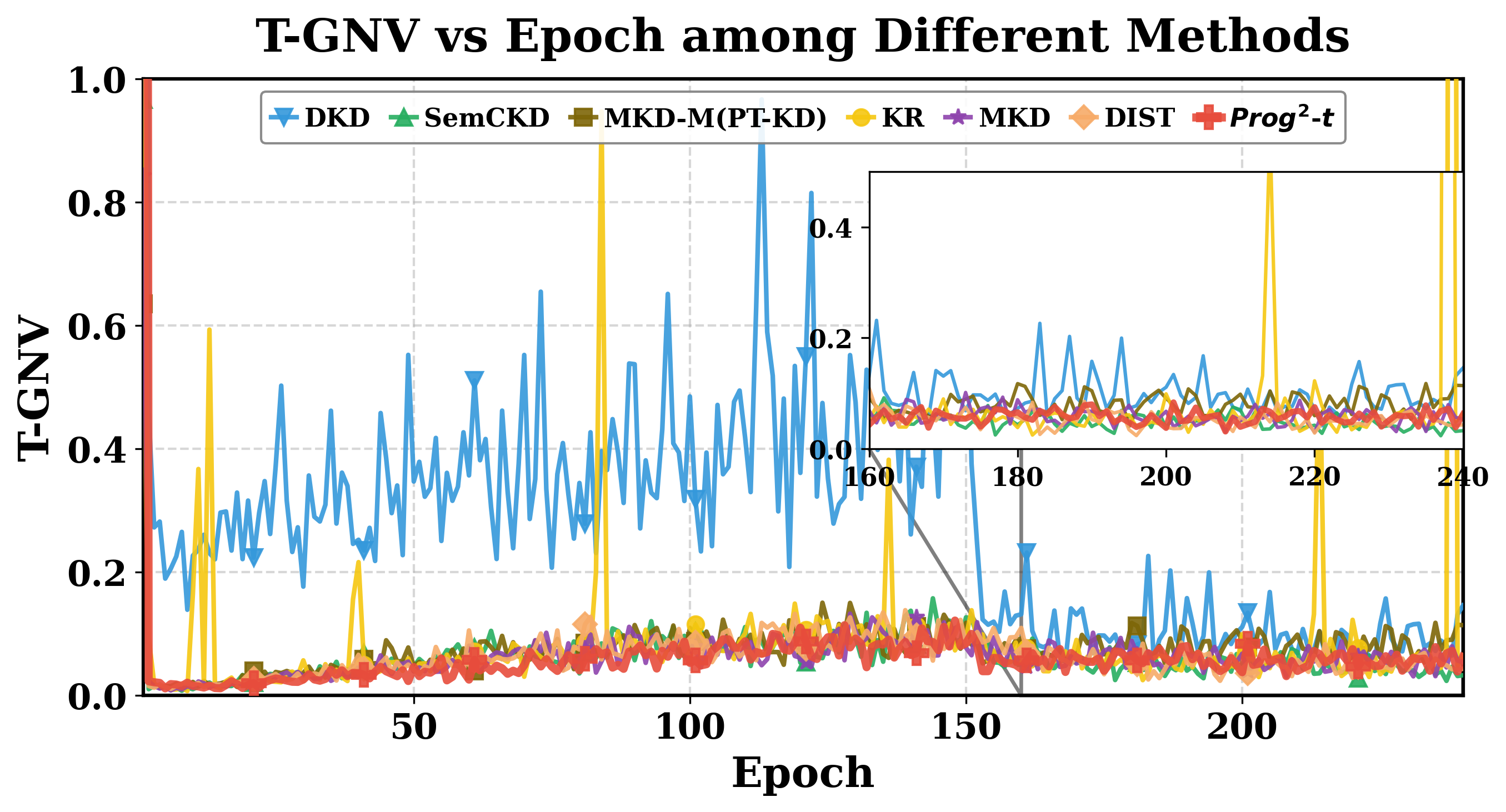}
    \caption{Visualization of the proposed gradient norm variance metric T-GNV across training epochs for the baselines.}
    \label{appen_fig:GNV}
\end{figure}

\begin{figure}
    \centering
    \includegraphics[width=0.96\linewidth]{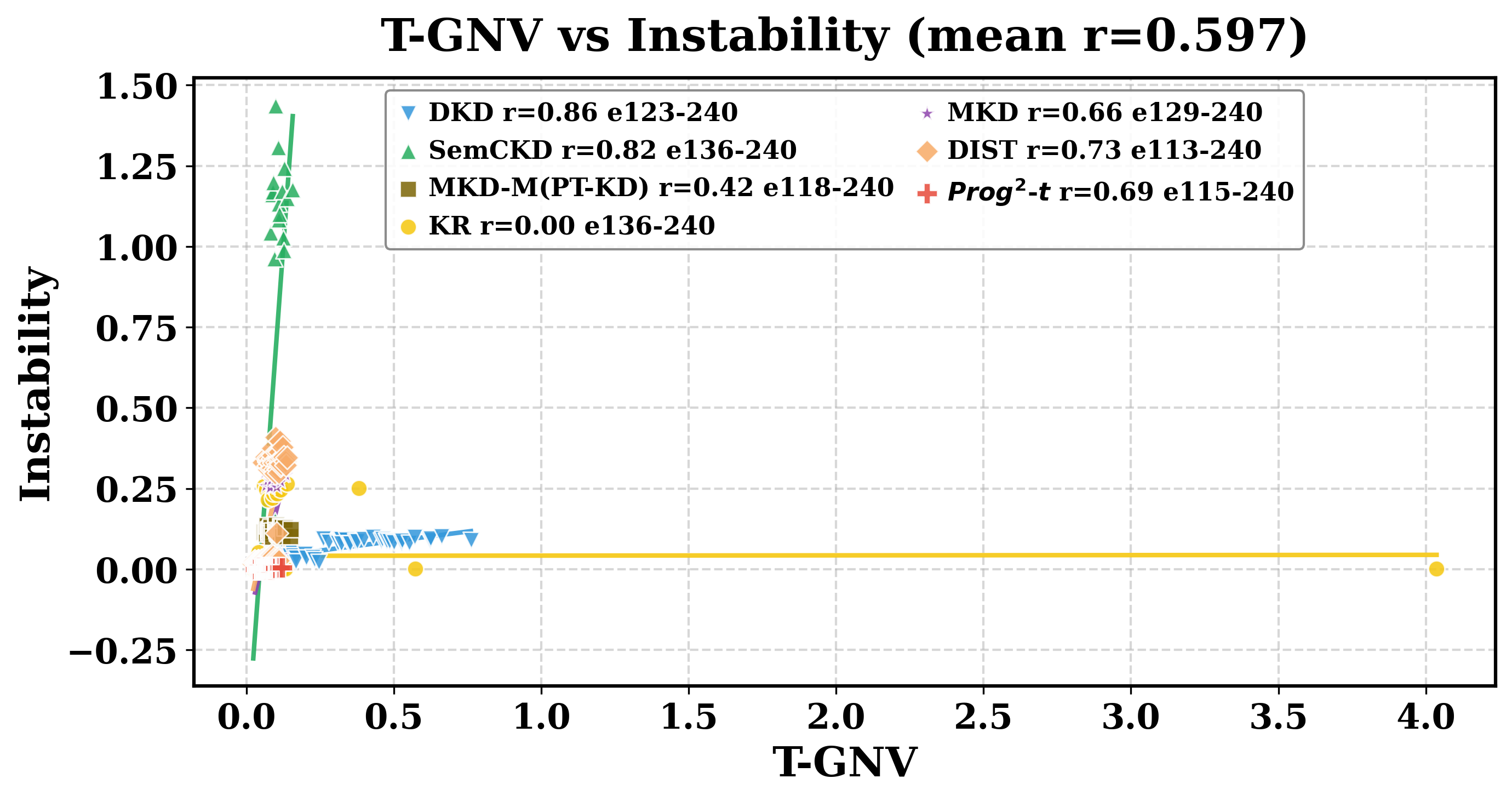}
    \caption{Correlation between the proposed instability metric $S_{\mathrm{instability}}$ and the Total Gradient Norm Variance (T-GNV) across training epochs. The strong positive correlation (Pearson $r = 0.86$) validates that $S_{\mathrm{instability}}$ effectively captures optimization fluctuations.}
    \label{appen_fig:GNV_S_correlation}
\end{figure}

As shown in Fig.~\ref{appen_fig:GNV}, we tracked both metrics across the training epochs of the baseline model. The results demonstrate a positive correlation (Fig~\ref{appen_fig:GNV_S_correlation}) between our $S_{\mathrm{instability}}$ and the Gradient Norm Variance (Pearson correlation coefficient $r = 0.86$ for DKD~\cite{zhao2022decoupled} method). In the Method section of the manuscript, we explicitly reiterate the references underlying the formulation of the existing instability metrics.

We evaluate the Pearson correlation coefficient between $\text{T-GNV}_t$ and the instability values $S_{\mathrm{instability}}$. For each method, the two sequences are aligned based on corresponding epochs. Given two aligned finite-valued sequences $\mathbf{x}=(x_1,x_2,\ldots,x_n)$ and $\mathbf{y}=(y_1,y_2,\ldots,y_n)$, where $x_i$ corresponds to the gradient norm variance and $y_i$ corresponds to $S_{\mathrm{instability}}$, the Pearson correlation coefficient is computed as follows.
\begin{equation}
    r(\mathbf{x},\mathbf{y})=\frac{\sum_{i=1}^{n}(x_i-\bar{x})(y_i-\bar{y})}{\sqrt{\sum_{i=1}^{n}(x_i-\bar{x})^2}\sqrt{\sum_{i=1}^{n}(y_i-\bar{y})^2}}
\end{equation}
. Here, $\bar{x}=\frac{1}{n}\sum_{i=1}^{n}x_i$ and $\bar{y}=\frac{1}{n}\sum_{i=1}^{n}y_i$. Practically, this coefficient is computed after filtering non-finite values. If the aligned sequences contain fewer than two valid points or exhibit a near-zero standard deviation in either sequence, the correlation is defined as 0.

The gradient dynamics during the early training phase typically do not follow a stable process. The initial epochs are affected by factors including random initialization, learning rate warmup, unaligned teacher and student representations, and unstabilized distillation losses. These factors cause the relationship between $\text{T-GNV}_t$ and $S_{\mathrm{instability}}$ to differ significantly from that in later stages. Therefore, we match epochs in stages to control the starting position for the Pearson correlation calculation. Denoting the gradient norm variance and instability for task $k$ as $G_k$ and $S_k$ respectively, we retain only the epochs where both $G_{k,e}$ and $S_{k,e}$ are available. For each method, the aligned sequence is defined as $\mathcal{D}_k=\{(e,G_{k,e},S_{k,e})\}$. For each possible start epoch $s$, we form a subset $\mathcal{D}_{k,s}=\{(e,G_{k,e}^{\mathrm{new}},S_{k,e}):e\geq s\}$ subject to the constraint $|\mathcal{D}_{k,s}|\geq L_{\min}$, where the default minimum segment length is $L_{\min}=3$. For each candidate starting point, we compute the correlation $r_{k,s}=r\left(\{G_{k,e}^{\mathrm{new}}\}_{e\geq s},\{S_{k,e}\}_{e\geq s}\right)$ and select the optimal start epoch that maximizes the correlation coefficient $s_k^\ast=\arg\max_s r_{k,s}$. This process reduces the interference of early anomalous gradients on the correlation evaluation. The resulting correlation between $\text{T-GNV}_t$ and $S_{\mathrm{instability}}$ for each method is presented in the corresponding figure.

\begin{table*}[t]
\centering
\small
\setlength{\tabcolsep}{3pt}
\renewcommand{\arraystretch}{1.25}
\caption{Summary of dataset-level experimental configurations.}
\label{tab:dataset_settings}
\begin{tabularx}{\textwidth}{
p{1.7cm}
p{1.8cm}
p{2.5cm}
p{2.9cm}
p{3.1cm}
p{2.1cm}
p{1.4cm}
p{1.5cm}
}
\toprule
\textbf{Dataset} 
& \textbf{Task} 
& \textbf{Data split / scale} 
& \textbf{Architecture} 
& \textbf{Optimizer, learning rate, and scheduler} 
& \textbf{Batch size} 
& \textbf{Epochs} 
& \textbf{Iterations per epoch} \\
\midrule

Cityscapes 
& Semantic segmentation 
& 2975 training images and 500 testing images from street scenes covering over 50 cities 
& SegNet with a VGG-16 backbone. The full teacher has 24.98M parameters. The pruned students are obtained by channel pruning, resulting in 6.25M parameters for $T_{25\%}$ and 1.57M parameters for $T_{6.3\%}$.
& Adam optimizer with learning rate $1\times 10^{-4}$ for the student, the teacher in the $Prog^2$-$ts$ setting, and the teacher-side adapter. The learning rate is reduced to $5\times 10^{-5}$ at epoch 100. The single-layer convolution for feature/channel alignment is optimized by Adam with learning rate $0.1$ and weight decay $5\times 10^{-4}$.
& 8 
& 200 for the single-stage setting; 600 for the multi-stage setting 
& 371 \\
\midrule

CIFAR-100 
& Image classification 
& 100 classes with 600 images per class; 500 training images and 100 testing images per class 
& VGG-16 backbone. The teacher and student models contain 14.78M and 3.71M parameters, respectively.
& SGD optimizer with learning rate $0.05$ for the student and the teacher-side adapter, Nesterov momentum $0.9$, and learning-rate decay factor $0.1$ at epochs 150, 180, and 210. The single-layer convolution for distillation-dimension alignment uses learning rate $0.1$.
& 128 
& 240 for the single-stage setting; 720 for the multi-stage setting
& 390 \\
\midrule

Tiny-ImageNet-200 
& Image classification 
& 200 classes; each class contains 500 training images, 50 validation images, and 50 testing images. Images are downsampled to $64\times64$ pixels.
& Same backbone setting as CIFAR-100 where applicable.
& SGD optimizer with learning rate $0.01$ for the student and the teacher-side adapter, Nesterov momentum $0.9$, and 240 training epochs. Other hyperparameters follow the CIFAR-100 setting where applicable.
& 128 
& Same settings as CIFAR-100 where applicable.
& 390 \\
\midrule

NYU-V2 
& Depth estimation 
& 3 cities, 464 scenes, and 1449 images, including 654 testing images 
& SegNet architecture which is the same as Cityscapes.
& SGD optimizer with learning rate $0.02$, momentum $0.9$, and weight decay $1\times 10^{-4}$ for the student. In the multi-stage distillation setting, the teacher model and teacher-side adapter use the same SGD configuration. The single-layer convolution for teacher-student feature alignment uses Adam with learning rate $0.1$.
& 2 
& Same settings as Cityscapes where applicable.
& 397 \\
\bottomrule
\end{tabularx}
\end{table*}

\begin{table*}[t]
\centering
\small
\setlength{\tabcolsep}{4pt}
\renewcommand{\arraystretch}{1.25}
\caption{
Summary of reproducibility, evaluation, baseline, and runtime-measurement protocols.}
\label{tab:general_protocol}
\begin{tabularx}{\textwidth}{
p{3.0cm}
p{13.8cm}
}
\toprule
\textbf{Reproducibility item} 
& \textbf{Setting / protocol} \\
\midrule

Number of independent runs 
& Each experiment is repeated three times, and the reported mean and standard deviation are computed over the three independent runs. \\

Random-seed protocol 
& Random seeds are not fixed to a predefined seed list. For each run, the seed is randomly sampled from \texttt{random.randint(0,10000)}. The exact sampled seed values are \texttt{None}. \\

Mean and standard deviation reporting 
& The notation ``$\pm$'' in the result tables denotes the standard deviation over three independent runs. \\

Dataset splits 
& The dataset splits follow the standard splits described for each benchmark. All methods are evaluated under the same dataset split protocol. \\

Preprocessing and data augmentation 
& Random cropping and random flipping are applied for data augmentation. The preprocessing protocols follow prior work, including MTAN~\cite{liu2019end}, InvPT~\cite{ye2022inverted}, and SemCKD~\cite{chen2021cross}, where applicable. \\

Early stopping 
& None. \\

Baseline implementation protocol 
& All baseline methods are reproduced within the same training pipeline based on their official open-source implementations. To ensure fair comparison, the hyperparameters and configurations of competing methods are maintained according to their original open-source settings whenever available. \\

Missing baseline hyperparameters 
& For experiments not provided in the official baseline implementations, the corresponding distillation weights are set empirically, generally to 1.0. \\

Teacher-model handling 
& On Cityscapes, the teacher model is re-run to obtain the reported result because of the potential overfitting risk of the teacher model on this dataset. \\

Hardware environment 
& Experiments are conducted on a server equipped with 6 NVIDIA RTX 3090 GPUs. Training uses 6 NVIDIA RTX 3090 GPUs where applicable. The server uses a Supermicro 4029GP platform with a 20-thread 2.4GHz CPU, 128GB memory, and a 2000W power supply. GPU memory is 24GB. \\

Software environment 
& Python 3.10, CUDA 11.8, and PyTorch 2.7.1. The operating system and cuDNN version are \texttt{None}. \\

Runtime / timing protocol 
& Baseline runtime measurements are performed in isolation on a single NVIDIA RTX 3090 GPU to determine training time on all datasets. The time includes data loading and evaluation overhead. \\

Code availability 
& It will be open-sourced on the GitHub community after the paper is accepted. \\

\bottomrule
\end{tabularx}
\end{table*}

\begin{table*}[t]
\centering
\small
\setlength{\tabcolsep}{4pt}
\renewcommand{\arraystretch}{1.25}
\caption{Summary of model compression and teacher-student configurations.}
\label{tab:ts_settings}
\begin{tabularx}{\textwidth}{
>{\raggedright\arraybackslash}p{2.3cm}
>{\raggedright\arraybackslash}X
>{\raggedright\arraybackslash}X
>{\raggedright\arraybackslash}X
}
\toprule
\textbf{Dataset} 
& \textbf{Teacher setting} 
& \textbf{Student setting} 
& \textbf{Compression / pruning protocol} \\
\midrule

Cityscapes 
& Full SegNet model with a VGG-16 backbone. The channel configuration is $[64,128,256,512,512,512,256,128,64,64]$, and the parameter size is 24.98M.
& Two pruned students are used. The $T_{25\%}$ student has channel configuration $[32,64,128,256,256,256,128,64,32,32]$ and 6.25M parameters. The $T_{6.3\%}$ student has channel configuration $[16,32,64,128,128,128,64,32,16,16]$ and 1.57M parameters.
& The students are obtained by channel pruning following~\cite{li2017pruning}. The shared-block channels are pruned by 50\% and 75\%, respectively. \\
\midrule

CIFAR-100 
& VGG backbone teacher with 14.78M parameters or ResNet 32$\times$4 backbone teacher with 7.43M parameters.
& VGG backbone student with 3.71M parameters or ResNet 32 backbone student with 0.47M parameters.
& Same settings as Cityscapes on VGG backbone where applicable or settings on the ResNet backbone that gradually decreases based on the number of different network layers (ResNet 32$\times$4: 7.43M $\rightarrow$ ResNet 8$\times$4\_double: 4.88M $\rightarrow$ ResNet 110: 1.74M $\rightarrow$ ResNet 8$\times$4: 1.23M $\rightarrow$ ResNet 56: 0.86M $\rightarrow$ ResNet 44: 0.67M $\rightarrow$ ResNet 32: 0.47M).\\
\midrule

Tiny-ImageNet-200 
& \multicolumn{3}{>{\raggedright\arraybackslash}p{\dimexpr\textwidth-2.3cm-8\tabcolsep\relax}}{Same settings as CIFAR-100 where applicable.} \\
\midrule

NYU-V2 
& Same settings as Cityscapes where applicable or Segformer-based Mit-B4 backbone teacher with 63.99M parameters.
& Same settings as Cityscapes where applicable or Segformer-based Mit-B0 backbone student with 3.71M parameters.
& Same settings as Cityscapes where applicable or settings on the Segformer backbone that gradually decreases based on the number of different network layers (Mit-B4: 63.99M $\rightarrow$ Mit-B3: 47.22M $\rightarrow$ Mit-B2: 27.35M $\rightarrow$ Mit-B1: 13.68M$\rightarrow$ Mit-B0: 3.71M).\\
\bottomrule
\end{tabularx}
\end{table*}

\bibliographystyle{IEEEtran}
\bibliography{TMM}